\PassOptionsToPackage{usenames,dvipsnames}{xcolor}
\documentclass[manuscript,screen]{acmart}

\AtBeginDocument{%
  \providecommand\BibTeX{{%
    \normalfont B\kern-0.5em{\scshape i\kern-0.25em b}\kern-0.8em\TeX}}}

\setcopyright{acmcopyright}
\copyrightyear{2024} 
\acmYear{2024} 
\acmDOI{aaaaaaa.aaaaaaa}

\usepackage[dvipsnames]{xcolor}
\usepackage{pbox}

\usepackage{longtable}

\usepackage{xltabular}

\usepackage{adjustbox}
\usepackage{makecell}
\usepackage{rotating}
\usepackage{multirow}
\usepackage{graphicx}
\usepackage{subcaption}
\usepackage{pifont}
\usepackage{xcolor}
\usepackage{array}

\usepackage{caption}
\usepackage{booktabs}
\usepackage{soul}

\definecolor{bluenode}{HTML}{3399FF} 

\definecolor{pinknode}{HTML}{FF00FF} 

\definecolor{yellownode}{HTML}{FFCC00}

\definecolor{greennode}{HTML}{99FF00} 

\definecolor{graynode}{HTML}{999999}

\newcommand\blfootnote[1]{%
  \begingroup
  \renewcommand\thefootnote{}\footnote{#1}%
  \addtocounter{footnote}{-1}%
  \endgroup
}

\usepackage{tikz}
\usetikzlibrary{arrows,shapes,positioning,shadows,trees}

\tikzset{
  basic/.style  = {draw, text width=3cm, font=\sffamily, rectangle},
  root/.style   = {basic, rounded corners=2pt, thin, align=center, fill=TealBlue!5},
  level 2/.style = {basic, rounded corners=6pt, thin,align=center, fill=TealBlue!10, text width=3cm},
  level 3/.style = {basic, thin, align=left, fill=TealBlue!15, text width=5cm},
  level 4/.style = {basic, thin, align=left, fill=TealBlue!20, text width=5.1cm},
}

\begin{document}


\title{A Survey of Pre-trained Language Models for Processing Scientific Text}



\author{Xanh Ho$^\ast$}
\affiliation{%
  \institution{National Institute of Informatics}
  \city{Tokyo}
  \country{Japan}}
\email{xanh@nii.ac.jp}

\author{Anh Khoa Duong Nguyen$^\ast$}
\affiliation{%
\institution{Independent Researcher}
  \country{Vietnam}}
\email{dnanhkhoa@live.com}

\author{An Tuan Dao$^\ast$}
\email{dtan@nii.ac.jp}
\author{Junfeng Jiang$^\ast$}
\email{jiangjf@is.s.u-tokyo.ac.jp}
\author{Yuki Chida$^\ast$}
\email{chida@nii.ac.jp}
\affiliation{%
  \institution{The University of Tokyo}
  \city{Tokyo}
  \country{Japan}}

\author{Kaito Sugimoto$^\ast$}
\affiliation{%
  \institution{NTT Communications Corporation}
  \city{Tokyo}
  \country{Japan}}
\email{kaito.sugimoto@ntt.com}

\author{Huy Quoc To}
\affiliation{%
  \institution{University of Information Technology}
  \city{Ho Chi Minh}
  \country{Vietnam}}
\email{huytq@uit.edu.vn}

\author{Florian Boudin}
\affiliation{%
  \institution{JFLI, CNRS, National Institute of Informatics}
  \city{Tokyo}
  \country{Japan}
  ~~~\&~~~
  \institution{LS2N, Université de Nantes}
  \city{Nantes}
  \country{France}
  }
\email{florian.boudin@univ-nantes.fr}


\author{Akiko Aizawa}
\affiliation{%
 \institution{National Institute of Informatics}
 \city{Tokyo}
 \country{Japan}}
\email{aizawa@nii.ac.jp}

\renewcommand{\shortauthors}{Ho et al.}

\newcommand{\rewrite}[1]{\textcolor{blue}{#1}}
\newcommand{\verify}[1]{\textcolor{orange}{#1}}
\newcommand{\todo}[1]{\textcolor{red}{#1}}
\newcommand{\checked}[1]{\textcolor{pink}{#1}}

\newcommand{\newupdate}[1]{\textcolor{RubineRed}{#1}}

\begin{abstract}

The number of Language Models (LMs) dedicated to processing scientific text is on the rise.
Keeping pace with the rapid growth of scientific LMs (SciLMs) has become a daunting task for researchers.
To date, no comprehensive surveys on SciLMs have been undertaken, leaving this issue unaddressed.
Given the constant stream of new SciLMs, appraising the state-of-the-art and how they compare to each other remain largely unknown.
This work fills that gap and provides a comprehensive review of SciLMs, including an extensive analysis of their effectiveness across different domains, tasks and datasets, and a discussion on the challenges that lie ahead.\footnote{Resources are available at \url{https://github.com/Alab-NII/Awesome-SciLM}.}


%

\end{abstract}





\maketitle

\blfootnote{$^\ast$The first six authors contributed equally to this research.}



\section{Introduction}



The introduction of Pre-trained Language Models (PLMs)~\cite[\textit{inter alia}]{devlin-etal-2019-bert,Radford2018ImprovingLU,liu2019roberta,clark2020electra,JMLR:v21:20-074} and Large Language Models (LLMs)~\cite[\textit{inter alia}]{brown2020language,chowdhery2022palm,openai2023gpt4,touvron2023llama} has had a profound impact on the landscape of NLP research~\cite{WANG2022}, showcasing their remarkable effectiveness throughout a variety of NLP tasks~\cite{devlin-etal-2019-bert,brown2020language}. 
%
This shift has prompted the development of LMs that are capable of solving complex tasks, often involving language understanding, logical inference, or commonsense reasoning, in both general and specific domains.
This is notably the case in the scientific domain, where many well-studied tasks, such as Named Entity Recognition (NER), Relation Extraction (RE), Question-Answering (QA), document classification,  or summarization to name a few, have benefited from the utilization of LMs.
%
One pivotal factor contributing to these successes is the abundant availability of scientific texts.
For example, almost 2 million biomedical articles where added to PubMed in 2022, contributing to a cumulative total of 36 million publications.\footnote{\url{https://www.nlm.nih.gov/bsd/medline_pubmed_production_stats.html}}
The ever increasing growth in the volume of scientific literature enables LMs to effectively learn and ingest scientific knowledge, fostering their capability to excel in a wide array of tasks.

However, despite the wealth of research on LMs for processing scientific texts (hereby reffered to as SciLMs), there is a currently no comprehensive survey on this subject.
%
Thus, a complete picture of the evolution of SciLMs over the past few years is currently lacking, resulting in an unclear understanding of the actual state of progress in these models.
This paper aims to bridge this gap and offers the first comprehensive review of SciLMs.
It provides the descriptions for over 110 models published in the last few years, conducts an extensive analysis of their effectiveness across different domains, tasks and datasets, and initiates a discussion on the challenges that will likely shape future research.

In the remainder of this section, we first show the overall structure of our survey.
We then outline the scope of our research, focusing on six aspects: time scope, target language models, target domains, target scientific text, target languages, and target modalities. 
We subsequently explain how we collected related papers. 
Following this, we clarify the distinctions between our survey and existing surveys. 
Finally, we present an overview of the landscape of SciLMs over the past few years in the form of an evolutionary tree.

\subsection{Structure of the Paper}

\begin{figure}[h]
  \centering
  \includegraphics[width=\linewidth]{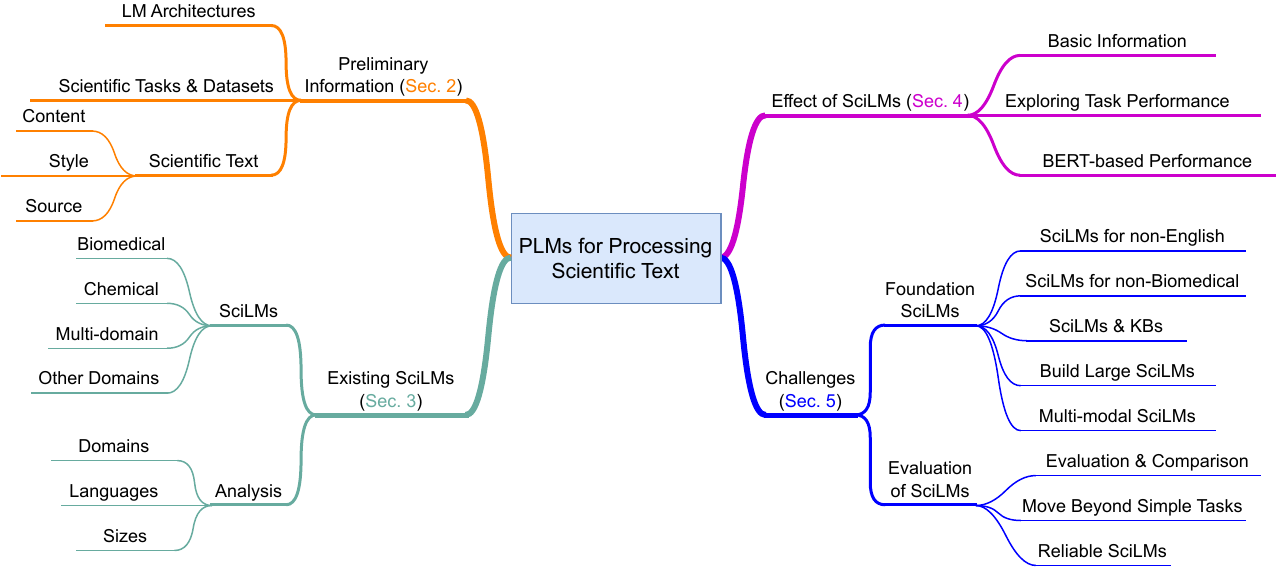}
  \caption{Overall structure of our survey.
  }
\label{overall_fig}
\end{figure}

Figure~\ref{overall_fig} shows the overall structure of our survey.
We provide background information in Section \ref{sec_background}, including details about LM architectures and existing scientific tasks, as well as the distinctions between scientific text and text in other domains.
Next, in Section \ref{sec_exist_lms}, we systematically review all existing PLMs and LLMs for processing scientific text from 2019 to September 2023, analyzing their popularity based on three main aspects: domain, language, and size.
After that, we analyze the effectiveness of SciLMs by considering the performance changes over time across different tasks and datasets in Section \ref{sec_effectiveness}.
We conclude by highlighting current challenges and open questions for future studies in Section \ref{sec_challenges}.

\subsection{Survey Scope}
\hfill

\textbf{Time Scope.}
The BERT model was released in October 2018. 
Our focus is on exploring PLMs for processing scientific text; therefore, we mainly consider papers released \textbf{from 2019 to September 2023}.

\textbf{Target LMs.}
Recently, most state-of-the-art (SOTA) approaches for NLP tasks are based on PLMs and LLMS. 
They have made a significant impact on scientific research; for example, the release of GPT-4 \cite{openai2023gpt4} has changed the research directions in NLP \cite{WANG2022,10.1145/3560815}.
In the scope of this paper, we use the term `language models' (LMs) to refer to both PLMs and LLM in general. 
We designate PLMs and LLMs proposed for processing scientific text as SciLMs.
In specific cases where we aim to emphasize the effectiveness of LLMs, we may use two terms: LLMs and SciLLMs, instead of LMs and SciLMs.
It is noted that we do not consider general LMs fine-tuning on scientific downstream tasks are SciLMs.
Furthermore, we exclusively concentrate on neural network LMs due to their popularity and superior performance compared to non-neural network ones on many tasks.
For example, we do not consider n-gram \cite{brown-etal-1992-class}, Conditional Random Fields \cite{10.5555/645530.655813}, and Hidden Markov Model \cite{baum1966statistical} in our research.

\textbf{Target Domains.}
The purpose of our survey is to explore LMs for processing scientific text; therefore, within the scope of this paper, we aim to cover as many scientific domains as possible. 
Specifically, the SciLMs in our survey span various domains, including computer science (CS), biomedicine, chemistry, and mathematics. 
We remain open to extending the domains in our research as new SciLMs are proposed for additional fields.

\textbf{Target Scientific Text.}
In this survey, we consider an LM to be a SciLM when its training data includes scientific text.
In specialized domains like chemistry, there are specific types of text that are used to represent and communicate information unique to that field such as the molecule structures in SMILE \cite{doi:10.1021/ci00057a005} or SEFIE \cite{krenn2020self}.
The LMs that are trained on these strings are also considered in this study.

\textbf{Target Languages.}
Similar to `Target Domains', we also aim to encompass as many available SciLMs in different languages as possible in order to conduct a comprehensive review of LMs for processing scientific text.

\textbf{Target Modalities.}
In addition to the text in scientific papers, there are other types of information, such as images or tables. 
However, in the scope of this paper, we mainly focus on LMs for scientific text.
For a more comprehensive discussion on multi-modal PLMs, we refer readers to an extensive survey available here~\cite{wang2022MMPTMSurvey,zhang2023visionlanguage}.


\subsection{Papers Collection}
Due to the rapid growth of the topic, after the first phase where we comprehensively obtain related papers (from 2019 to February 2023).
We also perform the second phase to update new SciLMs in our list. 


\textbf{Phase 1: From 2019 to February 2023.}
We use SciBERT~\cite{beltagy-etal-2019-scibert} as a seed paper, then manually check cited papers of SciBERT.
At the time of writing this paper,\footnote{February 2023} SciBERT has 1,857 citations.\footnote{Due to the limitation of Google Scholar, our review is limited to the top 1,000 citations.}
In addition, to ensure the coverage of our survey, we also use BERT as a seed paper, then manually check cited papers of BERT by using the function `Search within citing articles' of Google Scholar with three keywords: `\textit{scientific text}', `\textit{scientific papers}', and `\textit{scientific articles}'.
Moreover, when reading related papers and related survey papers, we also check mentioned papers in the paper, if we find that we are missing any related papers, we add them to our study.

    
    

\textbf{Phase 2: From February 2023 to September 2023.}
In this phase, we only check the cited papers of SciBERT from 2023 in Google Scholar.
At the time of checking (September 13), SciBERT had 639 citations.
We manually checked the titles and/or abstracts of these cited papers to find the newly released SciLMs.

\subsection{Related Surveys}
\citet{HAN2021225,10.1016/j.jbi.2021.103982,WANG2022,wang2023pretrained}, and \citet{zhao2023survey} are the most similar papers to ours. 
%
Specifically, both \cite{HAN2021225} and \cite{WANG2022} delve into PLMs themselves, discussing related topics such as the history of PLMs and LM architectures.
%
Meanwhile, both \cite{10.1016/j.jbi.2021.103982} and \cite{wang2023pretrained} focus on surveying PLMs in the biomedical domain. 
%
%
They summarize numerous existing PLMs which we also cover in Section ~\ref{sec_exist_lms}.
%
However, our paper concentrates more on exploring PLMs across all domains, not solely in the biomedical domain.
%
%
Additionally, we include sections that compare scientific text with text in other domains (Section~\ref{sec_comparison}) and analyse the effectiveness of SciLMs (Section~\ref{sec_effectiveness}).
%
%
In contrast, \citet{zhao2023survey} summarize newly released LLMs but do not emphasize scientific text as our paper does.
%







\begin{figure}[h]
  \centering
  \includegraphics[width=\linewidth]{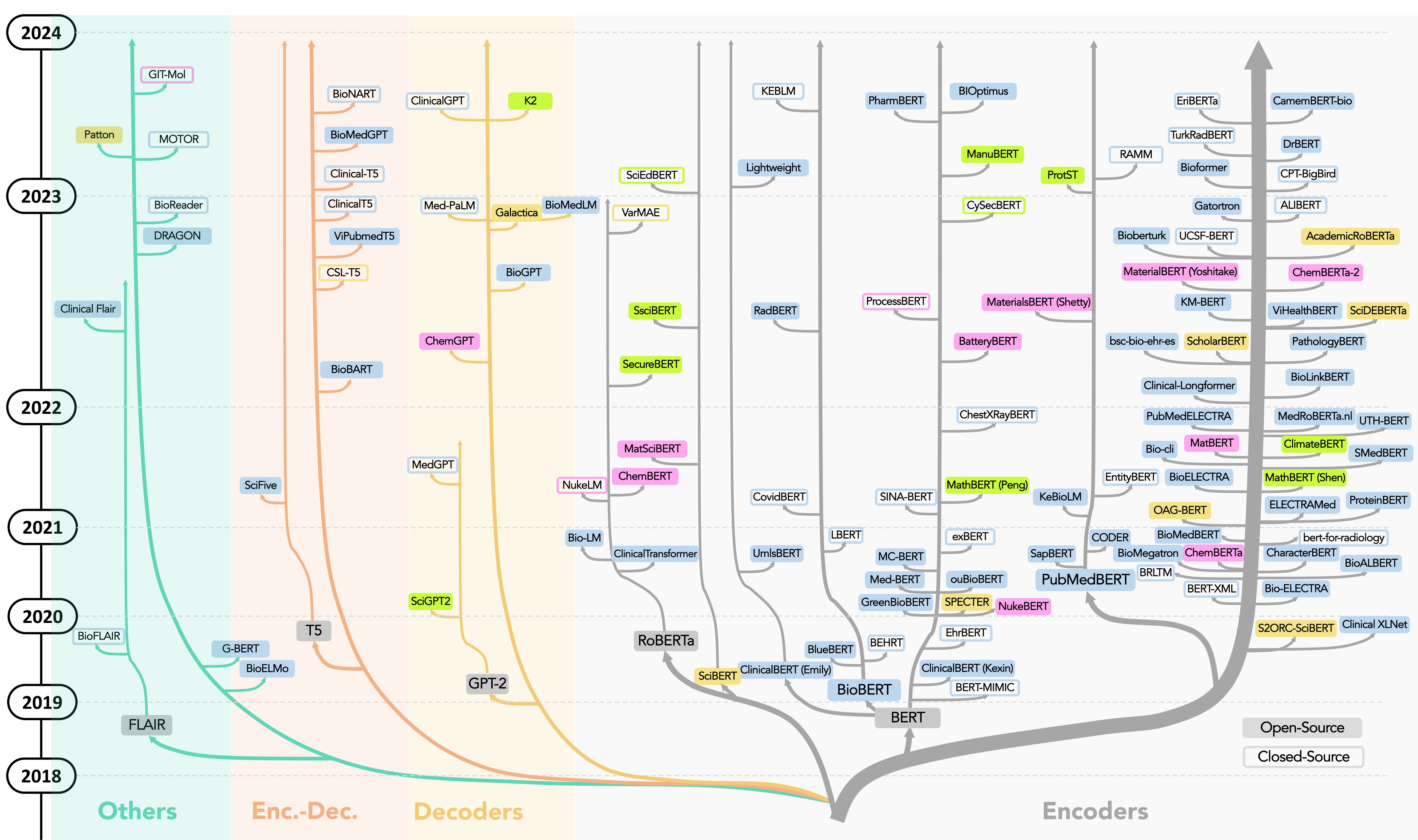}
  \caption{
    %
    Evolutionary tree of SciLMs.
The nodes are color-coded based on their domains: \colorbox{bluenode!20}{blue} for biomedical, \colorbox{pinknode!20}{pink} for chemical, \colorbox{yellownode!20}{yellow} for multi-domain, \colorbox{greennode!20}{green} for other domains, and \colorbox{gray!20}{gray} for general domain models. The node is filled in white if the model is closed-source; otherwise, it is open-source.
The English version of a model is used if it has multiple languages, and the most efficient variant is used if a model has multiple variants.
SciLMs that use continual pretraining are represented as children of the model whose weights they initialize. Only popular models are depicted as parent nodes in the tree for clarity. SciLMs trained from scratch are placed as leaves in the rightmost branch.
  }
\label{fig_tree}
\end{figure}

\subsection{Landscape of SciLMs}

Figure \ref{fig_tree} presents a tree illustrating the landscape of SciLMs from 2019 to September 2023.
We observe the following points:
(1) The tree is quite large and dense, indicating the existence of numerous proposed SciLMs during the period from 2019 to September 2023.
Additionally, more and more models are proposed each year, indicating an increase in the number of models annually.
(2) Most SciLMs are encoder-based models (91 models), and among these, BERT-based models are most commonly used. This suggests that research on SciLMs is still primarily focused on encoder-based architectures and has not yet generalized to other architectures.
(3) As depicted in the tree, many nodes are blue (biomedical domain nodes), indicating that the majority of SciLMs are proposed for the biomedical domain.
For details about all SciLMs in our study, as well as additional observations about existing SciLMs, we refer readers to Section \ref{sec_exist_lms}.



\section{Preliminary Information}
\label{sec_background}
We first briefly summarize existing LM architectures and other related information.
We then introduce existing tasks in processing scientific text.
Finally, we present the distinctions between scientific text and text in other domains.

\subsection{A Brief Summary of Existing LM Architectures}


\subsubsection{The core backbone}

Today, almost all LMs rely on the Transformer architecture~\cite{vaswani2017attention}.
Transformer is a type of neural network designed for sequence modeling. The core idea behind this architecture is to use self-attention mechanisms~\cite{cheng-etal-2016-long, DBLP:conf/iclr/LinFSYXZB17} without recurrent or convolutional networks. 
This allows the efficient computation of representations for input and output sequences, regardless of their lengths.

Transformer is comprised of two types of modules: the encoder and the decoder.
Both modules are made up of stacked network layers, each consisting of a self-attention sub-layer and a feed-forward sub-layer. A notable feature of the Transformer architecture lies in its multi-head attention mechanism, which enables the parallel computation of different attention patterns.

\subsubsection{Types of architecture}

Current LMs can be categorized into four distinct types: Encoder-Decoder style, Encoder-only style, Decoder-only style, and other styles.
The following subsections briefly introduce each type of LM.

\textbf{Encoder-Decoder style.}
LMs following the Encoder-Decoder style are based on the Transformer architecture and differ according to their pre-training objectives.
%
For example, T5 \cite{JMLR:v21:20-074} is trained on span corruption prediction task and BART \cite{lewis-etal-2020-bart} is trained on five tasks, which can be largely split into text infilling and sentence permutation.

\textbf{Encoder-only style.}
LMs of Encoder-only style are built upon the Transformer encoder and also differ according to their selected pre-training objectives.
Masked Language Modeling (MLM), Next Sentence Prediction (NSP), Sentence Order Prediction (SOP), and Replaced Token Prediction (RTP) are basic objectives. For instance, the origin of Encoder-only LM is BERT~\cite{devlin-etal-2019-bert} and it is trained by MLM and NSP. RoBERTa~\cite{liu2019roberta}, which is intended to improve BERT, is trained only MLM.
ELECTRA~\cite{clark2020electra} uses RTP with a smaller LM, generating token-replaced sentences and predicting the replaced token.
This type of LMs can utilize all of the information of the sentence by its nature (this feature is sometimes called bidirectional), and are often used in classification tasks.

\textbf{Decoder-only style.}
LMs of Decoder-only style are based on the Transformer decoder and its main pre-training objective is Next Token Prediction (NTP). This kind of models are used for generative tasks such as QA, dialog generation, and so on.

\textbf{Other types of LMs.}
In this part, we present several LMs that appear in Tables \ref{table_lms_1}, \ref{table_lms_2}, or \ref{table_lms_3} but do not belong to the three aforementioned types.

\begin{enumerate}
    \item ELMo~\cite{peters-etal-2018-deep} is an LM that concatenates the representations of forward and backward LSTMs. ELMo introduces the concept of contextualized word embeddings, which has later been standardized in Transformer-based LMs.
    \item Flair~\cite{akbik-etal-2019-flair} is a text embedding library supporting various types of LMs (including those that do not use the Transformer architecture).
    \item Graph Neural Network (GNN) \cite{DBLP:journals/debu/HamiltonYL17, DBLP:journals/tnn/WuPCLZY21} is a type of neural networks for processing graph structures. GNNs are often combined with LMs to handle texts and knowledge graphs  simultaneously.
\end{enumerate}

\subsection{Existing Tasks and Datasets in Scientific Articles}

\begin{figure}[b!]
    \centering
    \resizebox{.8\textwidth}{!}{%
        \tikzset{
  basic/.style  = {draw, text width=3cm, font=\sffamily, rectangle},
  root/.style   = {basic, rounded corners=2pt, thin, align=center, fill=TealBlue!5},
  level 2/.style = {basic, rounded corners=6pt, thin,align=center, fill=TealBlue!10, text width=3cm},
  level 3/.style = {basic, thin, align=left, fill=TealBlue!15, text width=5cm},
  level 4/.style = {basic, thin, align=left, fill=TealBlue!20, text width=5.1cm},
}

\begin{tikzpicture}[
  level 1/.style={sibling distance=75mm},
  edge from parent/.style={->,draw},
  >=latex]


\node [level 2] (c1) {Text Mining};
\node [level 2, right of = c1, xshift=7cm] (c2) {Text Application};

\node [level 3, below of = c1, xshift=2cm, yshift=.4cm] (c11) {Knowledge Graph Construction};
\node [level 4, below of = c11, xshift=1cm, yshift=-.4cm] (c12) {Named entity recognition\\
                                                        Relation extraction\\
                                                        Named entity disambiguation\\
                                                        Coreference resolution\\
                                                        Entity linking, ...};
\node [level 3, below of = c12, xshift=-1cm, yshift=-.4cm] (c13) {Scientific Dataset Construction};
\node [level 4, below of = c13, xshift=1cm, yshift=0cm] (c14) {Analyzing the research data\\
                                                        Science research corpus \\
                                                        Workflow scientific mining};

\node [level 3, below of = c2, xshift=2cm, yshift=.4cm] (c21) {Text Understanding};
\node [level 4, below of = c21, xshift=1cm, yshift=-.6cm] (c22) {Scientific verification\\
                                                       Natural language inference\\
                                                       Document analysis\\
                                                       Semantic search\\
                                                       Citation recommendation\\
                                                       Scientific reviewing, ...};
\node [level 3, below of = c22, xshift=-1cm, yshift=-.55cm] (c23) {Text Generation};
\node [level 4, below of = c23, xshift=1cm, yshift=-.15cm] (c24) {Automatic related work generation\\
                                                        Automated evidence synthesis\\
                                                        Citation text generation\\
                                                        Summarization};
\node [level 3, below of = c24, xshift=-1cm, yshift=-0.2cm] (c25) {Text Understanding \& Generation};
\node [level 4, below of = c25, xshift=1cm, yshift=.35cm] (c26) {Question answering};

\draw[->] ([xshift=10pt]c1.south west) |- (c11.west);
\draw[->] ([xshift=10pt]c1.south west) |- (c13.west);
\draw[->] ([xshift=10pt]c11.south west) |- (c12.west);
\draw[->] ([xshift=10pt]c13.south west) |- (c14.west);

\draw[->] ([xshift=10pt]c2.south west) |- (c21.west);
\draw[->] ([xshift=10pt]c2.south west) |- (c23.west);
\draw[->] ([xshift=10pt]c2.south west) |- (c25.west);
\draw[->] ([xshift=10pt]c21.south west) |- (c22.west);
\draw[->] ([xshift=10pt]c23.south west) |- (c24.west);
\draw[->] ([xshift=10pt]c25.south west) |- (c26.west);

\end{tikzpicture}
    }
    \caption{Existing tasks in processing scientific text.}
    \label{fig_task_classification}
\end{figure}
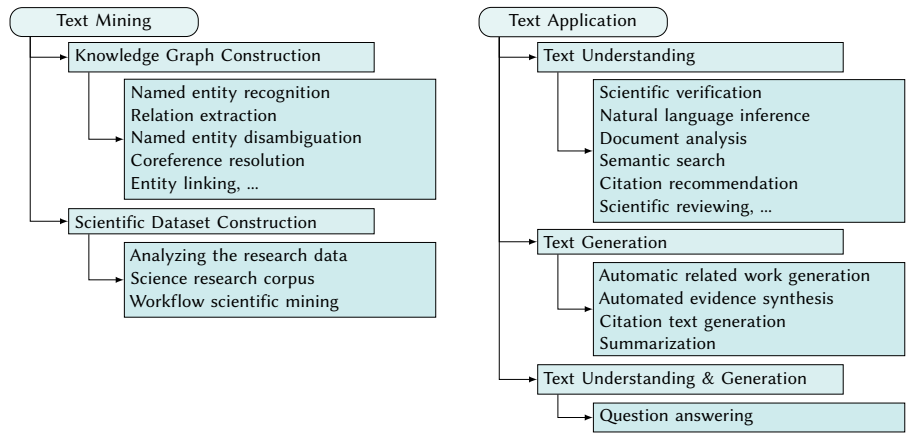

We divide existing tasks in scientific articles into two sub-groups: \textbf{scientific text mining} and \textbf{scientific text application}.
Figure~\ref{fig_task_classification} presents a summary of the tasks within each group in our classification.
%
%
%

\subsubsection{Scientific Text Mining}
\label{sub_sub_acquisition}

The purpose of this group is to mine existing scientific articles to extract knowledge, such as constructing a scientific dataset or a knowledge graph (KG) from unstructured text.
We begin by discussing tasks related to KG construction, such as NER and RE.
Following that, we present essential information on the list of existing scientific research datasets that we are aware of.

\textbf{Knowledge Graph Construction.}
There are many tasks related to KG construction. 
As listed in Figure \ref{fig_task_classification}, most of these tasks involve entities or relations within the KG, such as NER, entity linking, and RE.
In Section \ref{sec_effective_basic_infor}, when we analyse the effectiveness of SciLMs, we find that many tasks within this group are among the top 20 most popular tasks used for evaluation. 
Specifically, these tasks include NER, RE, PICO extraction, entity linking, disambiguation, and dependency parsing.
%
We also observe that among the top 20 popular datasets used for evaluation, many belong to the KG construction group. 
We provide important information related to these datasets in Table~\ref{tab_details_kg_construction}.

\begin{table}[hb!]

  \caption{Details information of the popular datasets related to KG construction. 
  (Sorted based on popularity in Section \ref{sec_effectiveness}).
  }
  \label{tab_details_kg_construction}
  \small

  \begin{tabular}{l l l l r l c}
    \toprule
\textbf{Rank} &  \textbf{Year} &  \textbf{Dataset} & \textbf{Task} &  \textbf{Size} & \textbf{Source} & \textbf{Domain}  \\
    \midrule

1 &  2014   &  NCBI-disease \cite{DOGAN20141} & NER & 793 &  PubMed abstracts  & Biomedical \\  

2 &  2016  & BC5CDR-disease \cite{10.1093/database/baw068} & NER & 1,500 & PubMed abstracts  & Biomedical \\ 
   
3 & 2004   &  JNLPBA \cite{collier-kim-2004-introduction} & NER & 2,404 & MEDLINE abstracts  & Biomedical \\  

4 &  2016  &  ChemProt \cite{10.1093/database/bav123} & RE & 1,820 &  PubMed abstracts &  Biomedical \\  
   
5 &  2016  &  BC5CDR-chemical \cite{10.1093/database/baw068} & NER & 1,500 & PubMed abstracts  & Biomedical \\

7 & 2008   &  BC2GM \cite{overview_smith} & NER & 20,000 &  MEDLINE sentences & Biomedical \\  
   
8 &  2013  &  DDI \cite{HERREROZAZO2013914} & RE & 1,025 &  MEDLINE abstracts  &  Biomedical \\

9 & 2011   &  i2b2 2010 \cite{10.1136/amiajnl-2011-000203} & NER or RE & 871 &  Patient reports  & Clinical \\ 
   
11 & 2015   &  GAD \cite{extraction_bravo} & RE & 5,330 & PubMed abstracts  &  Biomedical \\ 
   
12 & 2015   &  BC4CHEMD \cite{Krallinger2015CHEMDNERTD} & NER &  10,000 & PubMed abstracts  & Biomedical \\   
   
14 & 2013   &  Species-800 \cite{the_species_pafilis} & NER & 800 & MEDLINE abstracts   &  Biomedical \\  
   
15 &  2013  &  i2b2 2012 \cite{Sun2013EvaluatingTR} & NER & 310 & Clinical records  &  Clinical \\
   

17 &  2010  &  LINNAEUS \cite{Gerner2010LINNAEUSAS} & NER & 153 & PubMed
articles  &  Biomedical \\ 
   
18 &  2018  &  EBM-NLP \cite{nye-etal-2018-corpus} & PICO Extraction & 5,000 & MEDLINE abstracts  & Biomedical \\


  \bottomrule
\end{tabular}

\end{table}

\textbf{Scientific Dataset Construction.}
With the rapid growth of the research community, numerous papers are published every day. 
Therefore, collecting and processing existing research papers for downstream tasks or LMs plays a crucial role when working with scientific text. 
Table \ref{tab_scientific_research_data} presents a list of scientific research datasets that we are aware of.

\begin{table}[hb!]

  \caption{
  Existing scientific text datasets. 
  }
  \label{tab_scientific_research_data}
 \resizebox{\textwidth}{!}{%

  \begin{tabular}{l l l c c c}
    \toprule
  \textbf{Year} &  \textbf{Dataset} &  \textbf{Size} & \textbf{Source} & \textbf{Domain} & \textbf{Available Information} \\
    \midrule

 2009  & AAN \cite{radev-etal-2009-acl} & 25K & ACL Anthology &  Computational Linguistics &  Metadata, Citations \\   

2014   & CiteSeer$^X$ \cite{10.1007/978-3-319-06028-6_26} + RefSeer \cite{Huang2015ANP} & 1.0M & CiteSeer$^X$ + DBLP & Multi  &  Metadata, Citations \\

2018   &  CL Scholar \cite{singh2018cl} & 40K  & ACL Anthology  & Computational Linguistics  & Metadata, Full-text \\   

2019   & Bibliometric-Enhanced arXiv \cite{Saier2019BibliometricEnhancedAA} & 1M & arXiv & All domains in arXiv  & Metadata, Citations, Full-text \\

2019   & NLP4NLP \cite{10.3389/frma.2018.00036} & 65K & 34 Conferences and Journals & Speech and NLP  & Metadata, Citations \\

2020   & NLP Scholar \cite{mohammad-2020-nlp} & 45K & ACL Anthology &  Computational Linguistics & Metadata, Citations \\   
    
2020   & S2ORC \cite{lo-etal-2020-s2orc} & 81.1M &  Semantic Scholar & Multi  & \makecell{Metadata, Full-text, \\ Citations, Figures, Tables} \\    

   2022 & D3 \cite{wahle-etal-2022-d3} & 6.3M & DBLP & CS  & Metadata, Citations \\  

 2022  & NLP4NLP+5 \cite{10.3389/frma.2022.863126} & 90K & 34 Conferences and Journals & Speech and NLP  & Metadata, Citations \\ 


  \bottomrule
\end{tabular}

 }
\end{table}

\subsubsection{Scientific Text Application}
\label{sub_sub_app}
The purpose of this group is to focus on high-level tasks related to scientific understanding and scientific text generation. 
We further divide this group into three subcategories as follows.

\textbf{Text Understanding.}
Scientific text understanding is often more challenging than general text understanding because it requires domain knowledge to comprehend specific terms. 
Currently, various tasks are used to test the understanding ability of models on scientific text, including scientific claim verification, natural language inference (NLI), document analysis, and paper evaluation.
For more tasks, we refer readers to Figure~\ref{fig_task_classification}.

\textbf{Text Generation.}
This group emphasizes the ability to automatically generate scientific text.
Tasks in this group include automatic related work generation, automated evidence synthesis, citation text generation, and summarization.

\textbf{Text Understanding and Generation.}
We consider a QA task to belong to this group because it requires both understanding and generation abilities.
From Section \ref{sec_effective_basic_infor}, we observe that only two QA datasets appear in the top 20 popular datasets used to evaluate SciLMs. 
However, these datasets do not appear as frequently when compared to datasets from other tasks. 
Specifically, PubMedQA is used 10 times, BioASQ is used 7 times, while NCBI-disease is used 27 times. 
%
We argue that the QA task, which includes both testing understanding and generation, is an important evaluation criterion for future SciLMs.
Therefore, we briefly summarize important information about existing QA datasets in Table \ref{tab_qa_tasks} (in Appendix \ref{app_sec_tasks_datasets}).

\subsection{Comparison between Scientific Text and Text in Other Domains}
\label{sec_comparison}

Scientific text has many special characteristics compared to texts in other domains. In this section, we will discuss it from three aspects: content, style, and source, to show the main differences between scientific text and text in other domains and how these characteristics help SciLMs achieve superior performances in some scenarios. Note that we only discuss texts, leaving other modalities like figures and tables aside, because it is out of the scope of this paper.
Table~\ref{tab:comparison_between_texts} summarizes our comparison between scientific text and text in other domains.

\begin{table}[htp]
  \small
  \caption{Comparison between scientific text and text in other domains.}
  \label{tab:comparison_between_texts}

  \begin{tabular}{l | p{7cm} | p{5.5cm} }
    \toprule
  \textbf{Aspects} &  \textbf{Scientific Text} & \textbf{Text in Other Domains}  \\
    \midrule



    \multirow{5}{*}{\textbf{Content}} & - \textbf{Vocabulary}: Contains domain specific terminologies & - \textbf{Vocabulary}: Understandable by everyone \\

    & - \textbf{Knowledge}: Advanced knowledge in the scientific domains & - \textbf{Knowledge}: Common sense in the real world \\
    
    & - \textbf{Reasoning}: Many statements require rigorous logical reasoning & - \textbf{Reasoning}: Some statements contain shallow reasoning paths  \\
    
    & - \textbf{Citation}: Required in many cases & - \textbf{Citation}: Voluntarily included \\
    \midrule

    

    \multirow{5}{*}{\textbf{Style}} & - \textbf{Tone}: Mainly for researchers. Formal, objective, faithful & - \textbf{Tone}: Written for everyone. Informal, subjective, sometimes emotional \\
    & - \textbf{Structure}: Well-organized with rich structural information (e.g., title, abstract, sections, keywords, etc.) & - \textbf{Structure}: Casual \\
    
    & - \textbf{Language}: Long texts (e.g., books or articles) & - \textbf{Language}: Short texts (e.g., reviews or tweets) \\
    \midrule


    \multirow{4}{*}{\textbf{Source}} & - \textbf{Amount}: Growing rapidly, large-scale high-quality corpora available. Not sufficient for training  LLMs from scratch & - \textbf{Amount}: Unlimited (Internet crawling) 
    \\
    & - \textbf{Preprocessing}: 
    OCR or PDF parsing to extract texts from papers & - \textbf{Preprocessing}: Careful filtering \\

  \bottomrule
\end{tabular}

\end{table}

\subsubsection{Content}
\hfill

\textbf{Vocabulary.}
According to the definition of scientific text \cite{russell2012academic}, it is a type of written text that contains information discussing concepts, theories, or other series of topics that are based on scientific knowledge like medicine, biology, and chemistry. Therefore, compared to the text in other domains, scientific text usually has a larger vocabulary size including many terminologies. For example, biomedical domain texts contain a considerable number of domain-specific proper nouns (e.g. BRCA1, c.248T>C) and terms (e.g. transcriptional, antimicrobial) \cite{lee2020biobert}, which are understood by most of the biomedical researchers. Some of the vocabularies are new from before. Note that different from the new words from other domains like Social Network Sites (SNS), these new terminologies are usually followed by clear definitions and detailed explanations described in some sections like `Introduction' or can be found in their cited references. Therefore, pre-training on self-contained scientific text is beneficial for new word discovery. 

Based on this characteristic, some LMs \cite{beltagy-etal-2019-scibert,lo-etal-2020-s2orc,lewis-etal-2020-pretrained} that adopted pre-training from scratch can design tokenizers with domain-specific vocabulary and learn better representations for terminologies. Some other works \cite{tai-etal-2020-exbert,yao2021adapt,ma2022searching} chose to add new embeddings for extended vocabulary when performing continuous pre-training. Table \ref{tab:compare_vocabulary_overlapping} shows the overlapping rates between SciLMs and LMs in the general domain. We can see that the vocabulary overlapping rate between SciBERT and BERT is lower than that between SciBERT and PubMedBERT. This low Jaccard Index indicates that many words or terminologies from scientific domains are not included in the vocabulary set used by LMs in the general domain. Moreover, it also reflects that the distributions are different in scientific domains and the general domain, which enhances the necessity of pre-training on scientific text.

\begin{table}[htp]
  \centering
  \small
  \caption{Vocabulary Similarity Matrix (Jaccard Index).}
  \label{tab:compare_vocabulary_overlapping}

\begin{tabular}{lccc}
\toprule
& \textbf{BERT} & \textbf{SciBERT} & \textbf{PubMedBERT} \\
\midrule
\textbf{BERT} & 1.00 & 0.28 & 0.25 \\
\textbf{SciBERT} & 0.28 & 1.00 & 0.49 \\
\textbf{PubMedBERT} & 0.25 & 0.49 & 1.00 \\
\bottomrule
\end{tabular}

\end{table}

\textbf{Knowledge.}
The main purpose of the scientific text is to share and report advanced research findings, theories, knowledge, and analysis to others who are specialized in related fields in a clear, understandable, and logical manner. Researchers usually catch up with the advanced development of science and obtain some remarkable knowledge for specific tasks like drug discovery by reading published papers from journals and conferences. Therefore, to solve these professional and challenging tasks using LMs, we also need to pre-train them with scientific text that contains many advanced technologies and new knowledge. For example, many domain-specific LMs pre-trained on scientific papers \cite{Hebbar_Xie_2021,ahmad2022chemberta,gupta2022matscibert} can be applied to drug discovery and development, molecule synthesis, and materials discovery. When fine-tuning these LMs for downstream tasks in the same domain, it is easier to adapt and they usually perform better than those LMs pre-trained only in the general domain.

\textbf{Reasoning.}
Scientific text is usually complicated compared to general text. As we described in the previous section, one of the core purposes of scientific papers is to propose something new to peers in their fields. Therefore, most of the contents describe research findings with complicated reasoning processes. Also, some scientific papers contain complex formulas, which can be found in some mathematical and chemical papers. However, most of the current scientific LMs do not use such kind of information. Exploiting this information can yield better performances in understanding complex concepts and reasoning \cite{taylor2022galactica}. 

\textbf{Citation.}
In the scientific domain, supporting materials are required to be included in many cases. For example, in a scientific paper, researchers are required to cite related literature to support their statements. Except for providing supporting information, such citations also contain the relationship between the contents and the supporting materials. \citet{yasunaga-etal-2022-linkbert} utilized the information from the citations and introduced a novel training objective, document relation prediction, to improve the language understanding ability of their model.

\subsubsection{Style}
\hfill

\textbf{Tone.}
The audiences of the scientific text are mainly researchers, scholars, and academics who are knowledgeable in the specific field of research, whereas texts in other domains like news are available for everyone. Therefore, the prior consideration in writing scientific texts is whether the texts are formal, objective, and faithful to convey information and support authors' arguments with evidence. On the contrary, texts in other domains like blogs or novels may be informal, subjective, and emotional that emphasize entertaining, persuading, or expressing authors' personal opinions. Therefore, LMs pre-trained in scientific domains can learn to generate fluent and professional scientific texts for many applications including assisting academic writing \cite{10.1093/bib/bbac409}. 
Meanwhile, with the increasing push for open science and public knowledge, many researchers are also trying to promote their work to wider audiences now, including students, journalists, policymakers, and the general public. To achieve this goal, Scientific Text Simplification has become a promising research topic recently, which aims to simplify scientific texts for non-expert readers \cite{10.1093/jamia/ocac149,engelmann2023text}.

\textbf{Structure.}
Scientific papers are well-organized with rich structural information, including title, abstract, content, table, etc. For example, in the field of health sciences, \citet{sollaci2004introduction} pointed out that a very standardized structure is widely adopted in scientific writing, known as introduction, methods, results, and discussion (IMRAD). Especially, some useful knowledge can be extracted from this structural information. For example, titles and abstracts usually provide rich semantics of the whole paper. Therefore, some work \cite{miolo2021electramed,10.1007/s11192-022-04602-4} extracted the abstract or title as the summary of scientific papers for effective pre-training. Also, some work \cite{taylor2022galactica} adopted the keywords as a useful element to filter undesired papers from pre-training corpora. 
It should be noted that texts from some other domains also have structural information. For example, news articles are usually written in a certain way that readers are hooked by reading the first sentences, also known as lead sentences. Therefore, these sentences can be considered as a proxy for the summary of a piece of news. However, these annotations may not always be reliable, and previous work on PLMs in the general domain usually overlooked this information.

\textbf{Language.}
Compared to text in other domains, scientific text is longer and usually contains multiple pages like books and novels. We select several popular pre-training corpora in the general domain and scientific domains and calculate their linguistic statistics. In the following analysis, we compute the statistics of all subsets of the Pile \cite{pile} containing 22 sources from general and scientific domains. Since it is diverse and can be easily accessible, we believe it can serve as a representative set of commonly used pre-training corpora. Note that different tokenizers can produce different statistics results, like different numbers of tokens. Therefore, without loss of generality, we use the \texttt{PunktSentenceTokenizer} and \texttt{NLTKWordTokenizer} from the NLTK \cite{nltk} library to segment documents into sentences based on punctuations and segment sentences into words mainly based on space, respectively. The statistics can be found in Table \ref{tab:stat_corpora}.

Generally, we use crawled texts from the Internet to do pre-training \cite{Radford2018ImprovingLU,devlin-etal-2019-bert,brown2020language}. We notice that sentences from the Internet usually have an average shorter length (25.94 words), which is not enough for pre-training an LM to solve tasks that require long-term consistency ability. Though sentences from GitHub and Ubuntu IRC have many words, we believe it is because there are many punctuations within a sentence, producing more `words' than they actually have. 

Besides, we also calculate the depth of the syntactic tree and the readability of each sentence to show the complexities of scientific text and text in other domains. We adopt spacy \cite{spacy} to compute the syntactic tree depth of sentences and compute the Flesh-Kincaid Grade \cite{kincaid1975derivation} to evaluate their readabilities. With these metrics, we see that the complexity of scientific text is similar to the texts from the Internet, but scientific text is much more complex than texts from other domains like books, TV, news, email, etc, which makes them suitable for improving reasoning ability of SciLMs.

\begin{table}[htb]
  \small
  \caption{Statistics of some commonly used pre-training corpora.}
  \label{tab:stat_corpora}

  \resizebox{\textwidth}{!}{%

  \begin{tabular}{p{0.2cm}c c c c c c}
    \toprule
    & \textbf{Corpus} &   \textbf{Domain} & \textbf{\#Word/Sentence} & \textbf{Syntactic Tree Depth} & \textbf{Readability} & \textbf{\#Word (B)} \\

  \midrule
 \multirow{17}{*}{\rotatebox{90}{General Text}} & Pile-CC & \multirow{5}{*}{Web} & 23.47 & 24.82 &  \hphantom{0}9.90 & 46.79\\
 
 & OpenWebText2 &  & 24.27 & 25.10 &  \hphantom{0}9.85 & 23.73 \\
 & Wikipedia (en) & & 24.87 & 29.69 &  13.05 & 10.23 \\
 & StackExchange & & 38.33 & 32.61 &  \hphantom{0}9.43 & 13.40 \\
 & \textbf{Average (micro)} &  & 25.94 & 26.53 &  10.16 & \textbf{Sum} 94.15 \\
 
  \cmidrule{2-7}
 & Books3 & \multirow{4}{*}{Book} & 18.51 & 19.50 &  \hphantom{0}6.62 & 29.95 \\
 & BookCorpus2 &  & 14.48 & 16.98 &  \hphantom{0}5.44 & \hphantom{0}3.89 \\
 & Gutenberg (PG-19) &  & 19.59 & 19.34 &  \hphantom{0}5.30 & \hphantom{0}1.17 \\
 & \textbf{Average (micro)} & & 18.10 & 19.21 &  \hphantom{0}6.44 & \textbf{Sum} 35.01 \\
  \cmidrule{2-7}
  
 & OpenSubtitles & \multirow{3}{*}{TV} & 10.17 & 10.49 & \hphantom{0}1.64 & \hphantom{0}5.33 \\
 & YoutubeSubtitles & & 17.42 & 39.66 &  27.92 & \hphantom{0}0.87 \\
 & \textbf{Average (micro)} & & 11.19 & 14.58 &  \hphantom{0}5.33 & \textbf{Sum} \hphantom{0}6.20 \\
  \cmidrule{2-7}
  
 & Github & Code & 90.86 & 103.61 &  20.95 & 18.22 \\
 & Ubuntu IRC & Chat & 40.83 & 29.98 &  \hphantom{0}9.47 & \hphantom{0}1.15 \\
 & EuroParl & Multilingual & 26.78 & 22.67 &  11.79 & \hphantom{0}1.87 \\
 & HackerNews & News & 21.24 & 25.24 &  \hphantom{0}9.89 & \hphantom{0}1.19 \\
 & Enron Emails & Email & 21.71 & 22.52 &  \hphantom{0}6.66 & \hphantom{0}0.25 \\
  \midrule
 \multirow{9}{*}{\rotatebox{90}{Scientific Text}} &  arXiv & CS+Math+Physics & 41.67 & 27.44 &  10.06 & 25.25 \\
 & PubMed Abstracts & Biomed & 24.16 & 25.43 &  14.02 & \hphantom{0}6.59 \\
 & PubMed Central & Biomed & 33.42 & 28.76 &  11.81 & 34.47 \\
 & FreeLaw & Law & 20.15 & 19.43 &  \hphantom{0}6.31 & 13.99 \\
 & USPTO Backgrounds & Law & 25.89 & 27.85 &  12.86 & \hphantom{0}7.87 \\
 & DM Mathematics & Math & 19.22 & 12.66 &  \hphantom{0}1.84 & \hphantom{0}6.55 \\
 & PhilPapers & Philosophy & 23.52 & 24.54 &  11.74 & \hphantom{0}0.75 \\
 & NIH ExPorter & Grant & 27.97 & 29.93 &  16.12 & \hphantom{0}0.72 \\
 \cmidrule{2-7}
  & \textbf{Average (micro)} & Scientific & 31.32 & 25.63 &  10.14 & \textbf{Sum} 96.19 \\
  \bottomrule
\end{tabular}

}
\end{table}

Furthermore, scientific papers, especially peer-reviewed papers, are published after several manual discussions and proofreading, leading to extremely high quality compared to texts in other domains like SNS. Most of them merely have grammar issues. Therefore, language models pre-trained with scientific text can also produce high-quality texts. 


\subsubsection{Source}
\hfill

\textbf{Amount.}
In the domain of scientific text, large-scale unsupervised corpora are freely available and the amount is still growing rapidly. The PubMed Abstracts dataset and PMC Full-text articles that \citet{lee2020biobert} used contain 4.5B and 13.5 tokens respectively, whereas the entire English Wikipedia contains only 2.5B tokens. Moreover, the PubMed subset of a high-quality cleaned dataset, the Pile \cite{pile}, contains 50B tokens, which is enough for training a medium-size LM (e.g., 2.7B) following the recommendation from \citet{hoffmann2022training}. 

But it should also be aware that with the development of LLMs, the current amount for pretraining scientific LLMs from scratch may not be enough. Compared to texts in the general domain, with well-designed filtering strategies, texts from the Internet become a dominant resource for pre-training. For example, the filtered CommonCrawl that \citet{brown2020language} used contains 410B tokens, which is much more than the existing corpora in scientific domains. Furthermore, in some scientific fields, collecting enough high-quality data is still a challenge. For example, in nuclear science, \citet{jain2020nukebert} only collected 7k internal reports. After preprocessing, they only obtained 8M words for language modeling pre-training, which is limited. Therefore, how to explore more scientific texts for pre-training and how to pre-train an excellent scientific foundation model with limited scientific texts is a promising direction in the near future. We leave a detailed discussion of this challenge in Section \ref{sec:build_large_scilms}.

\textbf{Preprocessing.}
Except for the biomedical domain, careful preprocessing is usually needed to obtain high-quality scientific texts. As for some old papers published many years ago, we need to perform OCR (Optical Character Recognition) or PDF parsing to extract texts. The most commonly used PDF parsing tool is Grobid \cite{GROBID}, which was also used for preprocessing S2ORC dataset. Some other models (e.g., VILA \cite{10.1162/tacl_a_00466}) and datasets (e.g., PubLayNet \cite{zhong2019publaynet}) were also proposed to support high-accuracy PDF parsing.

\section{Existing LMs for Processing Scientific Text}
\label{sec_exist_lms}

We systematically organize and present 117 SciLMs in Tables~\ref{table_lms_1},~\ref{table_lms_2}, and~\ref{table_lms_3}.
Due to the space constraint, Tables \ref{table_lms_2} and \ref{table_lms_3} are presented in Appendix \ref{app_sec_existing_scilms}.
We categorize surveyed SciLMs into four sections based on their pretraining corpora: Biomedical, Chemical, Multi-domain, and Other Scientific Domains.
Finally, we further analyze and discuss the popularity of various aspects, such as data domain, language, and model size.
%
In addition to exploring SciLMs, we also provide an overview of LMs trained on non-scientific text during the same period. We summarize these models in Appendix~\ref{sec:scilms-non-scientific-domains} to offer a more detailed understanding of the LM landscape.


\begin{table}

  \caption{
  Existing LMs for scientific text. 
  In the \textbf{Training Objective} column, 
  \textit{MLM} denotes Masked Language Modeling,  
  \textit{NSP} denotes Next Sentence Prediction, and   \textit{SOP} denotes Sentence Order Prediction.
  In the \textbf{Type of Pre-training} column, \textit{CP} and \textit{FS} denote Continual Pretraining and From Scratch, respectively. 
 In the \textbf{Domain} column,  \textbf{CS} represents computer science, \textbf{Bio} represents Biomedical domain, \textbf{Chem} represents Chemical domain, and \textbf{Multi} represents multiple domains. 
It is noted that the date information is chosen from the first date the paper appears on the Internet.
  }
  \label{table_lms_1} 
      \resizebox{\textwidth}{!}{%

  \begin{tabular}{l l l c c c c c c }

    \toprule
  \textbf{No.} &  \textbf{Date} & \textbf{Model} & \textbf{Architecture} &  \textbf{Training Objective} & \makecell{\textbf{Type of} \\ \textbf{Pre-training} } & \makecell{\textbf{Model} \\ \textbf{Size}} & \textbf{Domain} &  \textbf{Pre-training Corpus}  \\

    \midrule
 1 &   2019/01 & BioBERT \cite{lee2020biobert} &  BERT & MLM, NSP & CP & 110M & Bio & PubMed \\ \midrule

2  &   2019/02 & BERT-MIMIC \cite{10.1093/jamia/ocz096} &  BERT & MLM, NSP & CP & \makecell{110M \& 340M} & Bio & MIMIC-III  \\ \midrule

3  &   2019/03 & SciBERT \cite{beltagy-etal-2019-scibert}  & BERT & MLM, NSP & CP \& FS & 110M & Multi & Semantic Scholar Corpus \\ \midrule

4  &   2019/04 & BioELMo \cite{jin-etal-2019-probing}  & ELMo & Bi-LM & FS & 93.6M & Bio & PubMed  \\ \midrule

5   &   2019/04 & Clinical BERT (Emily)  \cite{alsentzer-etal-2019-publicly} & BERT & MLM, NSP & CP & 110M & Bio & MIMIC-III \\ \midrule

6    &   2019/04 & ClinicalBERT (Kexin) \cite{Huang2019ClinicalBERTMC}  & BERT & MLM, NSP & CP & 110M & Bio & MIMIC-III  \\ \midrule

7  &   2019/06 & BlueBERT \cite{peng-etal-2019-transfer}  & BERT & MLM, NSP & CP & \makecell{110M \& 340M} & Bio & PubMed + MIMIC-III \\  \midrule

8 &    2019/06 & G-BERT \cite{ijcai2019p0825} & \makecell{GNN \\ + BERT} & \makecell{Self-Prediction, \\ Dual-Prediction} & CP & 3M & Bio & MIMIC-III \\ \midrule

9 &    2019/07 & BEHRT \cite{7801}  & BERT & MLM, NSP & CP & N/A & Bio & \makecell{Clinical Practice Research Datalink}  \\ 
\midrule

10 &    2019/08 & BioFLAIR \cite{sharma2019bioflair}  & Flair & Bi-LM & CP & N/A & Bio & PubMed \\ \midrule
    
11 &   2019/09 & EhrBERT \cite{info:doi/10.2196/14830}  & BERT & MLM, NSP & CP & 110M & Bio & Electronic Health Record Notes \\  \midrule

12 &   2019/11 & S2ORC-SciBERT \cite{lo-etal-2020-s2orc}  & BERT & MLM, NSP & FS & 110M & Multi  & S2ORC \\ \midrule

13 &   2019/12 & Clinical XLNet \cite{huang-etal-2020-clinical}  & XLNet & \makecell{Generalized Autoregressive \\ Pretraining}  & CP & 110M & Bio & MIMIC-III \\ \midrule

14 &   2020/02 & SciGPT2 \cite{luu-etal-2021-explaining}  & GPT2 & LM & CP & 124M & CS &  S2ORC \\ \midrule

15 &   2020/03 & NukeBERT \cite{jain2020nukebert}  &  BERT & MLM, NSP  & CP & 110M & Chem & NText \\ \midrule

16 &   2020/04 & GreenBioBERT \cite{poerner-etal-2020-inexpensive}  & BERT & \makecell{CBOW Word2Vec, \\ Word Vector Space Alignment} & CP & 110M & Bio & PubMed + PMC \\  \midrule

17 &   2020/04 & SPECTER \cite{cohan-etal-2020-specter} & BERT & Triple-Loss & CP & 110M  & Multi &  Semantic Scholar Corpus \\ \midrule

18 & 2020/05  & BERT-XML \cite{zhang2020bertxml} & BERT  & MLM, NSP &  FS  & N/A & Bio & Electronic Health Record Notes \\ \midrule

19 &  2020/05 &	Bio-ELECTRA \cite{ozyurt-2020-effectiveness}  & ELECTRA & Replaced Token Prediction & FS & 14M & Bio & PubMed + PMC \\  \midrule

20 &   2020/05 & Med-BERT \cite{Rasmy2020MedBERTPC}  & BERT & \makecell{MLM, \\ Prolonged LOS Prediction}  & FS & 110M & Bio & Cerner Health Facts (Version 2017)  \\ \midrule

21 &   2020/05 & ouBioBERT \cite{2005.07202}  & BERT & MLM, NSP & FS & 110M & Bio & PubMed \\ \midrule

22 & 2020/07  & PubMedBERT \cite{10.1145/3458754} & BERT & \makecell{MLM,  NSP, \\ Whole-Word Masking} & FS   & 110M & Bio & PubMed \\ \midrule

23 & 2020/08  & MC-BERT \cite{zhang2020conceptualized}  &  BERT & MLM, NSP &  CP  & \makecell{110M \& 340M} & Bio & \makecell{Chinese Biomedical Community QA \\ + Chinese Medical Encyclopedia \\ + Chinese Electric Medical Record} \\ \midrule

24 & 2020/09  & BioALBERT \cite{article_naseem}  & ALBERT & MLM, SOP   &  CP  & \makecell{12M \& 18M} & Bio & PubMed + PMC \\ \midrule

25 & 2020/09  & BRLTM \cite{9369833}  & BERT  & MLM &    FS  & N/A & Bio & Private Electronic Health Record \\ \midrule

26 &  2020/10 & BioMegatron \cite{shin-etal-2020-biomegatron} &  Megatron & MLM, NSP &  CP \& FS  & \makecell{345M \& \\  800M \& 1.2B} & Bio & PubMed + PMC \\ \midrule

27  &    2020/10  &    CharacterBERT \cite{el-boukkouri-etal-2020-characterbert} &   BERT + Character-CNN & MLM, NSP & FS & 105M & Bio & MIMIC-III  + PubMed \\ \midrule

28 & 2020/10  & ChemBERTa \cite{chithrananda2020chemberta}  & RoBERTa & MLM &  FS  & 125M & Chem & SMILES from PubCHEM \\ \midrule

29 &  2020/10 & ClinicalTransformer \cite{10.1093/jamia/ocaa189}  & \makecell{BERT \\ ALBERT \\ RoBERTa \\ ELECTRA} & \makecell{MLM, NSP \\ 
MLM, SOP \\
MLM \\
Replaced Token Prediction} & CP   & \makecell{110M \\ 
12M \\
125M \\
110M} & Bio &  MIMIC-III \\ \midrule

30 & 2020/10  & SapBERT \cite{liu-etal-2021-self}  & BERT & Multi-Similarity Loss &  CP  & 110M &  Bio &   UMLS  \\  \midrule

31 & 2020/10  & UmlsBERT \cite{michalopoulos-etal-2021-umlsbert}  & BERT  &  MLM &  CP  & 110M &  Bio & MIMIC-III  \\ \midrule

32  &    2020/11  &   bert-for-radiology \cite{10.1093/bioinformatics/btaa668}  &    BERT & MLM, NSP & CP \& FS &  110M & Bio & Chest Radiograph Reports \\  \midrule

33 &    2020/11 & Bio-LM \cite{lewis-etal-2020-pretrained}  & RoBERTa & MLM & FS & \makecell{125M \& 355M} & Bio & PubMed + PMC + MIMIC-III \\ \midrule

 34 &    2020/11  &    CODER \cite{yuan2021coder}  &   \makecell{PubMedBERT \\ 
mBERT} & Contrastive Learning & CP & \makecell{110M \\ 110M} & Bio & UMLS \\ \midrule

35  &    2020/11  &    exBERT \cite{tai-etal-2020-exbert}   & BERT & MLM, NSP  & CP & N/A & Bio & ClinicalKey + PMC  \\ \midrule

36  &    2020/12  &    BioMedBERT \cite{chakraborty-etal-2020-biomedbert}   &    BERT & MLM, NSP & CP \& FS & 340M & Bio & BREATHE Dataset v1.0 \\  \midrule

37  &    2020/12  &    LBERT \cite{article_warikoo}  &  BERT & MLM, NSP & CP & 110M & Bio & PubMed  \\   \midrule

38  &    2021/03  &    OAG-BERT \cite{10.1145/3534678.3539210}   & BERT & MLM & FS & 110M & Multi & AMiner + PubMed    \\ \midrule

39  &    2021/04  &    CovidBERT \cite{Hebbar_Xie_2021} &    BERT & MLM, NSP & CP & 110M & Bio & Covid-19 Related Corpora \\ \midrule

40  &    2021/04  &    ELECTRAMed \cite{miolo2021electramed}   &    ELECTRA  & Replaced Token Prediction & FS & N/A & Bio & PubMed \\ \midrule

41  &    2021/04  &    KeBioLM  \cite{yuan-etal-2021-improving}   &    PubMedBERT & \makecell{MLM, Entity Detection, \\ Entity Linking} & CP & 110M & Bio & \makecell{Pubmed Docs \\ from PubMedDS} \\

  \bottomrule
\end{tabular}

}

\end{table}

\subsection{Biomedical Domain}

This subsection provides a detailed summary of SciLMs specifically pretrained on biomedical corpora, ranging from widely recognized sources such as PubMed, MIMIC-III, and ClinicalTrials, to COVID-19-related and manually constructed datasets.
Since the release of BERT~\cite{devlin-etal-2019-bert}, we have identified 85 existing SciLMs within the biomedical domain, showcasing a diverse range of model architectures, pretraining objectives, and pretraining strategies.
Over time, the architecture of these SciLMs has evolved from LSTM-based architectures to Transformer-based architectures such as BERT~\cite{devlin-etal-2019-bert}, ALBERT~\cite{lan2019albert}, and RoBERTa~\cite{liu2019roberta}. Moreover, the size of these models has grown remarkably, starting from 12M to an impressive 540B parameters. Interestingly, models such as BERT and its variants with approximately 110M and 340M parameters have become the preferred choices for biomedical research due to their cost-effectiveness and high performance in downstream tasks~\cite{hao2019visualizing}.

\subsubsection{Bidirectional Language Modeling (Bi-LM)}

Bi-LM, a common pretraining objective before the rise of transformers, combines forward and backward LMs to compute word probabilities based on previous and future words~\cite{wang2023pretrained}. Since 2019, a few studies have utilized LSTM-based architectures to pretrain LMs for the biomedical domain. For example, BioELMo~\cite{jin-etal-2019-probing} was pretrained from scratch using ELMo~\cite{peters-etal-2018-deep} architecture, while BioFLAIR~\cite{sharma2019bioflair} and Clinical Flair~\cite{rojas-etal-2022-clinical} were pretrained using FLAIR~\cite{akbik-etal-2019-flair} architecture with a continual pretraining strategy.

\subsubsection{Masked Language Modeling (MLM)}

MLM and NSP are two pretraining objectives introduced in the BERT paper~\cite{devlin-etal-2019-bert}. MLM involves randomly masking tokens of input sequences and predicting the masked tokens with the masked input. On the other hand, NSP aims to predict whether a given sentence is the next sentence.


\textbf{MLM-Based Models - Continual Pretraining.}
Since 2019, BERT architecture has become popular for PLMs in NLP. Several SciLMs have been developed using the BERT architecture, including BioBERT~\cite{lee2020biobert}, BERT-MIMIC~\cite{10.1093/jamia/ocz096}, Clinical BERT (Emily)~\cite{alsentzer-etal-2019-publicly}, ClinicalBERT (Kexin)~\cite{Huang2019ClinicalBERTMC}, BlueBERT~\cite{peng-etal-2019-transfer}, BEHRT~\cite{7801}, EhrBERT~\cite{info:doi/10.2196/14830}, MC-BERT~\cite{zhang2020conceptualized}, exBERT~\cite{tai-etal-2020-exbert}, LBERT~\cite{article_warikoo}, CovidBERT~\cite{Hebbar_Xie_2021}, ChestXRayBERT~\cite{9638337}, KM-BERT~\cite{Kim2022-er}, and ClinicalTransformer~\cite{10.1093/jamia/ocaa189}. MC-BERT and KM-BERT were designed for Chinese and Korean languages, respectively. In addition to the original BERT, some studies have tried eliminating the NSP objective from the pretraining process. This modification is motivated by findings suggesting that the NSP objective could introduce unreliability and potentially hinder performance in downstream tasks~\cite{liu2019roberta, lan2019albert,DBLP:conf/nips/YangDYCSL19,joshi2020spanbert}. Notable studies in this context include UmlsBERT~\cite{michalopoulos-etal-2021-umlsbert}, SINA-BERT~\cite{taghizadeh2021sinabert}, PharmBERT~\cite{10.1093/bib/bbad226}, BIOptimus~\cite{pavlova-makhlouf-2023-bioptimus}, CamemBERT-bio~\cite{touchent2023camembertbio}, and EntityBERT~\cite{lin-etal-2021-entitybert}. SINA-BERT was designed for Persian and uses Whole-Word Masking instead of Subword Masking. EntityBERT proposed a method to tag all entities in the input using XML tags, known as Entity-centric MLM~\cite{lin-etal-2021-entitybert}.

Attempts to utilize compact BERT-based models aim to address environmental concerns, enable real-time processing, offer lighter and faster alternatives, support edge computing, optimize parameter usage, overcome memory and speed cxonstraints, and enhance performance in NLP tasks.
~\cite{sanh2019distilbert, lan2019albert}.
For instance, Lightweight Clinical Transformers~\cite{rohanian2023lightweight} uses DistilBERT~\cite{sanh2019distilbert} architecture to distill knowledge during the pretraining phase, which reduces the size of the BERT model by 40\% while retaining 97\% of its language understanding capabilities. Similarly, BioALBERT~\cite{article_naseem} uses ALBERT~\cite{lan2019albert} architecture, which is BERT-based but with much fewer parameters. BioALBERT has 12M-18M parameters, making it the smallest biomedical model in our survey. Additionally, It uses a self-supervised loss for SOP proposed in ALBERT, which helps maintain inter-sentence coherence.

BERT-based alternatives, which are more advanced than BERT, have also been applied in the biomedical domain. Some works rely on Longformer~\cite{beltagy2020longformer} or BigBird~\cite{zaheer2020big} architectures for handling longer input. 
For instance, Clinical-Longformer~\cite{li2022clinicallongformer}, CPT-Longformer~\cite{Louisa_heyarticle}, and EriBERTa-Longformer~\cite{delaiglesia2023eriberta} utilize the Longformer architecture, whereas Clinical-BigBird~\cite{li2022clinicallongformer} and CPT-BigBird~\cite{Louisa_heyarticle} use the BigBird architecture. Additionally, RadBERT~\cite{yan2022radbert} uses RoBERTa architecture~\cite{liu2019roberta}, which extends BERT with changes to pretraining including dynamic masking and no NSP.

Several specifically tailored architectures have been developed using BERT as the base model. DRAGON~\cite{yasunaga2022deep}, G-BERT~\cite{ijcai2019p0825}, KeBioLM~\cite{yuan-etal-2021-improving}, and CODER~\cite{yuan2021coder} focus on incorporating KGs through methods such as GNNs, KG Linking Tasks, or Contrastive Learning.
Moreover, several LMs have been developed for specific use cases. These models include MOTOR~\cite{lin2023medical} and RAMM~\cite{yuan2023ramm}, using additional Contrastive Learning and Image-Text Matching objectives for pretraining multi-modal LMs. ViHealthBERT~\cite{minh-etal-2022-vihealthbert} incorporates Capitalized Prediction to improve NER for Vietnamese. GreenBioBERT~\cite{poerner-etal-2020-inexpensive} uses CBOW Word2Vec~\cite{mikolov2013efficient} and proposes Word Vector Space Alignment to expand wordpiece vectors of a general-domain PLM. SapBERT~\cite{liu-etal-2021-self} presents Self-Alignment Pretraining to learn to self-align synonymous biomedical entities. KEBLM~\cite{LAI2023104392} incorporates lightweight adapter modules to encode domain knowledge in different locations of a backbone PLM. Finally, BioNART~\cite{asada-miwa-2023-bionart} proposes a non-autoregressive LM that enables fast text generation.


\textbf{MLM-Based Models - Pretraining From Scratch.}
Here, we also have several SciLMs that utilize the original BERT architecture. These include BRLTM~\cite{9369833}, AliBERT~\cite{berhe-etal-2023-alibert}, and Gatortron~\cite{Yang2022-pf}, where AliBERT is for French, while Gatortron has a model size of 8.9B, which surpasses the average model size in this domain by more than 40 times.
On the other hand, various models without the NSP objective include BERT-XML~\cite{zhang2020bertxml}, ouBioBERT~\cite{2005.07202}, UTH-BERT~\cite{aclinical_kawazoe}, PathologyBERT~\cite{santos2022pathologybert}, UCSF-BERT~\cite{sushil2022developing}, PubMedBERT~\cite{10.1145/3458754}, and Bioformer~\cite{fang2023bioformer}, where UTH-BERT is for Japanese, and PubMedBERT uses Whole-Word Masking.

Several other LMs were developed using variations of BERT, such as Bio-LM~\cite{lewis-etal-2020-pretrained}, MedRoBERTa.nl~\cite{Verkijk_Vossen_2021}, bsc-bio-ehr-es~\cite{carrino-etal-2022-pretrained}, DrBERT~\cite{labrak2023drbert}, EriBERTa~\cite{delaiglesia2023eriberta}, and Bio-cli~\cite{carrino2021biomedical}. These models were pretrained on biomedical data using the RoBERTa architecture. MedRoBERTa.nl, DrBERT, and Bio-cli were designed for Dutch, French, and Spanish, respectively. Additionally, Bio-cli uses Whole-Word Masking instead of Subword Masking.
SMedBERT~\cite{zhang2021smedbert} is an LM that was pretrained on Chinese corpora. It incorporates deep structured semantics knowledge from neighboring structures of linked entities, which consists of entity types and relations. 
It utilizes objectives like Masked Neighbor Modeling, Masked Mention Modeling, MLM, and SOP.

In addition, there exist different custom architectures of BERT. For instance, CharacterBERT~\cite{el-boukkouri-etal-2020-characterbert} eliminates the wordpiece~\cite{wu2016google} system and instead utilizes a CharacterCNN~\cite{zhang2015character} module, similar to ELMo's first layer representation, to represent any input token without splitting it into wordpieces. Med-BERT~\cite{Rasmy2020MedBERTPC}, on the other hand, incorporates a domain-specific pretraining task to predict the prolonged length of stay in hospitals (Prolonged LOS). This task helps the model learn more clinical and contextualized features for each input visit sequence and facilitates certain tasks. The visit sequence refers to the order in which visits occur in a patient's EHR data. Meanwhile, ProteinBERT~\cite{10.1093/bioinformatics/btac020}'s pretraining combines language modeling with a novel task of Gene Ontology annotation prediction, enabling it to capture a wide range of protein functions. Lastly, BioLinkBERT~\cite{yasunaga-etal-2022-linkbert} takes advantage of document links to capture knowledge dependencies or connections across multiple documents.

\textbf{MLM-Based Models - Both Pretraining Strategies.}
We also consider SciLMs that have experimented with both pretraining strategies. Some models, such as bert-for-radiology~\cite{10.1093/bioinformatics/btaa668}, BioMedBERT~\cite{chakraborty-etal-2020-biomedbert}, Bioberturk~\cite{PPR:PPR560833}, and TurkRadBERT~\cite{türkmen2023harnessing}, use the original BERT architecture - with Bioberturk and TurkRadBERT designed for the Turkish language. BioMegatron~\cite{shin-etal-2020-biomegatron}, on the other hand, implements a simple and efficient intra-layer model parallel approach that enables training transformer models with billions of parameters.

\subsubsection{Replaced Token Detection (RTP)}

Models such as BioELECTRA~\cite{kanakarajan-etal-2021-bioelectra}, Bio-ELECTRA~\cite{ozyurt-2020-effectiveness}, ELECTRAMed~\cite{miolo2021electramed}, and PubMedELECTRA~\cite{TINN2023100729} utilize ELECTRA~\cite{clark2020electra}'s pretraining objectives to pretrain their models from scratch. ELECTRA is a model that replaces MLM with a more sample-efficient pretraining task called RTP. It trains two transformer models: the generator and the discriminator. The generator replaces tokens in the sequence, and the discriminator tries to identify the tokens replaced by the generator.
BioELECTRA also found that the FS strategy performs better than the CP strategy on most BLURB~\cite{10.1145/3458754} and BLUE~\cite{peng-etal-2019-transfer} benchmark tasks.

\subsubsection{Generation-Based Models}

Several generation-based SciLMs have been developed based on auto-regressive pretraining objectives. SciFive~\cite{phan2021scifive}, ClinicalT5~\cite{lu-etal-2022-clinicalt5}, Clinical-T5~\cite{Clinical_T5_lehman}, BioReader~\cite{frisoni-etal-2022-bioreader}, and ViPubmedT5~\cite{phan2023enriching} are all based on T5~\cite{JMLR:v21:20-074} architecture, which uses the pretraining objective of generating the given sequences in an auto-regressive way, taking the masked sequences as input. All models, except Clinical-T5, were pretrained based on an initial model weight rather than being pretrained from scratch. Also, Clinical-T5 has experimented with both pretraining strategies. ClinicalGPT~\cite{wang2023clinicalgpt} is a BLOOM-based model~\cite{workshop2022bloom} that employs rank-based training for reinforcement learning with human feedback to improve performance further. In another work, BioReader uses a retrieval-enhanced text-to-text LM for biomedical, which augments the input prompt by fetching and assembling relevant scientific literature chunks from a neural database with about 60 million tokens centered on PubMed~\cite{frisoni-etal-2022-bioreader}. ViPubmedT5, on the other hand, was pretrained on Vietnamese corpora.
Moreover, BioBART~\cite{yuan-etal-2022-biobart} utilizes BART~\cite{lewis-etal-2020-bart} architecture, a denoising autoencoder, to pretrain a biomedical text-to-text LM via continual pretraining. On the other hand, BiomedGPT~\cite{zhang2023biomedgpt}, BioGPT~\cite{10.1093/bib/bbac409}, and BioMedLM~\cite{cite_BioMedLM} were pretrained from scratch using GPT~\cite{Radford2018ImprovingLU} architecture, while MedGPT~\cite{kraljevic2021medgpt} uses continual pretraining.
Finally, two biomedical PLMs that use an autoregressive Transformer are Clinical XLNet~\cite{huang-etal-2020-clinical} and Med-PaLM~\cite{singhal2023large}. Clinical XLNet was continuously pretrained using XLNet~\cite{DBLP:conf/nips/YangDYCSL19} architecture, which utilizes bidirectional contexts for masked word prediction. Unlike autoregressive models like GPT, XLNet considers all possible permutations of the input sequence, allowing it to capture dependencies between words in both directions and resulting in a better understanding of the context. Med-PaLM, on the other hand, is the largest LM specialized for the medical domain, with 540 billion parameters. It is an instruction prompt-tuned version of Flan-PaLM~\cite{chung2022scaling}.

\subsection{Chemical Domain}

We now focus on SciLMs specialized in the chemical domain, summarizing those pretrained on corpora like PubChem, Material Science, NIMS Materials Database, ACS publications, or custom datasets.
We also consider SciLMs that were pretrained on domains relevant to Material Science, Nuclear, and Battery, as these domains are part of the chemical domain~\cite{Shetty_2023, Trewartha2022-sl, acharya2023nuclearqa, doi:10.1021/acs.jcim.2c00035}. Ultimately, we have gathered a total of 13 qualified SciLMs.
%
Our survey shows a predominant reliance on the BERT architecture for PLMs in the chemical domain. Among the 13 surveyed models, 11 were based on BERT, featuring parameters ranging from 110M to 355M. This suggests limited diversity in model architecture choices for LM pretraining in the chemical domain.


\textbf{BERT-Based Models.}
NukeBERT~\cite{jain2020nukebert} and ProcessBERT~\cite{KATO2022957} were both pretrained from a BERT checkpoint. In contrast, MaterialsBERT (Shetty)~\cite{Shetty_2023} was pretrained from a PubMedBERT~\cite{10.1145/3458754} checkpoint with Whole-Word Masking instead of Subword Masking, both of which aim to specialize in the chemical domain. Meanwhile, MaterialBERT (Yoshitake)~\cite{yoshitake2022materialbert} utilizes the BERT architecture to pretrain from scratch on chemical corpora.
In addition, various studies have been conducted using different variants of BERT, such as non-NSP BERT or RoBERTa. For example, MatSciBERT~\cite{gupta2022matscibert} and NukeLM~\cite{burke2021nukelm} are non-NSP BERT models, while ChemBERT~\cite{doi:10.1021/acs.jcim.1c00284} and a variant of NukeLM use the RoBERTa architecture to continually pretrain on domain-specific data. On the other hand, MatBERT~\cite{Trewartha2022-sl} and ChemBERTa~\cite{chithrananda2020chemberta} are models based on non-NSP BERT and RoBERTa, respectively, that use the pretraining from scratch strategy. Additionally, ChemBERTa-2~\cite{ahmad2022chemberta}, a variant of ChemBERTa, employs Multi-task Regression as an additional objective. It is also worth mentioning that BatteryBERT~\cite{doi:10.1021/acs.jcim.2c00035}, a non-NSP BERT model, utilizes both pretraining strategies.

\textbf{Generation-Based Models.}
Several architectures have been developed for molecular modeling. One such model is ChemGPT~\cite{frey2022neural}, a large chemical GPT-based model with over one billion parameters. It is used for generative molecular modeling and was pretrained from scratch on datasets consisting of up to ten million unique molecules.

\textbf{Multi-Modal Models.}
GIT-Mol~\cite{liu2023gitmol} is a multi-modal LLM that integrates the \textbf{G}raph, \textbf{I}mage, and \textbf{T}ext information. To perform this task, \citet{liu2023gitmol} proposed a novel architecture called GIT-Former, which can map all modalities into a unified latent space.

\subsection{Multi-domain}

This subsection presents SciLMs pretrained on multi-domain corpora, incorporating text data from diverse domains. Referred to as `Multi-domain' for simplicity, these models leverage abundant data, accommodate various domains, and allow further fine-tuning for specific domains.
For instance, SciBERT~\cite{beltagy-etal-2019-scibert} was pretrained on the full text of 1.14M biomedical and CS papers from the Semantic Scholar corpus~\cite{lo-etal-2020-s2orc}.
%
%
Since the introduction of BERT, we have identified 11 multi-domain SciLMs. The majority use BERT or similar architectures, while two others adopt generation-based architectures. Most models prefer a size of 110M parameters, except for Galactica~\cite{taylor2022galactica}, which ranges from 125M to 120B, making it the largest in this domain.


\textbf{BERT-Based Models.}
Multiple works use the BERT architecture to create a general PLM for multiple domains. For example, SciBERT~\cite{beltagy-etal-2019-scibert} and S2ORC-SciBERT~\cite{lo-etal-2020-s2orc} pretrained their models from scratch with BERT's recipe. Other works like OAG-BERT~\cite{10.1145/3534678.3539210} and ScholarBERT~\cite{hong2023diminishing} eliminate the NSP pretraining objective due to its limited contribution to downstream task performance. AcademicRoBERTa~\cite{yamauchi-etal-2022-japanese} and VarMAE~\cite{hu-etal-2022-varmae} employ the RoBERTa architecture. It is worth mentioning that VarMAE uses the continual pretraining strategy, while OAG-BERT, ScholarBERT, and AcademicRoBERTa started from scratch. AcademicRoBERTa was built for the Japanese language. On the other hand, SciDEBERTa~\cite{jeong2022scideberta} further pretrained DeBERTa~\cite{he2020deberta} with the science technology domain corpus. This transformer-based architecture aims to improve BERT and RoBERTa models with two techniques: a disentangled attention mechanism and an enhanced mask decoder~\cite{he2020deberta}.

\textbf{Specialized Architecture-Based Models.}
Some works customize pretraining objectives. For instance, SPECTER~\cite{cohan-etal-2020-specter}, a new LM initialized with SciBERT~\cite{beltagy-etal-2019-scibert}, adds Triple-loss to learn high-quality document-level representations by incorporating citations. Another example is Patton~\cite{jin-etal-2023-patton}, which employs continual pretraining and a GNN-nested Transformer architecture. Its architecture includes two pretraining objectives: Network-contextualized MLM and Masked Node Prediction, which enable the creation of an LLM to capture inter-document structure information.

\textbf{Generation-Based Models.}
Several efforts have been made to pretrain multi-domain LLMs using generation-based architectures. 
For instance, CSL-T5~\cite{li-etal-2022-csl} was pretrained from scratch using T5 for Chinese, while Galactica~\cite{taylor2022galactica} is a 120B Autoregressive LM designed to store, combine, and reason about scientific knowledge.
The model was pretrained from scratch on a vast scientific corpus of papers, reference materials, knowledge bases, and other sources.

\subsection{Other Scientific Domains}


In addition to the well-explored domains mentioned earlier, our survey provides a comprehensive overview of SciLMs pretrained in less commonly explored domains such as Climate, Computer Science, Cybersecurity, Geoscience, Manufacturing, Math, Protein, Science Education, and Social Science. Across these domains, we observed a prevailing preference for a model size of 110M, with the exception of one LLM containing 7B parameters.


\textbf{BERT-Based Models.}
There are different models based on the BERT architecture, some of which use the original BERT while others use advanced variants like non-NSP BERT and RoBERTa. For example, CySecBERT~\cite{bayer2022cysecbert}, SsciBERT~\cite{10.1007/s11192-022-04602-4}, ManuBERT~\cite{ManuBERT_kumar}, and SciEdBERT~\cite{liu2023context} are continually pretrained SciLMs designed for Cybersecurity, Social Science, Manufacturing, and Science Education, respectively. While CySecBERT and SsciBERT use the original BERT architecture, ManuBERT and SciEdBERT use BERT without the NSP objective. Additionally, ClimateBert~\cite{webersinke2022climatebert}, SecureBERT~\cite{1a3e2a1ca6f94c7589c870bc0b3e1e31}, and MathBERT (Shen)~\cite{shen2023mathbert} take advantage of the RoBERTa architecture to further pretrain on Climate, Cybersecurity, and Math corpora, respectively. Note that ClimateBert is a distilled version of the RoBERTa-base model that follows the same training procedure as DistilBERT.

\textbf{Specialized Architecture-Based Models.}
There exist various BERT-based models specifically designed for specific use cases. MathBERT (Peng)~\cite{peng2021mathbert} is an example of a continual pretrained BERT-based model that utilizes MLM, Context Correspondence Prediction, and Masked Substructure Prediction to learn representations and capture semantic-level structural information of mathematical formulas. On the other hand, ProtST~\cite{xu2023protst} is a framework built upon the BERT architecture for multi-modality learning of protein sequences and biomedical texts. It includes tasks like Masked Protein Modeling, Contrastive Learning, and Multi-modal Masked Prediction to incorporate different granularities of protein property information into a protein LM. ProtST was continuously pretrained on Protein corpora.

\textbf{Generation-Based Models.}
There are several GPT-based models available.
SciGPT2~\cite{luu-etal-2021-explaining} is a model specifically designed for the CS domain, and it was created through a process of continued pretraining of GPT2~\cite{radford2019language} model. Another example is K2~\cite{deng2023k2}, which is an LLaMA model~\cite{touvron2023llama} comprising 7B parameters, and it was further pre-trained on a text corpus specific to Geoscience.

\subsection{Summary and Discussion} \label{sec:scilms-summary-and-discussion}

This subsection discusses the popularity of LLMs specifically designed for processing scientific text. Our analysis takes into account various factors such as domain, language, and model size. We present our findings using statistical data obtained from our survey on SciLMs, which are presented in Table \ref{table_summary_domain}, Table \ref{table_summary_multilingual}, and Figure \ref{fig_summary_size}.


\begin{table}[ht]

\begin{minipage}[b]{0.35\linewidth}
    \caption{Distribution of SciLMs across different languages.  
      }
\begin{tabular}{l l l}
    \toprule
\textbf{Language} &  \textbf{\#} & \textbf{Model Names}\\     \midrule
    Dutch & 1  & MedRoBERTa.nl \\
    Korean & 1 & KM-BERT \\ 
    Persian & 1 & SINA-BERT \\ 
    \midrule 
    Turkish & 2 & \makecell[l]{Bioberturk, \\ TurkRadBERT} \\ 
    \midrule 
    Vietnamese & 2 & \makecell[l]{ViHealthBERT, \\ ViPubmedT5} 
    \\ 
     \midrule 
    French & 3 & \makecell[l]{AliBERT, DrBERT, \\ CamemBERT-bio} \\

    \midrule  
    Japanese & 3 & \makecell[l]{ouBioBERT, \\ UTH-BERT, \\ AcademicRoBERTa}  \\

     \midrule  
    Spanish & 3 & \makecell[l]{Bio-cli, \\ EriBERTa, \\ bsc-bio-ehr-es} \\

     \midrule  
    Chinese &  4 & \makecell[l]{MC-BERT, CSL-T5, \\ SMedBERT, \\ ClinicalGPT} \\ 

    \midrule
    
    English   &  97   & \makecell[l]{The remaining models \\ in Tables \ref{table_lms_1}, \ref{table_lms_2} and \ref{table_lms_3}}  \\
    \bottomrule
    \end{tabular}
    \label{table_summary_multilingual}
\end{minipage}\hfill
\begin{minipage}[b]{0.6\linewidth}
\centering
  \caption{Distribution of SciLMs across different domains. }
  \begin{tabular}{l l}
\toprule
\textbf{Domain} &  \textbf{\#} \\     \midrule
    Biomedical & 85 \\
    Chemical & 13 \\
    Multi-domain & 11 \\
    \midrule
    Cybersecurity and Math & 2 per domain \\
    \midrule
    \makecell[l]{Manufaturing, Computer Science,  \\ Climate, Protein, Social Science, \\ Geoscience, and Science Education} & 1 per domain \\
    \bottomrule
\end{tabular}
\label{table_summary_domain}    
\vspace{0.7em}
\includegraphics[width=0.7\textwidth]{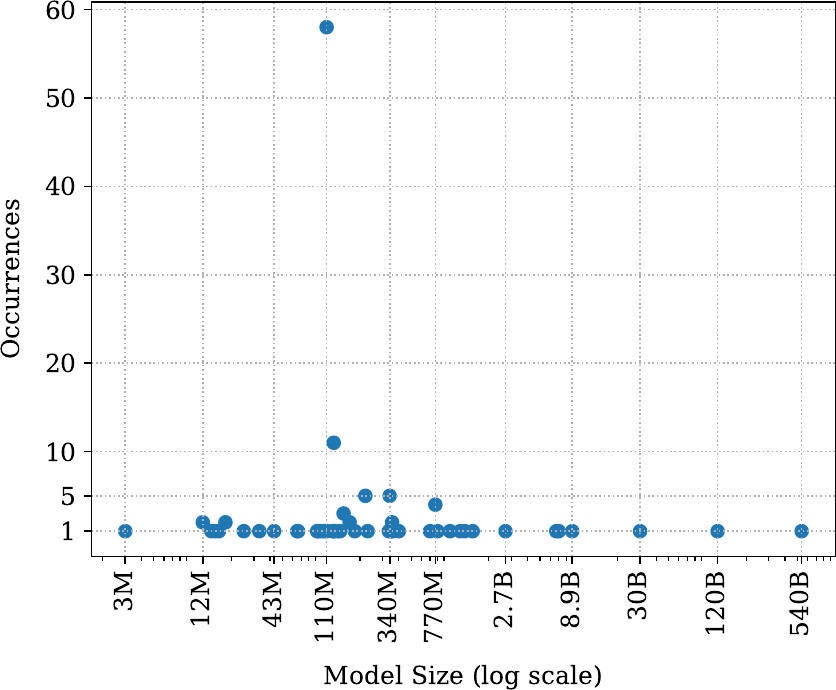}
\captionof{figure}{Distribution of model sizes. \label{fig_summary_size}}

\end{minipage}

\end{table}

\subsubsection{Domain-wise Distribution of SciLMs}

Table \ref{table_summary_domain} provides an overview of SciLM distribution across different domains.
The Biomedical domain has the highest number of existing models, with 85 in total. This dominance is due to the vast amount of scientific literature available in the biomedical field, with PubMed being a prime example, accounting for 17.5\% of The Pile dataset~\cite{pile}. 
The availability of vast and high-quality data within a specific field, such as the biomedical domain, has made it easier to develop domain-specific LLMs that perform well on downstream domain-specific NLP tasks~\cite{wang2023pretrained, sanchez2022effects}. Consequently, researchers have been motivated to create pretrained LLMs for the biomedical domain, leading to the growth of SciLMs in this field. The Chemical domain has 13 existing models, making it the second-highest number of models among all other domains. Interestingly, there is a significant overlap between the Chemical and Biomedical domains, as seen in various Biomedical or Chemical datasets such as BC5CDR, JNLPBA, BC4CHEMD, and others. This overlap presents an opportunity to leverage the vast amount of data available in the Biomedical domain to facilitate the development of more effective LMs for chemistry-related tasks~\cite{munkhdalai2015incorporating}. Besides, there are numerous potential applications for LM development in the Chemical domain, such as autonomous chemical research, drug discovery, materials design, and exploration of chemical space~\cite{boiko2023autonomous, moret2023leveraging, skinnider2021chemical, bran2023chemcrow}. These emphasize the importance of language processing in chemistry-related research. There are also 11 multi-domain models that aim to cater to a broader range of scientific domains. Moreover, pretraining LLMs with mixtures of domains can enhance their ability to generalize to different tasks and datasets~\cite{wang2023data, taylor2022galactica, arumae2020empirical}. By absorbing a wide range of knowledge, these models can gain a better understanding of multiple topics, resulting in better performance and greater versatility for diverse downstream tasks. Other domains like Cybersecurity, Math, Climate, CS, Geoscience, Manufacturing, Protein, Science Education, and Social Science each have one or two models, indicating potential areas for future research and development.

\subsubsection{Language-wise Distribution of SciLMs} 


Table \ref{table_summary_multilingual} shows SciLM prevalence across languages. English dominates with 97 models, underscoring its role as the primary language for scientific communication.
Other languages, such as Chinese, Spanish, Japanese, and French, also have a considerable presence, with multiple models developed for scientific text processing. However, Dutch, Korean, Persian, Turkish, and Vietnamese have fewer dedicated models, indicating that the need for scientific text processing in these languages has only recently attracted attention from the community. This diversification in language usage underlines the global nature of scientific research and underscores the importance of multilingual LMs to cater to researchers from diverse linguistic backgrounds. 
It is worth noting that scientific texts can come in various forms, such as medical records, theses, articles, speeches, textbooks, and books, which are often written in specialized technical language or non-English language for their intended audience.

\subsubsection{Distribution of Model Sizes}



The distribution of model sizes is shown in Figure \ref{fig_summary_size}.
Models within the size range of 100M to less than 400M are the most preferred, with a total of 103 models. Among these, models with sizes of 110M and 125M account for 58 and 11 models, respectively. 
This popularity can be attributed to the balance between efficiency and effectiveness. Many of these models are based on established architectures like BERT-Base and BERT-Large, allowing researchers to leverage prior work while balancing computational cost and performance.
There are 13 models with sizes less than 100M, which are likely favored for their cost-effectiveness and suitability for on-device applications. Researchers usually opt for these models when computational resources are limited, focusing on tasks where lighter models suffice.
There are 6 relatively large models with sizes from 700M to less than 1B. These models offer enhanced capabilities in handling complex scientific language nuances; however, their size presents challenges regarding training cost and computational requirements.
Recent developments have seen a surge in attempts to build large-scale LMs with sizes ranging from 1B to 540B, represented by 11 models. These models represent the cutting edge of language processing techniques, but they also pose significant challenges, including the need for vast corpora, extensive training time, and substantial computational resources. While these models offer remarkable performance, their feasibility and practicality for widespread adoption in the scientific community remain topics of active research and debate.












\section{Effectiveness of LMs for Processing Scientific Text
}
\label{sec_effectiveness}

In this section, 
%
we first present the basic information related to tasks and datasets used by SciLMs, highlighting the top 20 popular tasks and datasets (Section \ref{sec_effective_basic_infor}). We then explore the performance changes over time for five tasks: NER, Classification, RE, QA, and NLI (Section \ref{sec_effective_task_performance}). Additionally, we discuss the detailed information about models that outperform previous ones or achieve the SOTA, such as the number of tasks and datasets they used. Finally, we zoom in a bit closer to the performance when the architecture of the model is fixed (Section \ref{sec_effective_fix_architect}).

\subsection{Basic Information}
\label{sec_effective_basic_infor}

\begin{table}

  \caption{Examples of grouping task names.
  }
  \label{task_group}
  \small

  \begin{tabular}{l c | l c }
    \toprule
  \textbf{Original Name} &  \textbf{Grouped Name}  &  \textbf{Original Name} &  \textbf{Grouped Name}   \\
    \midrule

 Information Retrieval  & \multirow{3}{1cm}{Retrieval} & Dialogue  & \multirow{3}{1cm}{Dialogue} \\

Medical Question Retrieval & & Clinical Dialogue  &   \\

Mathematical Information Retrieval &   &  Medical Conversation  &  \\


 \midrule
Text Generation  & \multirow{3}{1cm}{Generation} & Sentiment Analysis  & \multirow{3}{1cm}{Sentiment Analysis} \\

Formula Headline Generation  & & Medical Sentiment Analysis &  \\

Keyword Generation  & & Sentiment Labeling  &  \\


 \midrule
Question Answering  & \multirow{3}{1cm}{Question Answering} &  Document Multi-label Classification  & \multirow{4}{1cm}{Classification}\\

Visual Question Answering  &  & Text Classification  &  \\


Medical Visual Question Answering  & & Discipline Classification  &  \\



  \bottomrule
\end{tabular}

 \end{table}

Due to the disparities in writing styles and terminologies among scientific papers, many tasks and datasets are labelled with different names, such as `relation extraction' and `relation classification', or the EU-ADR dataset \cite{VANMULLIGEN2012879} can be written as `EU-ADR' (in \citet{lee2020biobert}) or `EUADR' (in \citet{article_naseem}).
%
To ensure consistency in our analysis, we normalize the names of both tasks and datasets. 
If the task names differ but the dataset names are similar, we carefully review the task names to determine whether they should be grouped or kept separate.
After this tedious manual process, we found out that many different task names still look similar. 
Therefore, we rely on heuristics to group task names. 
Specifically, we rely on cue words such as `classification' to categorize name tasks into our list of predefined task names. For instance, if the task name contains `classification' (e.g., 
`sequence classification'), we categorize it as a classification task.
Table \ref{task_group} presents examples of groups of task names produced by our method.
It is noted that this method cannot group all different tasks perfectly, for example, the task `Relationship explanation task' (in \citet{luu-etal-2021-explaining}) is a type of generation task but the task name does not include the word `generation'.
For simplicity, we keep it with its original task name.
All the details of task and dataset names of all SciLMs are presented in Table \ref{tab_result_1} in Appendix \ref{app_sec_details_tasks_datasets}.


\begin{table}[!htb]
    \resizebox{\linewidth}{!}{%
\parbox{.45\linewidth}{
 
  \caption{Top-20 popular tasks and the number of SciLMs evaluated for each respective task.
  \label{table_popular_task} }
  \begin{tabular}{l l l}

\toprule
& \textbf{Task Name} &  \textbf{\#} \\     \midrule
1 & Named Entity Recognition & 58 \\  
 2 &       Classification & 47 \\  
 3 &         Relation Extraction & 33 \\  
 4 &         Question Answering & 31 \\  
5 &         Natural Language Inference & 20 \\  
     6 &    Sentence Similarity & 8 \\  
7 &        Summarization & 8 \\  
 8 &        PICO Extraction & 7 \\  
 9 &        Retrieval & 6 \\  
10 &        Generation & 6 \\  
11 &         Sentiment Analysis & 4 \\  
 12 &         Regression & 4 \\  
13 &         Recommendation & 3 \\  
14 &         Entity Linking & 3 \\  
15 &        Disambiguation & 3 \\  
16 &        Intrinsic Evaluation & 3 \\  
17 &        Dialogue & 3 \\  
18 &        Dependency Parsing & 2 \\  
19 &        Disease Prediction & 2 \\  
20 &        Citation Prediction & 2 \\  
    \bottomrule
\end{tabular}
}
\parbox{.6\linewidth}
{
         \vspace{2em}
    \caption{
Top-20 popular datasets, along with the number of SciLMs evaluated for each respective dataset and information on the task names.
    \label{table_popular_dataset}
      }
\begin{tabular}{l l l l }
    \toprule 
& \textbf{Dataset Name} &  \textbf{\#} & \textbf{Task Names}\\     \midrule 

   1 &     NCBI-disease & 27  & NER (23) or EN (1) or EL (3) \\  
   2 &        BC5CDR-disease & 21 & NER (19) or EL (2)\\  
   3 &        JNLPBA & 21 & NER \\  
   4 &        ChemProt & 21 & RE \\  
   5 &        BC5CDR-chemical & 19 & NER (18) or EL (1) \\  
   6 &        MedNLI & 18 & NLI \\  
   7 &        BC2GM & 16 & NER \\  
   8 &        DDI & 15  & RE \\  
   9 &        i2b2 2010 & 14 & NER (9) or RE (5) \\  
   10 &        HOC & 13 & Document Multi-label Classification \\  
   11 &        GAD & 12 & RE \\  
   12 &        BC4CHEMD & 10 & NER \\  
   13 &        PubMedQA & 10 & QA \\  
   14 &        Species-800 & 8  & NER \\  
   15 &        i2b2 2012 & 8  & NER \\  
   16 &        BC5CDR & 8 & NER \\  
   17 &        LINNAEUS & 7  & NER \\  
   18 &        EBM-NLP & 7 & PICO Extraction \\  
   19 &        BIOSSES & 7  & Sentence Similarity \\  
   20 &        BioASQ & 7 & QA \\  
        
    \bottomrule
    \end{tabular}
}

}

\end{table}

In summary, after grouping, there are 79 tasks and 337 datasets used to evaluate 117 SciLMs in our survey. 
It is worth noting that these 79 tasks could be further grouped if we carefully examine the details. 
However, we find that this process is not necessary, so we choose to skip it.
Tables \ref{table_popular_task} and \ref{table_popular_dataset} present the top-20 most popular tasks and datasets, and the number of SciLMs evaluated for each respective task and dataset. 
We observe that NER, Classification, RE, QA, and NLI emerge as the top five most popular tasks. 
For specifics, the top five datasets for the NER task are NCBI-disease, BC5CDR-disease, JNLPBA, BC5CDR-chemical, and BC2GM. Regarding the RE task, ChemProt and DDI stand out as the top two datasets. 
MedNLI claims the top spot for the NLI task, while HOC leads as the most popular dataset for the Document Multi-label Classification task, which is grouped under the classification task. 
For the QA task, PubMedQA and BioASQ are recognized as the two most popular datasets, although fewer SciLMs have been evaluated on these compared to other datasets.
In the next subsection, we delve deeper into exploring the task performance of the SciLMs on these popular tasks and datasets.

\subsection{Exploring Task Performance}
\label{sec_effective_task_performance}

In this section, we first present charts for the five most popular tasks to visualize how SciLM performance changes over time. 
We then analyze the list of SciLMs that outperform previous models or achieve SOTA results.

\begin{figure}[h]
  \centering
  \includegraphics[width=\linewidth]{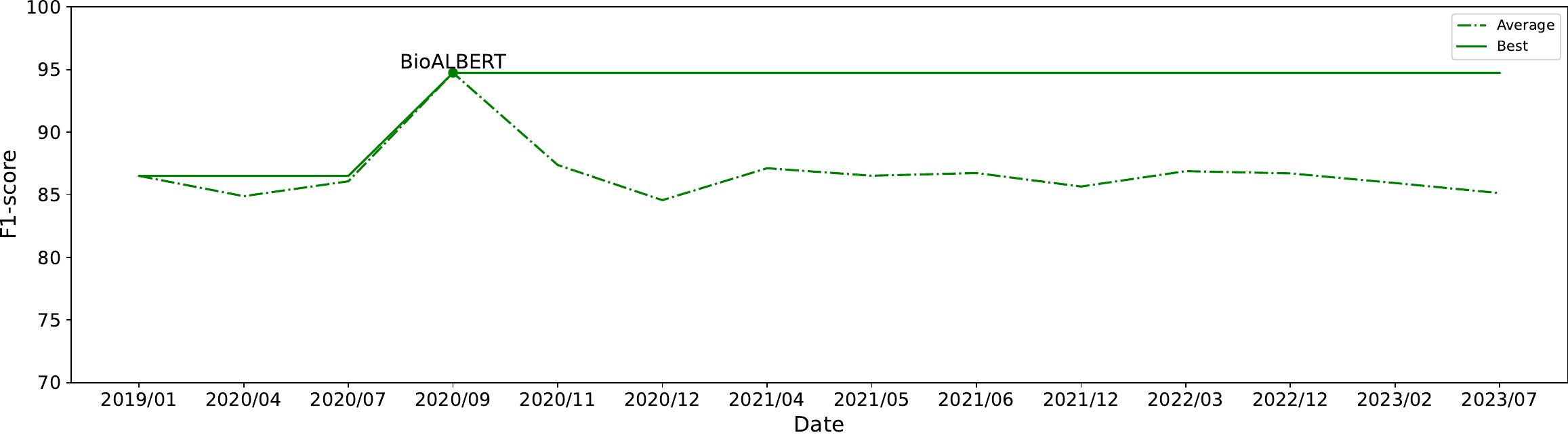}
  \caption{Average performance changes in the NER task. These scores are the average from the five NER datasets: NCBI-disease, BC5CDR-disease, JNLPBA, BC5CDR-chemical, and BC2GM.
  }
\label{result_ner}
\end{figure}

\subsubsection{Performance Changes Over Time}
\label{sec_5tasks_effectiveness}

\hfill

\textbf{NER Task.}
As shown in Tables \ref{table_popular_task} and \ref{table_popular_dataset}, 
NER is the most popular task used to evaluate SciLMs. 
There are eleven popular NER datasets among the top-20 datasets used by SciLMs to evaluate their performance, and we use five of them for drawing charts.
From the Table \ref{tab_result_1} (in Appendix \ref{app_sec_details_tasks_datasets}), we obtain a list of SciLMs that were evaluated on these five NER datasets. 
Subsequently, we assess the performance of the following SciLMs: BioBERT, GreenBioBERT, PubMedBERT, BioALBERT, Bio-LM, BioMedBERT, KeBioLM, SciFive, BioELECTRA, PubMedELECTRA, BioLinkBERT, BioReader, Bioformer, and BIOptimus.
%
For clarity, we display the average performance (F1-score) changes of all five datasets in Figure \ref{result_ner}. 
Performance changes for each of the five datasets are detailed in Figure \ref{result_ner_details} (in Appendix \ref{app_sec_exploring_task_performance}).

We note a consistent pattern in the average performance changes across the five datasets.
In September 2020, BioALBERT achieved the highest score with an average F1-score of 94.7.\footnote{We searched for papers discussing the highest scores achieved by BioALBERT, but we couldn't find any. Additionally, BioALBERT has released its model weights on the \href{https://github.com/usmaann/BioALBERT}{GitHub repository}. Therefore, we defer the in-depth analysis of fair comparisons and reliable results for BioALBERT to future studies.}
To date, the performance of BioALBERT on these NER datasets remains unmatched by any of the proposed models.
Our observations raise two main research questions related to the NER task and dataset. 
The first question is: 
\emph{Does it imply that the NER task is already solved by current SciLMs?}
We believe that the answer is no. 
Based on the detailed results in Figure \ref{result_ner_details} (in Appendix \ref{app_sec_exploring_task_performance}), we observe that the F1-score of BioALBERT on the JNLPBA dataset is only 84.0. 
This suggests that the high scores on other datasets may be due to overfitting between the training data of BioALBERT and these NER datasets. 
Additionally, it would be interesting to evaluate the models further, such as assessing their performance on adversarial sets.
The second question is: 
\emph{Why are newly proposed models unable to surpass the performance of BioALBERT?}
%
We believe there are several reasons for this, and here we discuss two that we consider are most important:
(1) The architecture of later SciLMs differs; they may experiment with alternative architectures, such as using T5 for SciFive, ELECTRA for BioELECTRA and PubMedELECTRA.
(2) The research focus varies; subsequent studies may explore additional methods for solving the task rather than solely aiming for the highest performance. 
For instance, \citet{yasunaga-etal-2022-linkbert} proposed new types of LMs by incorporating link information between documents into the training dataset and loss. Additionally, \citet{pavlova-makhlouf-2023-bioptimus} introduces a SciLM by pre-training with a curriculum learning schedule.

%





\begin{figure}[h]
    \begin{subfigure}{0.48\linewidth}
        \includegraphics[width=\linewidth]{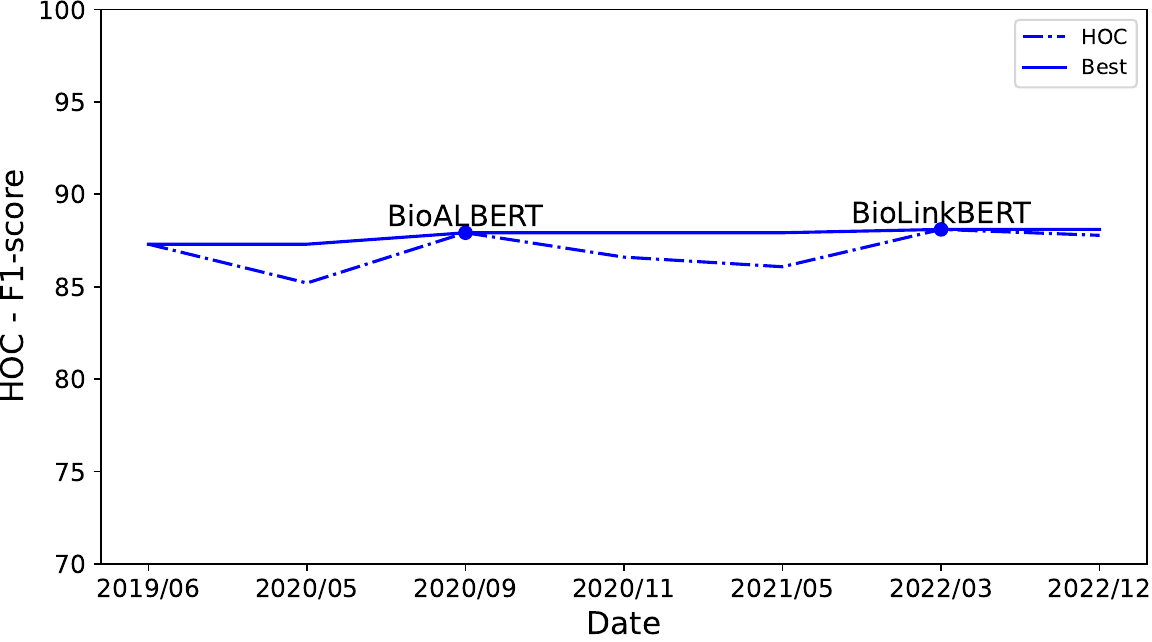}
    \end{subfigure}
    \begin{subfigure}{0.48\linewidth}
        \includegraphics[width=\linewidth]{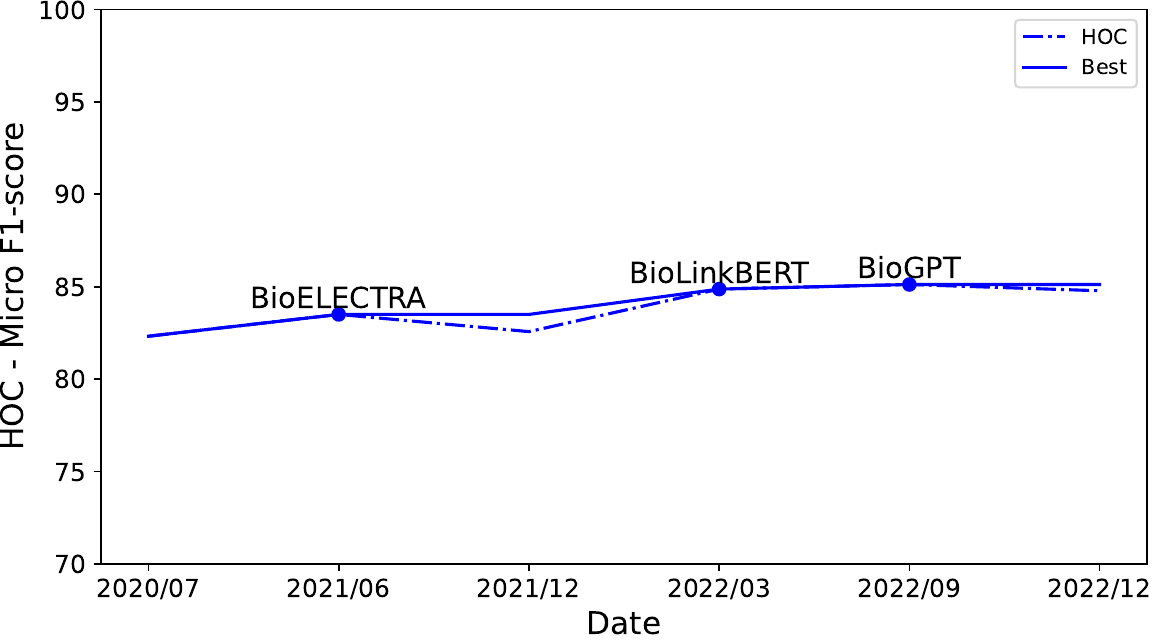}
    \end{subfigure}

    \caption{Performance changes in the HOC dataset. Left: measure by \textbf{F1-score}; Right: measure by \textbf{Micro F1-score}.}
    \label{result_hoc}
\end{figure}

\textbf{Classification Task.}
As shown in Table \ref{table_popular_task}, the classification task is the second most popular task.
However, there are many different types of classification tasks, such as citation intent classification (e.g., ACL-ARC \cite{jurgens-etal-2018-measuring}) or formula topic classification (e.g., TopicMath-100K \cite{peng2021mathbert}).
%
This explains why only one classification dataset, namely HOC, appears in the top 20 most popular datasets used to evaluate SciLMs. 
HOC denotes `Hallmarks of Cancer'; the HOC dataset consists of 1,499 cancer-related PubMed abstracts that have been annotated by experts. It includes 10 classes, each corresponding to one of the hallmarks of cancer.
This is a multi-label classification task, and we note that the F1-score and micro F1-score are commonly used for comparison. We observe that only BioLinkBERT obtains both scores. Therefore, we create two charts for two lists of SciLMs.
The performance changes of the HOC dataset are presented in Figure \ref{result_hoc}. 
On the left side, they are evaluated using F1-score, with the list of SciLMs as follows: BlueBERT, ouBioBERT, BioALBERT, Bio-LM, SciFive, BioLinkBERT, and BioReader. 
On the right side, they are evaluated using micro F1-score, with the list of SciLMs as follows: PubMedBERT, BioELECTRA, PubMedELECTRA, BioLinkBERT, BioGPT, and clinicalT5.
We believe that it is hard to draw any reliable conclusion when some models show scores on one metric, and others show scores on another type of metric.

%

\begin{figure}[h]
  \centering
  \includegraphics[width=\linewidth]{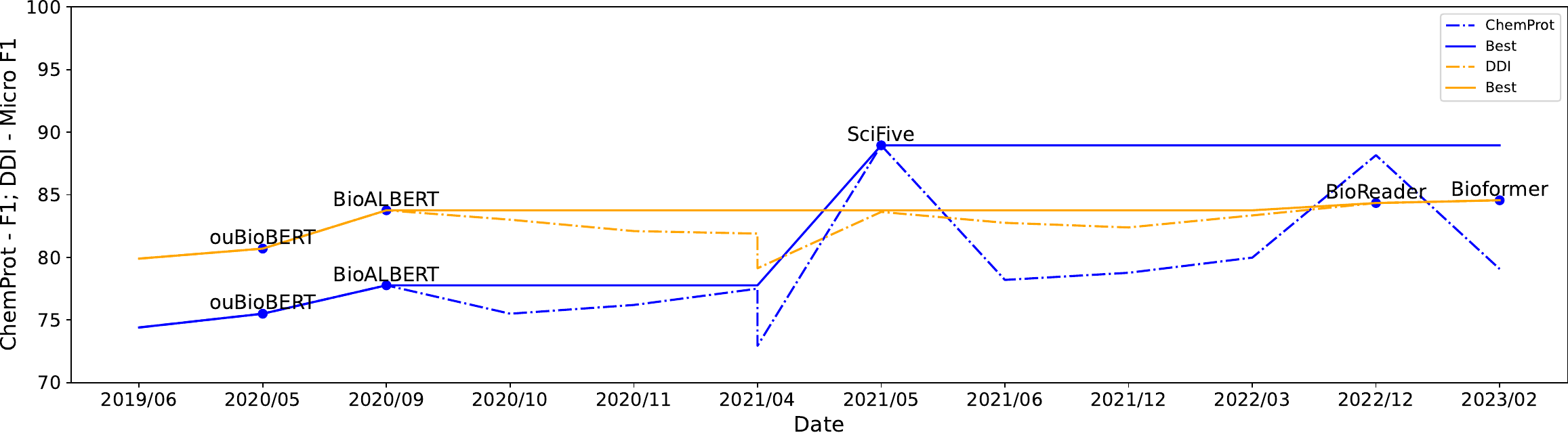}
  \caption{Performance changes in the RE task. ChemProt is evaluated with F1-score, while DDI uses Micro F1-score.
  }
\label{result_re}
\end{figure}

\textbf{RE Task.}
The third popular task is RE. 
Three popular RE datasets (ChemProt, DDI, and GAD) appear in the top 20 datasets. 
However, after merging the list of models evaluated on these three datasets, the number is quite small, with only 9 remaining models. 
Therefore, we only draw a chart for the first two datasets, ChemProt and DDI, with the following list of SciLMs: BlueBERT, ouBioBERT, BioALBERT, CharacterBERT, Bio-LM, KeBioLM, ELECTRAMed, SciFive, BioELECTRA, PubMedELECTRA, BioLinkBERT, BioReader, and Bioformer.\footnote{LBERT is excluded due to scores being on a different scale from other SciLMs. PubMedBERT is omitted as its F1 score for ChemProt is unavailable.}
Figure \ref{result_re} presents the performance changes of these two datasets. 
Similar to the NER task, BioALBERT achieved SOTA results in September 2020. 
However, for the RE task, there are proposed models that have surpassed the score of BioALBERT. 
Specifically, on the ChemProt dataset, SciFive achieved SOTA in May 2021 and still holds the SOTA title. 
In the case of the DDI dataset, BioReader achieved SOTA in December 2022, and Bioformer achieved SOTA in February 2023.

\begin{figure}[h]
  \centering
  \includegraphics[width=\linewidth]{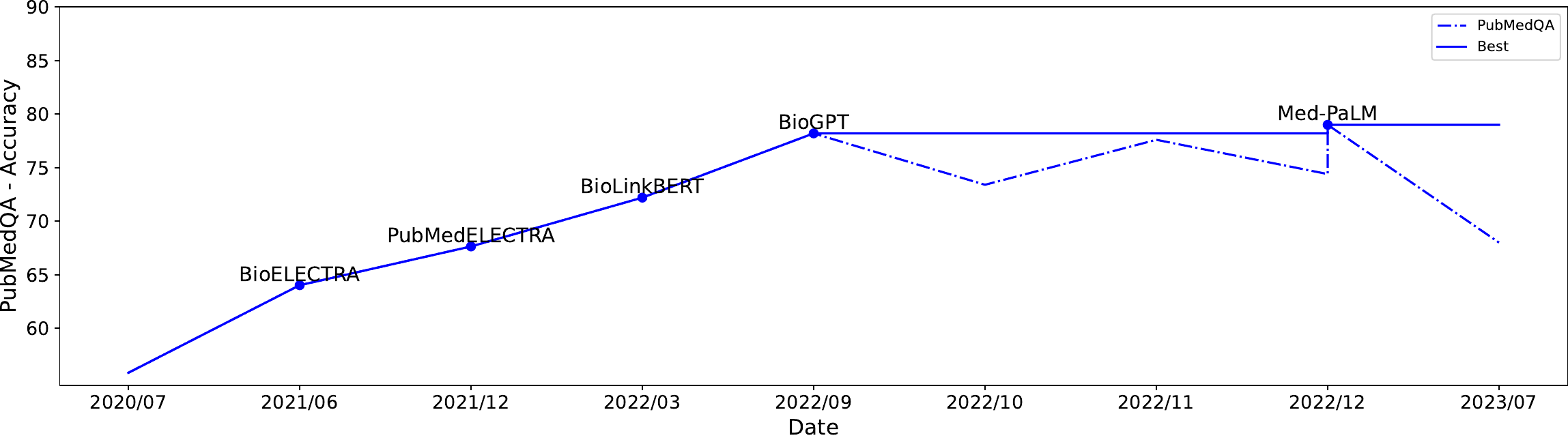}
  \caption{Performance changes in PubMedQA. 
  The range is from \textbf{60 to 90}, which differs from other tasks due to space constraints.
  }
\label{result_qa}
\end{figure}

\textbf{QA Task.}
There are only two QA datasets (PubMedQA and BioASQ) in the top 20 popular datasets. 
However, these two datasets are not as common as others, such as MedNLI or HOC. 
Few SciLMs are evaluated on the BioASQ dataset; therefore, we have decided to only draw a chart for the PubMedQA dataset.
Figure \ref{result_qa} presents the performance changes in PubMedQA. 
We observe that the performance of SciLMs on PubMedQA shows a gradual improvement over time. 
BioELECTRA surpasses the score of PubMedBERT (55.8) and achieves better performance (64.0). 
Subsequently, PubMedELECTRA surpasses the score of BioELECTRA, demonstrating even better performance (67.6). 
By utilizing citation link information in training the SciLM, BioLinkBERT outperforms all previous models and obtains the new best score (72.2) in March 2022.
However, most of these models lack the ability to generate. 
By using the new generation model, GPT, \citet{10.1093/bib/bbac409} proposed BioGPT and achieved a new SOTA result in the PubMedQA dataset, reaching 78.2. 
Afterward, several proposed SciLMs were introduced, but their scores are still lower than the score of BioGPT.
Recently, \citet{singhal2023large} introduced Med-PaLM by performing instruction prompt tuning on the Flan-PaLM model.
However, the improvement here is smaller than the improvement from BioLinkBERT to BioGPT.

\begin{figure}[h]
  \centering
  \includegraphics[width=\linewidth]{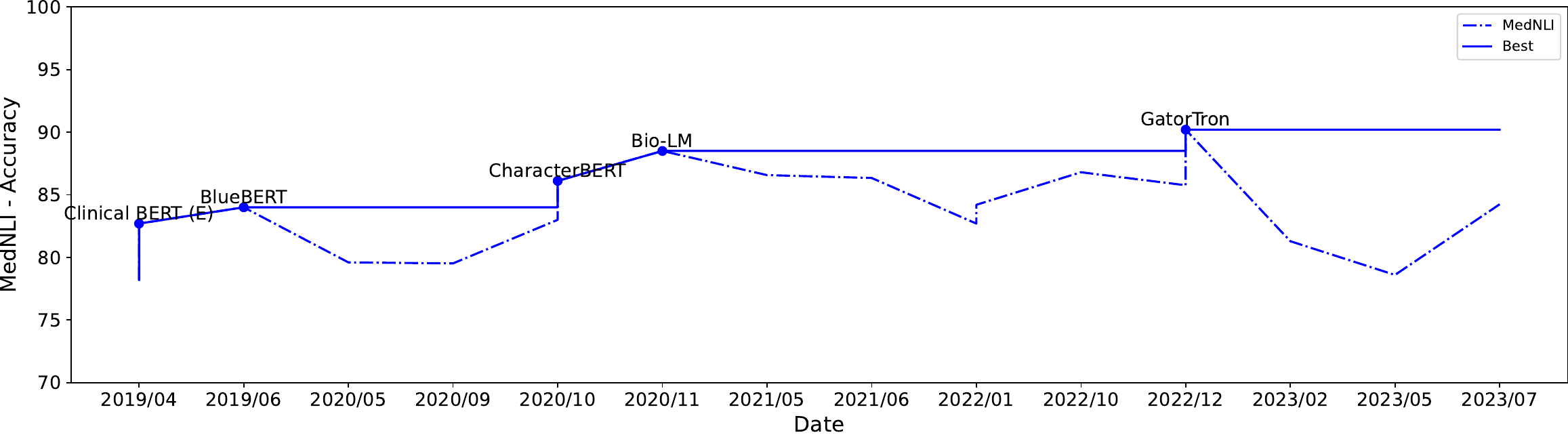}
  \caption{Performance changes in the MedNLI dataset.
  }
\label{result_nli}
\end{figure}

\textbf{NLI Task.}
The last popular task is NLI. 
MedNLI was the sole dataset for the NLI task focused on processing scientific text in English until recently when \citet{jullien-etal-2023-nli4ct} introduced the NLI4CT dataset.
Additionally, there are other NLI datasets for different languages, such as ViMedNLI \cite{phan2023enriching} for Vietnamese. 
MedNLI has been evaluated by many models from April 2019 until the present.
Figure \ref{result_nli} presents the performance changes in the MedNLI dataset.
As shown in the figure, in April 2019, Clinical BERT (Emily) achieved the SOTA score of 82.7 on the MedNLI dataset. 
Subsequently, in June 2019, BlueBERT surpassed the score of Clinical BERT (Emily) and achieved a new SOTA of 84.0. 
CharacterBERT surpassed the scores of BlueBERT and achieved SOTA in November 2020 (86.1). 
One month later, Bio-LM established a new SOTA with a score of 88.5. 
Recently, GatorTron surpassed Bio-LM, achieving a new SOTA with a score of 90.2.

\subsubsection{Number of Models Outperform Previous Models or Achieve SOTA Results}
\label{number_of_task_dataset_sota}

\begin{figure}[h]
    \centering
    \begin{subfigure}{0.4\linewidth}
        \includegraphics[width=\linewidth]{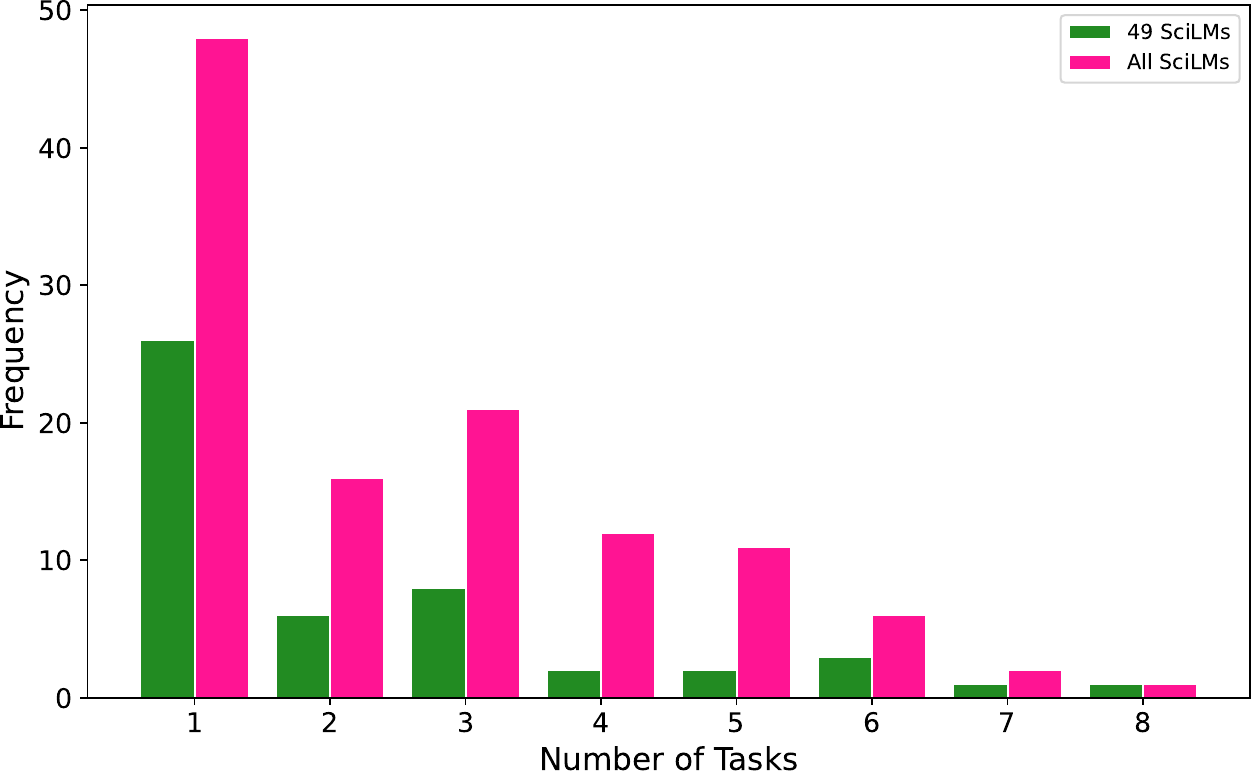}
    \end{subfigure}
    \begin{subfigure}{0.54\linewidth}
        \includegraphics[width=\linewidth]{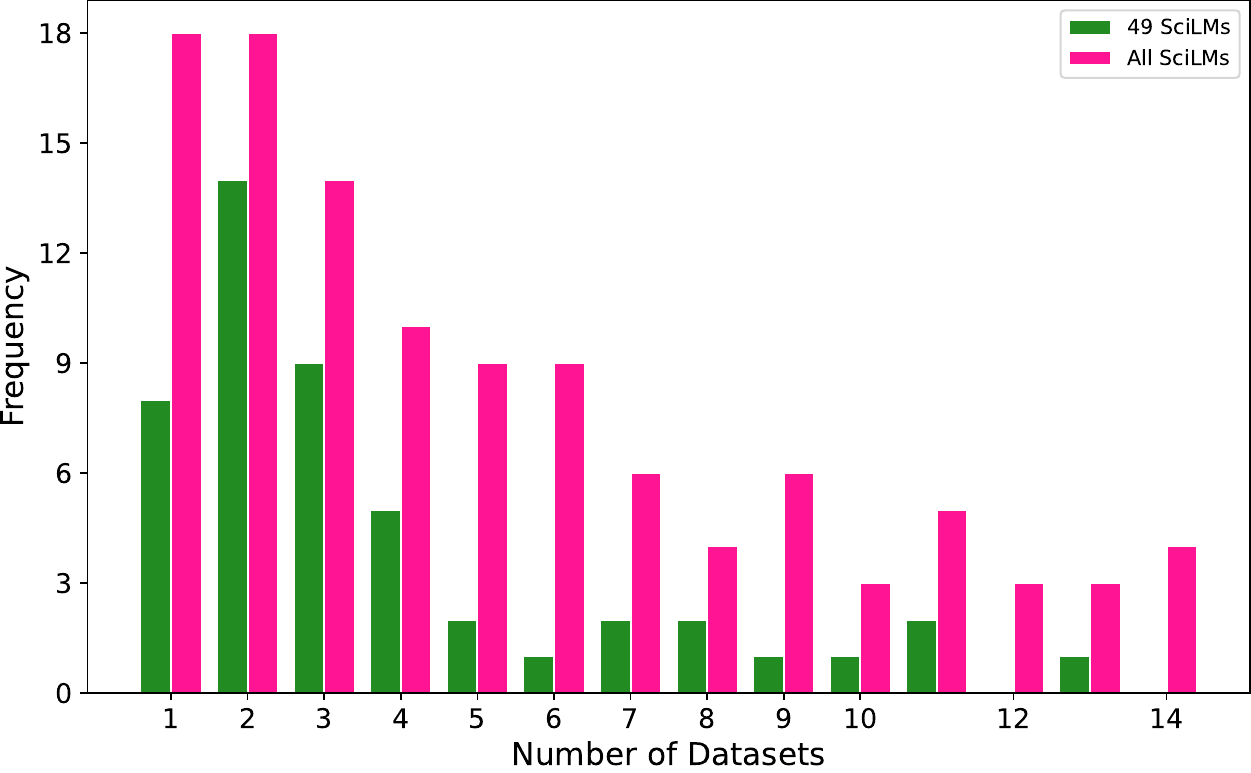}
    \end{subfigure}

    \caption{Histogram of number of tasks and datasets of all SciLMs and 49 SciLMs that outperform previous models.}
    \label{histogram_task_dataset}
\end{figure}

It is difficult and time-consuming to precisely obtain the number of SOTA SciLMs on all different datasets separately. 
In each proposed SciLM paper, the authors often mention whether their model achieves SOTA results or outperforms previous models. 
Therefore, we utilize this information for analysis. 
If the SciLM outperforms previous models or achieves SOTA results, we add the highlight word \hl{(All)} at the end of the column `Datasets' for each model  in Table \ref{tab_result_1} in Appendix \ref{app_sec_details_tasks_datasets}.
In summary, there are 49 SciLMs that outperform previous models or achieve SOTA results in the list of 117 SciLMs. 
However, this number alone does not provide a comprehensive understanding of the effectiveness of SciLMs. 
In some cases, SciLMs only evaluate on one dataset or one task, making it unfair for comparison. 
To effectively represent this number, we conducted a statistical analysis to count how many datasets and tasks these 49 SciLMs used in their evaluations. Figure \ref{histogram_task_dataset} displays histograms for the number of tasks (Left) and datasets (Right) among these 49 SciLMs (green columns).
As depicted in the figure, many SciLMs conducted their evaluation on only one task (26 out of 49 SciLMs) or on only a few datasets (8 out of 49 on one dataset, 14 out of 49 on two datasets). 
These numbers raise two main issues: (1) the generalization ability of proposed models remain unclear and (2) comparing the performance of proposed models may not be meaningful. 
For the first issue, if the model is only evaluated on one task, it indicates that the abilities of the model on other tasks have not been fully evaluated yet. 
Regarding the second issue, if the comparison is performed on two or three datasets, and it's quite domain-specific, then even if the model achieves SOTA scores or outperforms previous models, it's challenging to draw any reliable conclusions in this case.

Motivated by the above observations and to get a better understanding of the gravity of the identified issues, we have also created histograms for both the number of tasks and datasets in the complete list of 117 SciLMs.
These histograms are the pink columns in Figure \ref{histogram_task_dataset}.
As depicted in the figure, many SciLMs evaluated their models on a limited number of tasks and datasets. 
This raises concerns about the reliability of the evaluation conducted on these SciLMs, which is discussed in Section \ref{sec_evaluating_issue}.

\subsection{Variations in BERT-based Models Performance Across Tasks}
\label{sec_effective_fix_architect}

%
In this section, we extensively explore variations in performance over time with fixed model architectures.

\begin{figure}[h!]
    \centering
    \begin{subfigure}{0.495\linewidth}
        \includegraphics[width=\linewidth]{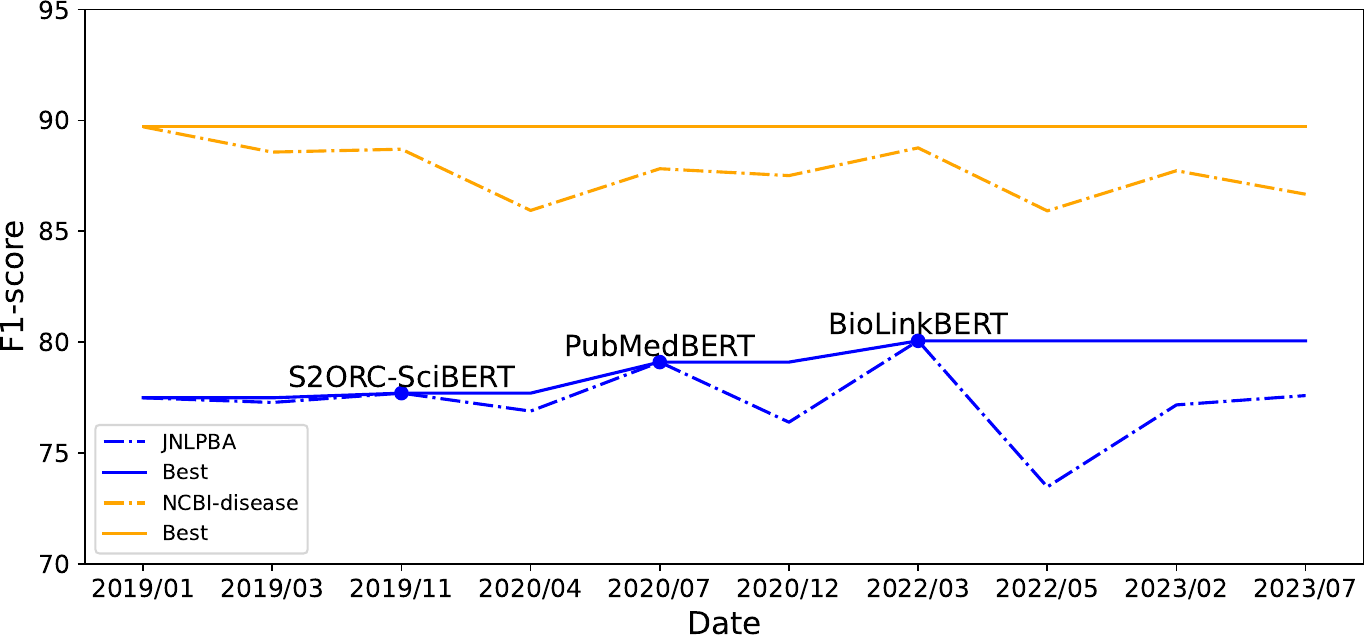}
    \end{subfigure}
    \begin{subfigure}{0.495\linewidth}
        \includegraphics[width=\linewidth]{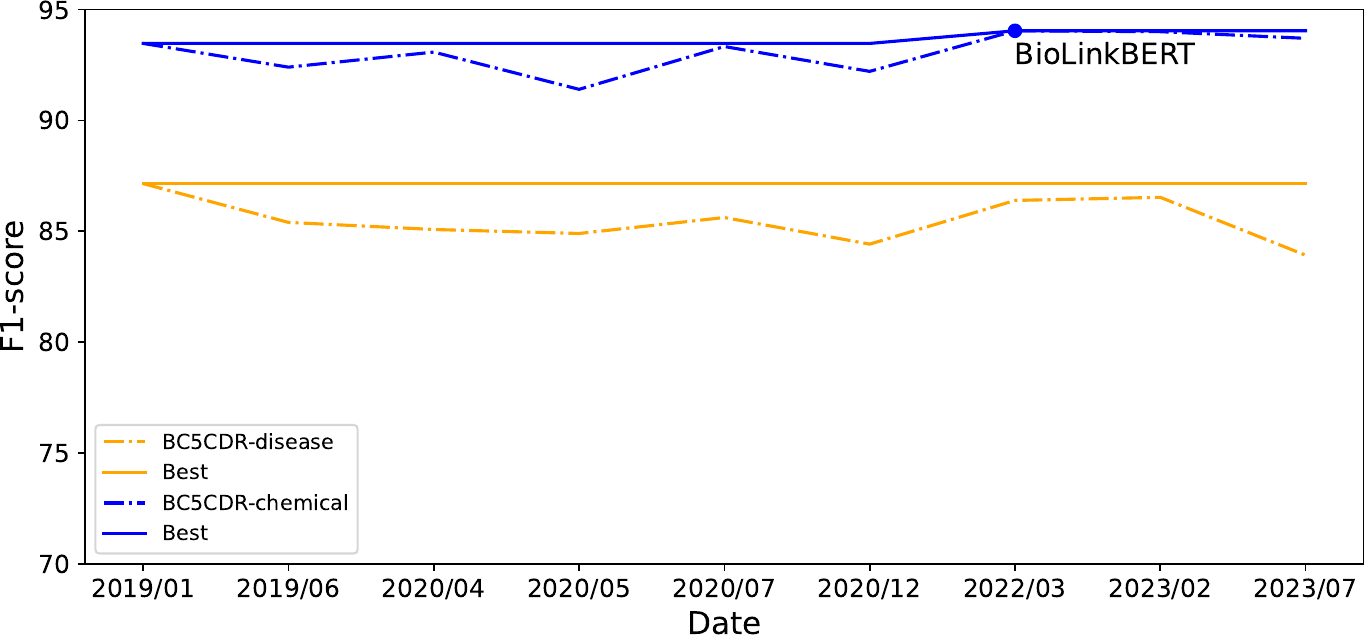}
    \end{subfigure}

    \caption{Details of performance changes for four NER datasets, JNLPBA, NCBI-disease, BC5CDR-disease, and BC5CDR-chemical when we fix the architecture information of the models.}
    \label{result_ner_details_bert_archtec}
\end{figure}

\textbf{NER Task.}
From the list of SciLMs in Section \ref{sec_5tasks_effectiveness}, we select the subset of models utilizing the BERT architecture.
%
We find that only seven of these models evaluate their performance on BC2GM.
Therefore, in this part, we only draw charts for four NER datasets: JNLPBA, NCBI-disease, BC5CDR-disease, and BC5CDR-chemical.
The list of SciLMs for the JNLPBA and NCBI-disease datasets is: BioBERT, SciBERT, S2ORC-SciBERT, GreenBioBERT, PubMedBERT, BioMedBERT, BioLinkBERT, ScholarBERT, Bioformer, and BIOptimus.
The list of SciLMs for the BC5CDR-disease and BC5CDR-chemical datasets is: BioBERT, BlueBERT, GreenBioBERT, ouBioBERT, PubMedBERT, BioMedBERT, BioLinkBERT, Bioformer, and BIOptimus.
Figure \ref{result_ner_details_bert_archtec} details performance changes for JNLPBA, NCBI-disease, BC5CDR-disease, and BC5CDR-chemical NER datasets with fixed model architectures.
For the NCBI-disease and BC5CDR-disease datasets (orange lines), we observe that no proposed models have surpassed the performance of BioBERT, introduced in January 2019.
For the BC5CDR-chemical dataset, BioLinkBERT, proposed in March 2022, can improve the performance of BioBERT, but the improvement is small.
For the JNLPBA dataset, we observe a slow but steady progress in performance changes over time. 
Specifically, S2ORC-SciBERT, proposed in November 2019, slightly improves the performance of BioBERT (77.5 to 77.7). 
Following that, PubMedBERT, introduced in July 2020, outperforms S2ORC-SciBERT by a large margin, increasing from 77.7 to 79.1. 
In March 2022, BioLinkBERT outperforms all previous BERT-based models and obtains the best score until now with an 80.1 F1-score when compared with BERT-based models only.

\textbf{Classification Task.}
We observe that there are only three BERT-based models (BlueBERT, ouBioBERT, and BioLinkBERT) using F1 score, while there are two BERT-based models (PubMedBERT and BioLinkBERT) using micro F1 score for evaluation.
Therefore, we have decided not to draw charts for the HOC dataset.
In terms of model performance, there is an improvement from BlueBERT (87.3 F1-score) to BioLinkBERT (88.1 F1-score).

\textbf{QA Task.}
Similar to the classification task, there are also only three BERT-based models (PubMedBERT, BioLinkBERT, and KEBLM) for the PubMedQA dataset. Therefore, we do not draw charts for it.
We observe an improvement from PubMedBERT (55.8\% accuracy) to BioLinkBERT (72.2\% accuracy), but KEBLM shows a performance drop with only 68.0\% accuracy. 
The main reason may be that KEBLM focuses on proposing models that can incorporate information from multiple types of knowledge, instead of relying solely on unstructured text.

\begin{figure}[bh]
    \centering
    \begin{subfigure}{0.495\linewidth}
        \includegraphics[width=\linewidth]{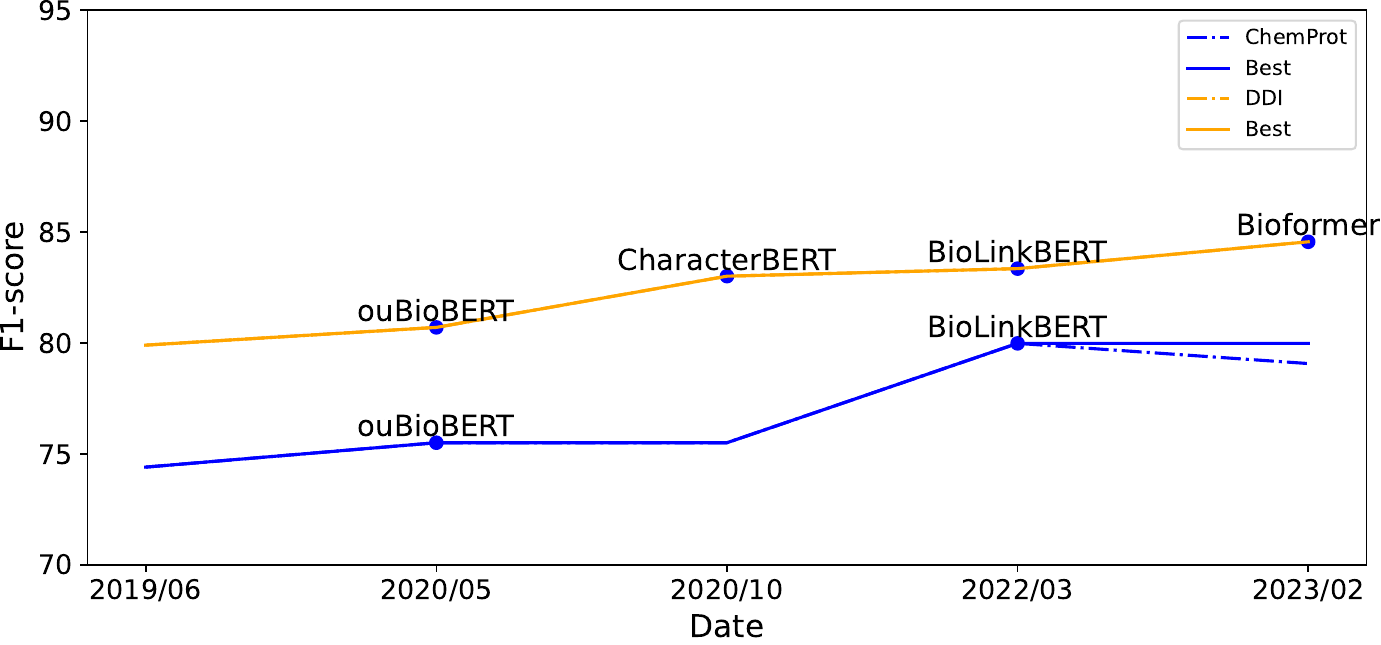}
    \end{subfigure}
    \begin{subfigure}{0.495\linewidth}
        \includegraphics[width=\linewidth]{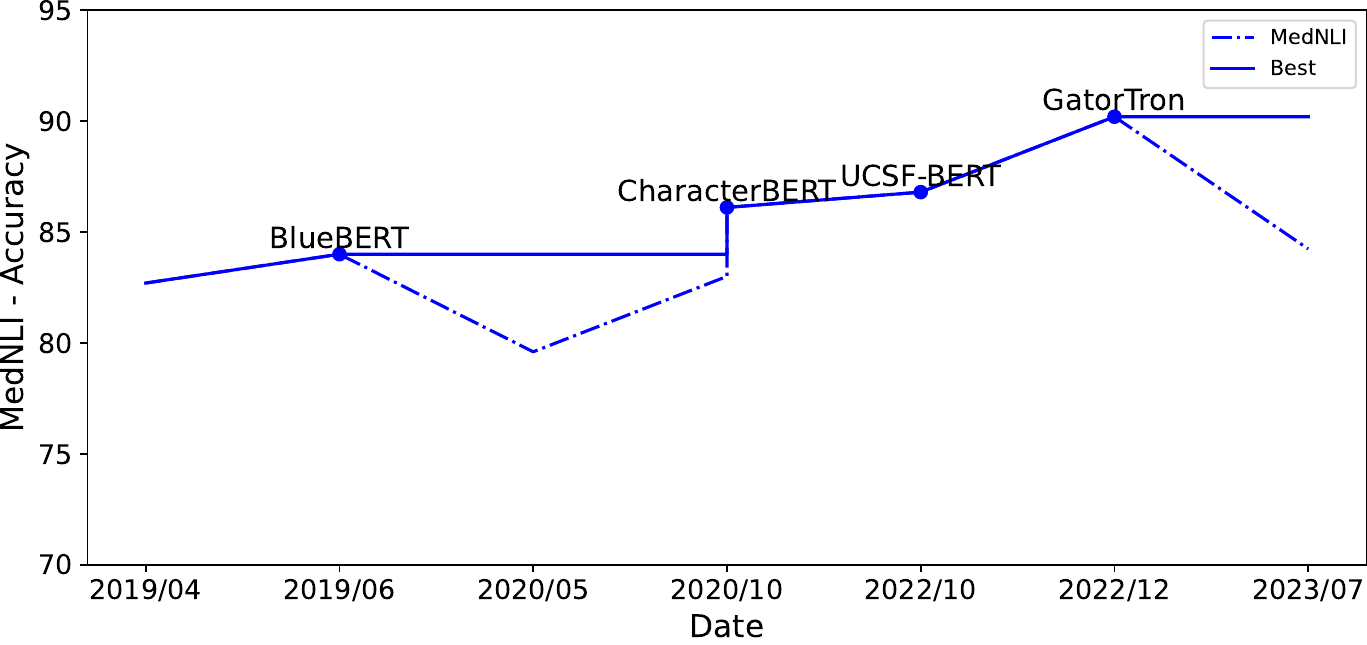}
    \end{subfigure}

    \caption{Details of performance changes for the RE datasets (Left) and the MedNLI dataset (Right) of BERT-based SciLMs.}
    \label{result_others_details_bert_archtec}
\end{figure}

\textbf{RE Task.}
From the list of SciLMs in Section \ref{sec_5tasks_effectiveness}, we retain only the models that utilize the BERT architecture. This yields five SciLMs: BlueBERT, ouBioBERT, CharacterBERT, BioLinkBERT, and Bioformer.
Figure \ref{result_others_details_bert_archtec} (Left) illustrates the performance changes for the DDI and ChemProt datasets.
We observe a gradual performance improvement over the past four years in the RE task. 
Specifically, for the DDI dataset, BioBERT first outperforms BlueBERT, and then CharacterBERT surpasses BioBERT. 
After that, BioLinkBERT and Bioformer also demonstrate improvements over the previous SciLMs.
ChemProt shows a quite similar pattern to the DDI dataset. However, the performance of CharacterBERT and ouBioBERT is similar (75.5 F1-score), and Bioformer does not outperform BioLinkBERT on the ChemProt dataset.

\textbf{NLI Task.}
Similar to previous tasks, we only retain the models that utilize the BERT architecture. This results in eight SciLMs: Clinical BERT (Emily), BlueBERT, ouBioBERT, UmlsBERT, CharacterBERT, UCSF-BERT, GatorTron, and KEBLM. 
Figure \ref{result_others_details_bert_archtec} (Right) illustrates the performance changes for the MedNLI dataset.
We observe a clear performance improvement in the MedNLI dataset from April 2019 to December 2022. BlueBERT surpasses Clinical BERT (Emily), followed by CharacterBERT improving over all previous SciLMs. Subsequently, UCSF-BERT and GatorTron also demonstrate improvement over all previous SciLMs. 
Currently, GatorTron stands as the best BERT-architecture model for the MedNLI dataset.

\section{Current Challenges and Future Directions}
\label{sec_challenges}

\subsection{Foundation SciLMs}

\subsubsection{SciLMs for non-English Language}
Study on multilingual and monolingual models for non-English languages has received significant attention in recent couples of years. Such LMs attempt to address the limitations in the solving of NLP tasks for non-English languages \cite{lai2023chatgpt}. The development of monolingual PLMs for other languages also witnessed a massive increase. This evolution improves the performance of PLMs in various downstream tasks, such as classification, summarization, and machine-reading comprehension in languages other than English. As a result, many benchmarking datasets and evaluations are performed in low-resource languages and have benefited the NLP research community. 

\par With respect to scientific text, most documents are written in English; therefore, SciLMs are initially designed to handle only English text. Few attempts have been made to investigate the performance of multilingual SciLMs in other languages. Table \ref{table_summary_multilingual} summarizes the current SciLMs for different languages. We find that nine languages other than English have pre-trained SciLMs. Chinese is the language spoken most among them; therefore, it receives a lot of attention from researchers. French, Japanese, and Spanish are also popular languages for which SciLMs have been evaluated. The lack of studies in other languages is probably due to the lack of large-scale scientific datasets. One possible explanation is that most research articles and scientific reports are written in English, making it a time-consuming task to collect and create datasets for non-English languages. The development of machine translation models is rapidly advancing \cite{WANG2022143} and can be integrated into future SciLMs. However, as far as we know, models for low-resource languages are not able to capture scientific phrases and academic writing styles; hence, it is essential to conduct more research on multilingual or non-English SciLMs.

\subsubsection{SciLMs for non-Biomedical Domain}

The landscape of SciLMs extends beyond the biomedical domain to encompass various scientific disciplines. 
In the chemical domain, there are 13 specialized SciLMs, primarily based on the BERT architecture.
The survey expands to less commonly explored domains such as Climate, CS, Cybersecurity, Geoscience, Manufacturing, Math, Protein, Science Education, and Social Science. 
In the multi-domain category, 
models like SciBERT, S2ORC-SciBERT, OAG-BERT, ScholarBERT, AcademicRoBERTa, and VarMAE are designed to handle diverse domains.
%
Despite progress, there is a significant gap in SciLM representation across scientific domains. 
While the biomedical domain has 85 identified models, other domains often have only one or two dedicated models, leading to concerns about \textit{limited generalizability}, \textit{neglect of domain-specific nuances}, and \textit{impediments to domain-specific applications}. 
The dominance of biomedical SciLMs raises questions about their generalizability across diverse scientific disciplines, potentially lacking contextual understanding for accurate representation in other fields.

To address these challenges, strategies include encouraging domain-specific research collaboration, open access to specialized datasets, incorporating transfer learning techniques, and establishing shared evaluation benchmarks. 
Collaboration between NLP researchers and domain experts can foster the development of SciLMs tailored to specific scientific domains. Open access to specialized datasets and leveraging transfer learning techniques allow adaptation to specific domains, even with limited data.
%
%
Shared benchmarks incentivize researchers, encouraging contributions to SciLM development across various domains and advancing research in scientific disciplines.

\subsubsection{Integrating Knowledge into SciLMs}

The exploration of integrating external knowledge, specifically Knowledge Bases (KBs), into SciLMs fills a crucial gap in the existing literature~\cite{pan2024unifying,wang2020language}. 
KBs play a pivotal role in enhancing LMs' capabilities within the scientific domain, providing structured information retrieval, domain-specific precision, contextual enrichment, informed reasoning, and task performance improvement. 
Tailored to scientific disciplines, KBs offer comprehensive knowledge coverage, enriching the context for LMs.

\begin{table}[ht]
\centering
\small
\caption{SciLMs with Knowledge Integrating. The No. column referred to Tables~\ref{table_lms_1},~\ref{table_lms_2}, and~\ref{table_lms_3}.}
\begin{tabular}{|c|l|c|c|c|c|c|}
\hline
\textbf{No.} & \textbf{Model} & \textbf{Domain} & \textbf{Arch.} & \textbf{Base-model} & \textbf{KBs Pretraining Task} & \textbf{KBs Used} \\
\hline
30 & SapBERT~\cite{liu-etal-2021-self} & Bio & En & PubMedBERT & Synonyms Clustering & UMLS \\
31 & UmlsBERT~\cite{michalopoulos-etal-2021-umlsbert} & Bio & En & ClinicalBERT (Emily) & CUI Words Connecting & UMLS  \\
34 & CODER~\cite{yuan2021coder} & Bio & En & PubMedBERT & Contrastive Learning & UMLS  \\
41 & KeBioLM~\cite{yuan-etal-2021-improving} & Bio & En & PubMedBERT & KG Embeddings (TransE~\cite{bordes2013translating}) & UMLS \\
45 & ProteinBERT~\cite{10.1093/bioinformatics/btac020} & Bio & En & ProteinBERT & Gene Ontology Prediction & UniRef90 \\
84 & DRAGON~\cite{yasunaga2022deep} & Bio & Others & BioLinkBERT-Large & MLM, KG Link Prediction & UMLS \\
\hline
\end{tabular}
\label{tab:SciLMsandKBs}
\end{table}

Table~\ref{tab:SciLMsandKBs} summarizes key models, such as UmlsBERT, ProteinBERT, DRAGON, KeBioLM, CODER, and SapBERT, each tailored to specific domains and pretraining tasks.
The integration process involves methodologies categorized into key approaches, such as integrating knowledge into the training objective and integrating knowledge into LM inputs. UmlsBERT, KeBioLM, CODER, and DRAGON exemplify the former, embedding knowledge directly into the learning process during pretraining. ProteinBERT, on the other hand, aligns more closely with the latter, incorporating external knowledge, such as Gene Ontology annotations, into the LM inputs to enhance context and semantics.
Challenges in knowledge integration include \textit{knowledge noise}, \textit{domain mismatch}, \textit{interpretability}, and \textit{coverage issue}. 
\textit{Knowledge noise} refers to challenges stemming from irrelevant or noisy information in KBs, encompassing outdated or incorrect data, ambiguous terms, and irrelevant concepts, which can significantly impact the precision of SciLMs, posing specific challenges in scientific domains where accuracy and precision are critical.
\textit{Domain mismatch} addresses disparities between KB language and scientific text nuances, requiring navigation for effective integration. 
\textit{Interpretability} concerns maintaining transparency in decision-making post-integration, crucial for validating reliability. 
The \textit{coverage issue} stems from the limited size of KBs, necessitating strategies to handle gaps in knowledge for accurate predictions.
Successfully overcoming these challenges is pivotal for enhancing SciLM efficacy in processing scientific text, allowing for reliable and precise outcomes.

\subsubsection{Build Large SciLMs}
\label{sec:build_large_scilms}
As shown in Figure \ref{fig_summary_size}, the majority of existing SciLMs have less than 1B parameters (i.e. BERT-level). One reason is that BERT-based SciLMs perform relatively well in various downstream tasks with limited budgets. Another reason is that building larger SciLMs requires much more computation resources and data. Galactica~\cite{taylor2022galactica} is the first attempt to scale SciLM up to 100B+ parameters. However, training such a large model requires a significantly larger amount of data and computation resources compared to BERT-like models. SciBERT was pre-trained with a single TPU v3 with 8 cores (similar to 2 A100 GPUs), whereas \citet{taylor2022galactica} used 128 A100 GPUs to pre-train their Galactica-120B. Therefore, an effective solution for pre-training SciLMs is a big challenge these days, especially for researchers who work in a university.

 Since the existence of high-quality open-sourced LLMs \cite{touvron2023llama, touvron2023llama2}, some researchers pay much attention on continual pre-training with these LLMs with extra scientific texts, relaxing the need for collecting large-scale scientific corpora and reducing the need for tons of computation resources as we train SciLMs from scratch. Therefore, effective continual pre-training is a promising direction for building large SciLMs. Meditron \cite{chen2023meditron} included a small proportion of the original pre-training corpus used by Llama-2 \cite{touvron2023llama2} to avoid forgetting when continual pre-training on data in the medical domain. 
 Some attempts have been made to guide the continual pretraining in the general domain \cite{gupta2023continual}, however, practical methods designed specifically for the scientific domain are still rare. 

As found by \citet{beltagy-etal-2019-scibert}, training SciLMs from scratch can benefit from designing domain-specific tokenizers for scientific domains, achieving better performances. Compared to other languages, there have been already a lot of English scientific texts for pre-training SciLMs. However, if we want to train a large SciLM (i.e. with more than 100B parameters) from scratch, the scale of existing data may not be enough. Detailedly, SciBERT used only 3.17B tokens during pre-training \cite{beltagy-etal-2019-scibert}, whereas Galactica consumed 450B tokens in total by repeating 106B tokens for approximately four epochs, which are 140 times more than those for training BERT-like models. According to the Chinchilla Scaling Laws \cite{hoffmann2022training}, LLMs with 63B parameters require 1.4T tokens for pre-training, and language models pre-trained with deduplicated texts perform better generalization ability, whereas the scientific texts in The Pile \cite{pile} contain much less tokens (96B tokens) than the recommended amount. Therefore, how to collect enough scientific texts for pre-training large SciLMs still remains a challenge these days. 

\subsubsection{Multi-modal SciLMs}
%
LMs capable of handling both language and non-language information such as image, audio, and video, have received considerable attention in recent years~\cite{wang2022MMPTMSurvey,DBLP:conf/ijcai/LongCHY22,DBLP:journals/pami/XuZC23}.
Notably, vision-and-language models pre-trained on vast amounts of language and image data have achieved significant success in various tasks, such as image captioning~\cite{DBLP:journals/pami/StefaniniCBCFC23} and image generation from text instructions \cite{DBLP:journals/corr/abs-2303-07909}. Lately, with the successful deployment of LLMs like GPT-4 \cite{openai2023gpt4}, many studies are emphasizing on training adapters that can transform non-language information to be treated in the same embedding space as language \cite{DBLP:journals/corr/abs-2304-10592,DBLP:journals/corr/abs-2305-06500}. Such architectures are expected to handle non-language data while retaining the extensive problem-solving capabilities of LLMs.

In the scientific domain, the advent of multi-modal models is also gaining momentum. Multi-modal SciLMs can be built by doing additional training on mono-modal or multi-modal PLMs on general domains, and thus take advantage of the robust performance of general-domain models. However, there are challenges yet to be overcome. For instance, in scientific domains, there is less data available compared to the general domain \cite{yuan2023ramm, lin2023medical}, which makes it difficult to sufficiently train or fine-tune the multi-modal SciLMs. In addition, in scientific domains, models handling more than two modalities are anticipated. Typically, scientific papers include many different types of information, such as tables, equations, figures, and codes. Therefore, their multi-modal integration into SciLMs should be considered as a crucial step forward. Also, biomedical SciLMs that incorporate a wide spectrum of data, such as CT, MRI, and ultrasound, are desirable \cite{lin2023medical}. However, research dealing with more than three modalities is relatively sparse.

Addressing these challenges requires strategies to increase the amount of available training data, including PDFs and LaTeX files. It should be encouraged to explore data augmentation techniques and learning methods that integrate external knowledge. In addition, the recent upsurge in LLMs signifies the need to develop multi-modal SciLMs based on publicly available LLMs (e.g., Llama 2 \cite{touvron2023llama2}) in scientific domains as well. Building such models will bring us closer to fully realizing the immense potential of multi-modal models in academic research.

\subsection{Evaluating the Effectiveness, Efficiency, and Trustworthiness of SciLMs}

\subsubsection{Issues with Evaluation and Comparison}
\label{sec_evaluating_issue}
As discussed in Section \ref{number_of_task_dataset_sota}, many SciLMs evaluate their models on a limited number of tasks and datasets.
This raises issues related to both evaluation and comparison. 
First, concerning evaluation, when proposed SciLMs assess their models on only one or a few tasks, it implies that their model is not comprehensively tested, and its performance may only be effective for some tasks while performing poorly on others. 
This can be explained by the fact that most existing SciLMs are based on an encoder-based architecture, such as BERT \cite{devlin-etal-2019-bert}, rather than being text-to-text models like T5 \cite{JMLR:v21:20-074}. 
Therefore, their models are not easily adaptable for evaluation across various NLP tasks.
With regard to the second issue of comparison, if we only compare different SciLMs on one or a few datasets, it becomes challenging to draw any reliable conclusions.

One promising direction to enhance the evaluation and comparison of different models is to create a benchmark comprising various tasks and datasets.
In the general domain, GLUE \cite{wang2018glue} and SuperGLUE \cite{NEURIPS2019_4496bf24} were introduced as standardized benchmarks for comparison. 
Motivated by this, in the biomedical domain, \citet{peng-etal-2019-transfer} and \citet{10.1145/3458754} introduced the BLUE and the BLURB benchmarks, respectively. 
The BLUE benchmark comprises five different tasks with ten datasets across these tasks, while the BLURB benchmark encompasses six different tasks with thirteen datasets across these tasks.
However, we observe that currently, there are only a few SciLMs in our list of 117 SciLMs that conduct evaluations on these benchmarks (for BLUE, the models are: BlueBERT, ouBioBERT, BioALBERT, and BioELECTRA; for BLURB, the models are: PubMedBERT, BioELECTRA, PubMedELECTRA, and BioLinkBERT).
Perhaps because many datasets in these benchmarks already yield high scores, researchers may be less motivated to evaluate on them. 
However, we believe that creating benchmarks with diverse tasks and datasets is a promising direction for future research to enable fair and reliable comparisons. 
We encourage future researchers to develop benchmarks with more tasks and datasets, even across different domains.


\subsubsection{Move Beyond Simple Tasks}
As shown in Tables \ref{table_popular_task} and \ref{table_popular_dataset}, most existing SciLMs focus their evaluation on simple tasks in NLP, such as NER and RE. 
In the top-20 popular datasets used to evaluate SciLMs, there are eleven NER datasets, but there is only one NLI dataset and two QA datasets.
As we know, NER and RE are basic tasks in NLP, while NLI and QA emphasize language understanding, testing the models' broader comprehension skills. However, datasets for these two tasks are not commonly used. 

With the current issues, we suggest that future work on SciLMs should shift their focus to the evaluation of more complex understanding tasks in NLP, such as NLI and QA. 
To accomplish this, the first step is to create additional datasets specifically designed for the NLI and QA tasks, serving as benchmarks for meaningful comparisons among SciLMs.
Learning from general domain, there are some directions that we can go.
For example, for the QA task, we can propose datasets for testing different skills, such as reasoning over multiple documents \cite{welbl-etal-2018-constructing,yang-etal-2018-hotpotqa} and conversational QA \cite{choi-etal-2018-quac,reddy-etal-2019-coqa}.
In the case of the NLI task, there has been only one dataset dedicated to English scientific text over the past few years: the MedNLI dataset \cite{romanov-shivade-2018-lessons}.
Recently \citet{jullien-etal-2023-nli4ct} introduce an NLI4CT dataset for clinical trial reports.
One notable feature of the dataset is that it is the first dataset with interpretation for the NLI task in processing scientific text. However, the size of the dataset is quite small, with only 2,400 instances. We believe that introducing more NLI datasets with larger sizes and containing adversarial samples would be helpful for evaluating SciLMs.

\subsubsection{Reliable SciLMs}
In addition to the issues regarding evaluation and comparison, we also need to pay much attention to three other aspects ---robustness, generalization, and explanation--- to obtain more reliable SciLMs.

In terms of robustness, we observe that not many SciLMs undergo evaluations on various types of adversarial tests. 
As seen in general domain tasks, many models exhibit strong performance on original datasets but experience a significant drop in performance on adversarial versions of those datasets \cite{jia-liang-2017-adversarial,ribeiro-etal-2018-semantically,jiang-bansal-2019-avoiding}. 
Therefore, to ensure robustness, it is essential to test SciLMs on adversarial tests during model development.

For the second aspect, generalization, we also observe a similar situation to that of robustness, where not many proposed SciLMs consider testing the generalization of their models.
It is noted that there are multiple ways to define the generalization ability of models. 
In this research, we simplify the definition by only considering the ability to generalize from one dataset to another dataset within the same task, whether in the same domain or a different domain.
Another concern regarding generalization ability is the lack of available datasets for testing models across tasks.
For example, in the NLI task, only one dataset, MedNLI, is available (fortunately, recently \citet{jullien-etal-2023-nli4ct} introduce an NLI4CT dataset). 
As a result, researchers cannot test the generalization ability of their models--for instance, training on one dataset and conducting evaluations on another dataset.
We suggest that future studies focus on evaluating models across multiple datasets with different distributions in the training set. 
By doing so, we can obtain a clearer understanding of the generalization ability of the models.

For the third aspect, we also observe a lack of research on SciLMs that emphasizes the aspect of explanation.
In the NLP domain, explanations are considered as the reasons for ``why [input] is assigned [label],'' and they are crucial for ensuring the reliability of the models \cite{wiegreffe2021teach}.
However, this point is not well-discussed and emphasized in the scientific domain.
For example, to the best of our knowledge, there are currently only four datasets in the scientific domain that include explanation information---PubMedQA \cite{jin-etal-2019-pubmedqa} (long answer can be considered as explanation), SciFact \cite{wadden-etal-2020-fact}, QASPER \cite{dasigi-etal-2021-dataset}, and NLI4CT \cite{jullien-etal-2023-nli4ct}.
We believe that, to enhance the explanation ability of SciLMs, more datasets dedicated to explanation should be proposed for use in evaluating and analyzing the models' capabilities.

\section{Conclusion}
\label{sec_conclusion}
We surveyed existing LMs for processing scientific text. 
Specifically, we reviewed 117 SciLMs across various scientific domains, languages and architectures.
We presented an extensive analysis of these SciLMs, highlighting the general bias of previous work towards the biomedical domain and BERT-based encoders, and the uprising of non-English SciLMs.
We also provided new insights on the current SOTA performance of SciLMs across commonly used tasks and datasets, and its evolution through the recent years.
Finally, we discussed challenges and showed the two potential directions. 
The first direction is about establishing foundational SciLMs, for which we provided practical recommendations to improve their performance, such as integrating knowledge bases, developing larger models, or leveraging the multi-modal content present in scientific papers.
%
Additionally, we proposed actionable recommendations on the development of SciLMs for low-resource languages, extending beyond the English language and biomedical domains.
The second direction is about improving the evaluation of SciLMs, where we proposed to incorporate a broader range of tasks, datasets, and more challenging evaluation criteria.
We also underscored the urgent need to establish standardized benchmarks across various domains and tasks, ensuring a fair and reliable comparison between proposed SciLMs.

%
%


\bibliographystyle{ACM-Reference-Format}
\bibliography{all}

\appendix

\newpage

\section{Preliminary Information}

\subsection{Existing Tasks and Datasets in Scientific Articles}
\label{app_sec_tasks_datasets}

We present important information of existing QA datasets in Table \ref{tab_qa_tasks}.

\begin{table}

  \caption{Existing question answering datasets. 
  }
  \label{tab_qa_tasks}
 \resizebox{\textwidth}{!}{%

  \begin{tabular}{l l c c c c c }
    \toprule
  \textbf{Year} &  \textbf{Dataset} & \textbf{Answer Style}  &  \textbf{Size}  & \textbf{Domain} & \textbf{Question Source} & \textbf{Level} \\
    \midrule

 2015  & BioASQ \cite{BIOASQ_dataset} & Yes/No, Extractive & 3,743 &   Biomedical &  Expert &  Full paper \\

 2018  & Biomed-Cloze \cite{dhingra-etal-2018-simple} &  Cloze-style & 1M &  Biomedical & Automated & Passage \\

 2018  & emrQA \cite{pampari-etal-2018-emrqa} &  Extractive &  400K & Clinical documents  &  Automated & Clinical note  \\

 2018  & BioRead \cite{pappas-etal-2018-bioread} &  Multiple-choice & 16.4M & Biomedical & Automated &  900 tokens \\

 2018  & MedHop \cite{welbl-etal-2018-constructing} & Multiple-choice & 2,508 & Molecular biology  & Automated  &  Abstract\\

 2019  & PubmedQA \cite{jin-etal-2019-pubmedqa} & Yes/no/maybe & 273.5K & Biomedical  & Expert + Automated & Abstract \\

 2020  &  BioMRC \cite{pappas-etal-2020-biomrc} & Multiple-choice & 821K & Biomedical  & Automated &  Abstract\\

 2021  & QASPER \cite{dasigi-etal-2021-dataset} & \makecell{Extractive, 
Abstractive,  \\  
Yes/No, 
Unanswerable}  & 5,049 & NLP papers  & NLP practitioner & Full paper \\
  \bottomrule
\end{tabular}

 }
\end{table}





\section{Existing LMs for Processing Scientific Text}
\label{app_sec_existing_scilms}

\begin{table}

  \caption{
  A continuation of the table from Table~\ref{table_lms_1}.
  In the \textbf{Training Objective} column, 
  \textit{MLM} denotes Masked Language Modeling,  
  \textit{NSP} denotes Next Sentence Prediction, and   \textit{SOP} denotes Sentence Order Prediction.
  In the \textbf{Type of Pre-training} column, \textit{CP} and \textit{FS} denote Continual Pretraining and From Scratch, respectively. 
 In the \textbf{Domain} column,  \textbf{CS} represents computer science, \textbf{Bio} represents Biomedical domain, \textbf{Chem} represents Chemical domain, and \textbf{Multi} represents multiple domains. 
It is noted that the date information is chosen from the first date the paper appears on the Internet. 
  }
  \label{table_lms_2} 

      \resizebox{\textwidth}{!}{%

  \begin{tabular}{l l l  c c c c c c }


    \toprule
  \textbf{No.} &  \textbf{Date} & \textbf{Model}  & \textbf{Architecture} &  \textbf{Training Objective} & \makecell{\textbf{Type of} \\ \textbf{Pre-training} } & \makecell{\textbf{Model} \\ \textbf{Size}} & \textbf{Domain} &  \textbf{Pre-training Corpus}  \\

   \midrule
42  &    2021/04  &    SINA-BERT \cite{taghizadeh2021sinabert}&  BERT & MLM & CP & 110M & Bio & \makecell{Manually Collected \\ (2.8M Documents)}   \\

\midrule
43  &    2021/05  &  MathBERT (Peng) \cite{peng2021mathbert}   &   BERT  & \makecell{MLM, Masked Substructure Prediction, \\ Context Correspondence Prediction} & CP & 110M & Math & \makecell{Formula-Context Pairs \\ Extracted from arXiv} \\ \midrule

44  &    2021/05  &   NukeLM \cite{burke2021nukelm}   & \makecell{BERT \\ RoBERTa-Base \\ RoBERTa-Large} & MLM & CP & \makecell{110M \\ 125M \\  355M} & Chem & \makecell{U. S. Department of \\ Energy Office Scientific and \\ 
Technical Information Database} \\ \midrule

45  &    2021/05  &    ProteinBERT \cite{10.1093/bioinformatics/btac020} &  BERT &  Corrupted Token, Annotation Prediction & FS & 16M & Bio & UniProtKB/UniRef90 + GO \\ \midrule

46 &    2021/05  &    SciFive  \cite{phan2021scifive}   & T5 & Span Corruption Prediction  & CP & \makecell{220M  \\ \&  770M} & Bio & PubMed + PMC \\ \midrule

47  &    2021/06  &    BioELECTRA \cite{kanakarajan-etal-2021-bioelectra}  &   ELECTRA & Replaced Token Prediction & CP \& FS & 110M & Bio & PubMed + PMC \\ \midrule

48  &    2021/06  &    ChemBERT \cite{doi:10.1021/acs.jcim.1c00284}  &  RoBERTa & MLM & CP & 110M  & Chem & 
ACS Publications \\ \midrule

49  &    2021/06  &    EntityBERT \cite{lin-etal-2021-entitybert}  & BERT  & Entity-Centric MLM & CP & 110M   & Bio & MIMIC \\ \midrule

50  &    2021/06  & MathBERT (Shen) \cite{shen2023mathbert} & RoBERTa & MLM &  CP & 110M & Math & \makecell{Mathematics Curricula \\  + Mathematics Textbooks \\ + Mathematics Course Syllabi \\ + Mathematics Paper Abstracts} \\ \midrule

51  &    2021/07  &    MedGPT  \cite{kraljevic2021medgpt}  &  \makecell{GPT2 + GLU \\ + RotaryEmbed}  & LM & CP & N/A & Bio &  Electronic Health Record Notes \\ \midrule

52  &    2021/08  &    SMedBERT  \cite{zhang2021smedbert}  & SMedBERT & \makecell{Masked Neighbor Modeling, SOP, \\
Masked Mention Modeling, MLM} & FS & N/A & Bio & DXY BBS (Bulletin Board System)  \\ \midrule

53  &    2021/09  &  Bio-cli \cite{carrino2021biomedical} & RoBERTa & \makecell{MLM, Subword Masking \\ or Whole Word Masking} & FS & 125M & Bio & Clinical Corpus + Biomedical Corpus \\ \midrule

54  &    2021/09  &    MatSciBERT \cite{gupta2022matscibert}  &  BERT & MLM & CP & 110M & Chem & Material Science Corpus \\ \midrule

55  &    2021/10  &    ClimateBert  \cite{webersinke2022climatebert}  & DistilROBERTA & MLM & CP & 66M & Climate & \makecell{News Articles  + Research Abstracts \\ + Corporate Climate Reports} \\ \midrule

56  &    2021/10  &    MatBERT \cite{Trewartha2022-sl}  &  BERT & MLM & FS & 110M & Chem &  \makecell{Peer-Reviewed Materials \\ Science Journal Articles} \\ \midrule

57  &    2021/11  &   UTH-BERT \cite{aclinical_kawazoe}    & BERT & MLM, NSP & FS & 110M & Bio & \makecell{Clinical Texts from The \\ University of Tokyo Hospital}   \\ \midrule

58  &    2021/12  &    ChestXRayBERT \cite{9638337}   &  BERT & MLM, NSP & CP & 110M & Bio & 20617 Scientific Papers in PMC  \\ \midrule 

59  &    2021/12  &    MedRoBERTa.nl \cite{Verkijk_Vossen_2021}  &   RoBERTa & MLM & FS & 123M & Bio & \makecell{Dutch Hospital Notes, from EHRs} \\ \midrule

60  &    2021/12  &    PubMedELECTRA \cite{TINN2023100729}   &   ELECTRA & Replaced Token Prediction & FS & \makecell{110M \\ \& 335M} & Bio & PubMed  \\ \midrule

61  &    2022/01  &   Clinical-Longformer  \cite{li2022clinicallongformer}   &  \makecell{BigBird \\ Longformer}  & MLM & CP & \makecell{166M \\ 149M}  & Bio & MIMIC-III  \\ \midrule

62 &    2022/03 & BioLinkBERT \cite{yasunaga-etal-2022-linkbert}   & BERT & \makecell{MLM, \\ Document Relation Prediction}  & FS & \makecell{110M \\ \& 340M} & Bio  & PubMed + PMC \\  \midrule

63  &    2022/04  &    BioBART  \cite{yuan-etal-2022-biobart}   &  BART  & \makecell{Text Infilling,\\ Sentence Permutation} & CP & \makecell{140M \\ \& 400M} & Bio & PubMed  \\ \midrule

64  &    2022/04  &    SecureBERT  \cite{1a3e2a1ca6f94c7589c870bc0b3e1e31}   & RoBERTa & MLM & CP & 125M & Cybersecurity & \makecell{Manually Collected \\ (98,411 Documents)}   \\ \midrule

65   &    2022/05  &    BatteryBERT \cite{doi:10.1021/acs.jcim.2c00035}   &  BERT & MLM & CP \& FS & 110M & Chem & \makecell{Battery-Related Scientific Papers \\ from RSC, Elsevier, Springer} \\ \midrule

66  &    2022/05  &   bsc-bio-ehr-es  \cite{carrino-etal-2022-pretrained}   &  RoBERTa & MLM & FS & 125M & Bio & \makecell{Electronic Health Record Corpus
\\ + Biomedical Corpus} \\ \midrule

67   &    2022/05  &    ChemGPT  \cite{frey2022neural}  & GPT & Autoregressive LM & FS & 1B & Chem & SMILES from PubCHEM \\ \midrule

68  &    2022/05  &    PathologyBERT \cite{santos2022pathologybert}    &  BERT & MLM, NSP & FS & 110M & Bio &  \makecell{Manually Labeled Pathology Reports \\
+ Unlabeled Pathology Reports from EUH} \\ \midrule

69  &    2022/05  &    ScholarBERT \cite{hong2023diminishing}   & BERT &  MLM & FS & 770M & Multi & Public Resource Dataset \\ \midrule

70 &    2022/06 & RadBERT \cite{yan2022radbert}   & RoBERTa & MLM & CP & 110M & Bio &  Radiology Reports \\ \midrule

71  &    2022/06  &    SciDeBERTa \cite{jeong2022scideberta}   &  DeBERTa &  MLM & CP & N/A & Multi & S2ORC \\  \midrule

72  &    2022/06  &    SsciBERT \cite{10.1007/s11192-022-04602-4} & BERT & MLM, NSP & CP & 110M & Social Science  & Collected from Web of Science \\  \midrule

73 &    2022/06 & ViHealthBERT \cite{minh-etal-2022-vihealthbert}  & PhoBERT & \makecell{MLM, NSP, \\ Capitalized Prediction} & CP & 110M & Bio & \makecell{OSCAR Dataset \\ + Their Text Mining Corpus} \\  \midrule

74  &    2022/07  &    Clinical Flair \cite{rojas-etal-2022-clinical}   &  \makecell{Flair} & Character-Level Bi-LM & CP & N/A & Bio & Free-Text Diagnostic Suspicions \\  \midrule

75  &    2022/08  &    KM-BERT  \cite{Kim2022-er}  &  BERT & MLM, NSP & CP & 99M & Bio &  \makecell{Medical Textbooks \\ + Health Information News \\ + Medical Research Articles}  \\ \midrule

76  &    2022/08  &    MaterialsBERT (Shetty) \cite{Shetty_2023}   &  PubMedBERT & MLM, NSP, Whole-Word Masking & CP & 110M & Chem &  Polymer Scholar  \\ \midrule

77   &    2022/08  &    ProcessBERT \cite{KATO2022957}   &  
 BERT & MLM, NSP & CP & 110M & Chem & ChemECorpus  \\

  \bottomrule
\end{tabular}

}
\end{table}

\begin{table}
  \caption{
  A continuation of the table from Table~\ref{table_lms_1} and Table~\ref{table_lms_2}. 
  }
  \label{table_lms_3} 
      \resizebox{\textwidth}{!}{%

  \begin{tabular}{l l l  c c c c c c }


    \toprule
  \textbf{No.} &  \textbf{Date} & \textbf{Model}   & \textbf{Architecture} &  \textbf{Training Objective} & \makecell{\textbf{Type of} \\ \textbf{Pre-training} } & \makecell{\textbf{Model} \\ \textbf{Size}} & \textbf{Domain} &  \textbf{Pre-training Corpus}  \\

\midrule
78 &    2022/09 & BioGPT \cite{10.1093/bib/bbac409}   & GPT & Autogressive LM & FS & \makecell{347M  \\ \& 1.5B} & Bio & PubMed \\

\midrule
79  &    2022/09  &    ChemBERTa-2 \cite{ahmad2022chemberta}  &  RoBERTa &  MLM, Multi-Task Regression & FS & 125M & Chem & SMILES from PubCHEM   \\

 \midrule
80  &    2022/09  & CSL-T5   \cite{li-etal-2022-csl}   &   T5 & \makecell{Fill-in-the-blank-style \\ Denoising Objective} & FS & 220M & Multi & Chinese Academic Journals \\

 \midrule 
81  &    2022/09  &    MaterialBERT (Yoshitake) \cite{yoshitake2022materialbert}  &  BERT & MLM, NSP & FS & 110M &  Chem & Scientific Articles from NIMS \\

\midrule
82  &    2022/10  & AcademicRoBERTa \cite{yamauchi-etal-2022-japanese} & RoBERTa & MLM & FS & 125M & Multi & CiNii Articles \\

\midrule
 83  &    2022/10  & Bioberturk \cite{PPR:PPR560833}  & BERT & MLM, NSP & CP \& FS & N/A & Bio & \makecell{Turkish Medical Articles \\ +
Turkish Radiology Thesis} \\

\midrule
84  &    2022/10  &    DRAGON \cite{yasunaga2022deep}   &  GreaseLM & \makecell{MLM, KG Link Prediction} & CP & 360M & Bio & PubMed + UMLS KG \\  \midrule

85  &    2022/10  &    UCSF-BERT \cite{sushil2022developing}   &   BERT & MLM, NSP & FS & 135M & Bio &  Clinical Notes of UCSF Health  \\ \midrule

86  &    2022/10  & ViPubmedT5 \cite{phan2023enriching}  & T5 & Spans-masking Learning & CP & 220M & Bio & ViPubmed \\ \midrule

87 &    2022/11 & Galactica \cite{taylor2022galactica}  & GPT & Autogressive LM & FS & \makecell{125M \\ \& 1.3B \& 6.7B \\ \& 30B \& 120B} & Multi &  \makecell{Papers, Reference Material, Knowledge Bases, \\Common Crawl, Code, <work> Datasets} \\ 
\midrule

88  &    2022/11  &    VarMAE \cite{hu-etal-2022-varmae}  &  RoBERTa & MLM & CP & 110M & Multi  & \makecell{Semantic Scholar Corpus \\ + Finance-Related Online Platforms}  \\ \midrule

89  &    2022/12  &    AliBERT \cite{berhe-etal-2023-alibert}  & BERT & MLM & FS & 110M & Bio & \makecell{Drug Database 550MB + Cochrane 27MB  + \\
 Articles 4300MB  + Thesis 300MB 
+ RCP 2200MB}  \\ \midrule

90 &    2022/12 & BioMedLM \cite{cite_BioMedLM}   & GPT2 & Autogressive LM & FS & 2.7B &  Bio  &  PubMed + PMC \\ \midrule

91  &    2022/12  &    BioReader \cite{frisoni-etal-2022-bioreader}  & T5 + RETRO & MLM & CP & 229.5M & Bio & PubMed  \\ \midrule

92  &    2022/12  &    ClinicalT5 \cite{lu-etal-2022-clinicalt5}  & T5 & Span-mask Denoising Objective  & CP & \makecell{220M \\ \& 770M} & Bio & MIMIC-III  \\ \midrule

93  &    2022/12   & CySecBERT \cite{bayer2022cysecbert}  & BERT & MLM, NSP & CP & 110M  & Cybersecurity & \makecell{Blog Data + arXiv Data + Twitter Data \\ + National Vulnerability Database} \\ \midrule

94  &    2022/12  &    GatorTron \cite{Yang2022-pf}  & BERT & MLM & FS & 8.9B & Bio &  \makecell{De-identified Clinical Notes \\ + PubMed + Wikipedia}  \\ \midrule

95  &    2022/12   & Med-PaLM \cite{singhal2023large}  & Flan-PaLM & Instruction Prompt Tuning & CP & 540B & Bio & N/A \\ \midrule


96  &    2023/01  &    Clinical-T5 \cite{Clinical_T5_lehman} &  T5   & \makecell{Fill-in-the-blank-style \\ Denoising Objective} & \makecell{CP \& FS} & \makecell{220M \\ \& 770M}  & Bio & MIMIC-III + MIMIC-IV  \\ \midrule

97  &    2023/01  &    CPT-BigBird \cite{Louisa_heyarticle}   &  \makecell{BigBird \\ Longformer}  & MLM & CP & \makecell{166M \\ 149M}  & Bio & Senior Scholars' Basket of Journals  \\ \midrule 

98  &    2023/01  &   ProtST  \cite{xu2023protst}   &  BERT   & \makecell{Masked Protein Modeling \\ + Contrastive Learning \\
+ Multi-modal Masked Prediction} & CP \& FS  &  N/A  &  Protein  &  ProtDescribe  \\ \midrule

99  &    2023/01  &  SciEdBERT   \cite{liu2023context}    &  BERT   & MLM & CP  & 110M &  Science Education &  \makecell{The Academic Journal Dataset \\ + Students’ Responses}   \\ \midrule

100  &    2023/02  &    Bioformer \cite{fang2023bioformer}  &  BERT   & MLM, NSP & FS & 43M & Bio & PubMed + PMC   \\ \midrule

101  &    2023/02  &   \makecell[l]{Lightweight  Clinical \\ Transformers \cite{rohanian2023lightweight} }  &  DistilBERT   &  \makecell{MLM, \\ Knowledge Distillation Objective} & CP  & \makecell{15M \& 18M \\ \& 25M \& 65M} & Clinical & MIMIC-III   \\ \midrule

102  &    2023/03  &    ManuBERT \cite{ManuBERT_kumar} & BERT & MLM & CP & \makecell{110M  \\ \& 126M} & Manufaturing & \makecell{Manufacturing Process Journals \\ + Six Commonly Used Textbooks}  \\ \midrule

103  &    2023/03  &   RAMM  \cite{yuan2023ramm}   &  PubmedBERT   & \makecell{MLM, Image-Text Matching, \\ Image-Text Contrastive Learning} & CP  & N/A & Bio & ROCO, MIMIC-CXR, PMCPM   \\ \midrule

104  &    2023/04  &  DrBERT  \cite{labrak2023drbert}  &  RoBERTa & MLM & FS & 110M & Bio & \makecell{Online Sources \\ + Private Hospital Stays Reports}  \\ \midrule

105  &    2023/04  &  MOTOR   \cite{lin2023medical}   &  BLIP   & \makecell{MLM, Image-Text Matching, \\ Image-Text Contrastive Learning} &  CP   &  N/A & Bio &   MIMIC-CXR \\ \midrule

106  &    2023/05  &   BiomedGPT \cite{zhang2023biomedgpt}  &  BART & \makecell{Text Infilling,\\ Sentence Permutation} & FS & \makecell{33M \\ \& 93M \\ \& 182M}  & Bio &  \makecell{CheXpert + PathVQA + Retinal Fundus \\ + International Skin Imaging Collaboration \\  + Peir Gross   + CytoImageNet + MIMIC-III \\ + DeepLesion + NCBI BioNLP Corpus \\ + MedICaT + PubMed + SLAKE + IU X-ray  } \\ \midrule

107  &    2023/05  &   Patton  \cite{jin-etal-2023-patton}    &  \makecell{GNN-nested \\ Transformer}  & \makecell{Network-contextualized MLM,  \\ Masked Node Prediction} & CP  & N/A &  Multi &  \makecell{Academic Networks from Microsoft Academic Graph, \\ E-commerce Networks from Amazon}  \\ \midrule

108  &    2023/05  &  TurkRadBERT \cite{türkmen2023harnessing} & BERT & MLM, NSP & CP \& FS & 110M & Bio &  \makecell{General Turkish Corpus + Head CT Reports \\ + Turkish Biomedical Corpus \\ 
+ Turkish Electronic Radiology Theses} \\ \midrule

109  &    2023/06  &   CamemBERT-bio  \cite{touchent2023camembertbio}  & CamemBERT    & Whole Word Masking  & CP  &  111M & Bio &  Biomed-fr: ISTEX, CLEAR, and E3C  \\ \midrule

110  &    2023/06  &   ClinicalGPT \cite{wang2023clinicalgpt} & BLOOM & \makecell{Supervised Fine Tuning \\ 
Rank-based Training} & Instruction Tuning & CP & Bio & Supervised Task Datasets  \\  \midrule

111  &    2023/06  &   EriBERTa  \cite{delaiglesia2023eriberta}  &  \makecell{RoBERTa \\ Longformer}   & MLM & CP \& FS  & \makecell{125M \\ 149M} & Bio &  \makecell{MIMIC-III, EMEA, ClinicalTrials, PubMed, \\ SNOMED-CT, SPACCC,  UFAL, Wikipedia Med, \\ Private Clinical Documents, Medical Crawler}  \\ \midrule

112  &    2023/06  &  K2   \cite{deng2023k2}  &  LLaMA   & Cosine Loss   & CP  & 7B & Geoscience &  \makecell{Geoscience-Related Wikipedia Pages, \\ 
Open-access Geoscience Papers, \\
Geoscience Paper’s Abstracts}   \\ \midrule

113  &    2023/06  &    PharmBERT \cite{10.1093/bib/bbad226}   &  BERT   & MLM & CP  & 110M & Bio &  DailyMed  \\ \midrule


114  &    2023/07  &    BioNART \cite{asada-miwa-2023-bionart}   &  BERT   & Connectionist Temporal Classification & CP  & 110M &  Bio &  PubMed/MEDLINE   \\ \midrule

115  &    2023/07  &    BIOptimus \cite{pavlova-makhlouf-2023-bioptimus}   &  BERT   & MLM & CP  & 110M & Bio &  PubMed   \\ \midrule

116  &    2023/07  &    KEBLM \cite{LAI2023104392}  &  BERT & \makecell{MLM, Ranking Objective, \\
Contrastive Learning}  & CP  & N/A  & Bio &   UMLS, PubChem, MSI  \\ \midrule

117  &    2023/08  &    GIT-Mol \cite{liu2023gitmol}   &  \makecell{BLIP2}  & \makecell{Xmodal-Text Matching, \\ 
Xmodal-Text Contrastive Learning}  &  N/A &  700M & Chem &  PubChem \\


  \bottomrule
\end{tabular}

    }

\end{table}

\subsection{Non-Scientific Domains}
\label{sec:scilms-non-scientific-domains}

While our primary focus remains on SciLMs tailored for scientific text, it is essential to briefly summarize LMs trained on non-scientific text during the same period. By providing a concise overview of these models, we hope to expand our knowledge of the overall LM landscape.

According to our survey, we have found 21 LMs that fall under this category. Most of these models were built upon the BERT or similar architecture. Moreover, we observed that models with a size of 110-125M are typically preferred.


\textbf{BERT-Based Models.}
There are 9 non-SciLMs based on the BERT architecture, which have been trained with 110-125M parameters for various domains, including General, Bio, Financial, Aviation, Historical, and Political. Among them, CT-BERT~\cite{10.3389/frai.2023.1023281}, FinBERT~\cite{araci2019finbert}, CancerBERT~\cite{zhou2022cancerbert}, Bioreddit-BERT~\cite{basaldella-etal-2020-cometa}, Aviation-BERT~\cite{chandra2023aviation}, and SafeAeroBERT~\cite{andrade2023safeaerobert} were continually pretrained on previous work's checkpoints. In contrast, MacBERTh~\cite{manjavacas-arevalo-fonteyn-2021-macberth} and HmBERT~\cite{schweter2022hmbert} were pretrained from scratch using their own corpora. ConfliBERT~\cite{hu-etal-2022-conflibert} was pretrained using both strategies.

Non-NSP BERT-based models such as PetroBERT~\cite{rodrigues2022petrobert} and AnchiBERT~\cite{9534342} were continually pretrained using model weights from previous work. They were built for the petroleum domain in the Portuguese language and the Ancient Chinese domain, respectively. PeerBERT~\cite{10.1145/3506860.3506892} is a RoBERTa-based model pretrained from scratch on Peer Review Comments.

\textbf{Specialized Architecture-Based Models.}
Several models combine MLM with additional objectives to solve specific tasks. For instance, MMBERT~\cite{khare2021mmbert} is a multi-modal BERT model that utilizes a CNN to capture image and text features for medical domain visual QA tasks. It was pretrained from scratch on a corpus of images and text. DisorBERT~\cite{aragon-etal-2023-disorbert}, on the other hand, explores lexical knowledge to continually pretrain the BERT model to pay more attention to words related to mental disorders. icsBERTs~\cite{LIU2023127} adapts Multi-Task Learning and Sentence Structural Objective to improve the performance of existing PLMs in the Chinese business domain.
Moreover, TravelBERT~\cite{zhu2023pre} is a continued pretraining LM aiming to learn entity and topic knowledge by incorporating triple classification objective and title matching objective, respectively. The model was developed for the Chinese language in the tourism domain. 
Finally, BudgetLongformer~\cite{niklaus2022budgetlongformer} uses LongFormer to train from scratch on legal data.

\textbf{Generation-Based Models.}
Several autoregressive Transformer architectures have also been employed. For instance, FinBART~\cite{hongyuan-etal-2023-finbart} is a BART-based model pretrained from scratch on its Chinese-language financial data. Similarly, JaCoText~\cite{espejel2023jacotext} was pretrained from CoTexT~\cite{phan2021cotext} checkpoints, basically T5-Base and T5-Large for the Code domain. 
After the release of GPT-3~\cite{brown2020language}, there has been a rise in the development of large-scale LMs. For example, PANGU-$\alpha$ \cite{zeng2021pangu} is a large-scale autoregressive pretrained Chinese LM built for the General domain with increasing magnitude of parameter sizes, namely 2.6B, 13B, and 200B. The model introduces a query layer on top of the stacked Transformer layers to explicitly induce the expected output, which is the prediction of the next token in the pretraining stage of the autoregressive model.

\section{Effectiveness of LMs for Processing Scientific Text}

\subsection{Details about Tasks and Datasets of SciLMs}
\label{app_sec_details_tasks_datasets}

\newcolumntype{P}[1]{>{\centering\arraybackslash}p{#1}}

{\small \tabcolsep=2.2pt

\begin{xltabular}{\textwidth}{l p{2.15cm} p{4.2cm} X}




\caption{Tasks and Datasets are used in existing LMs for scientific text.
  The underlying datasets are those in which the proposed SciLMs outperform previous models or achieve SOTA results.} 

\label{tab_result_1} \\

    \toprule
 \textbf{No.} &   \textbf{Model} & \textbf{Tasks } & \textbf{Datasets}\\
 \\ 





\midrule
\multirow{4}{*}{1} & \multirow{4}{*}{BioBERT} & \multirow{2}{*}{\textcolor{blue}{Named Entity Recognition}} & \underline{NCBI-disease}, \underline{BC5CDR-disease}, \underline{BC5CDR-chemical}, \underline{BC4CHEMD}, \underline{BC2GM}, JNLPBA, LINNAEUS,  i2b2 2010, Species-800 \\

  &  & \textcolor{purple}{Relation Extraction} & \underline{ChemProt}, GAD, EU-ADR  \\
 
&    & \textcolor{orange}{Question Answering} & \underline{BioASQ 5b-factoid}, \underline{BioASQ 6b-factoid}, BioASQ 4b-factoid \\

 \midrule
 \multirow{1}{*}{2} &    \multirow{1}{*}{BERT-MIMIC} &  \multirow{1}{*}{\textcolor{blue}{Named Entity Recognition}} & \underline{i2b2 2010}, \underline{i2b2 2012}, \underline{SemEval 2014 Task 7}, \underline{SemEval 2015 Task 14} \hl{(All)}  \\

 \midrule
\multirow{5}{*}{3} &    \multirow{5}{*}{SciBERT} &  \textcolor{blue}{Named Entity Recognition} &  \underline{BC5CDR}, \underline{SciERC}, JNLPBA, NCBI-disease \\

&    & \textcolor{magenta}{PICO Extraction} & \underline{EBM-NLP} \\
    
&    & \textcolor{teal}{Text Classification} & \underline{ACL-ARC}, \underline{SciCite}, Paper Field \\ 

&    & \textcolor{purple}{Relation Extraction} & \underline{ChemProt}, SciERC \\ 

&    & \textcolor{cyan}{Dependency Parsing}  & GENIA-LAS, GENIA-UAS\\

\midrule
 \multirow{2}{*}{4} &    \multirow{2}{*}{BioELMo} &  \textcolor{blue}{Named Entity Recognition} & \underline{BC2GM} (Probing task),  CoNLL 2003 \\

& & \textcolor{Bittersweet}{Natural Language Inference} & \underline{MedNLI} (Probing task), SNLI\\

\midrule
 \multirow{2}{*}{5} &    \multirow{2}{*}{Clinical BERT (E)} &  \textcolor{blue}{Named Entity Recognition} &  i2b2 2006, i2b2 2010, i2b2 2012, i2b2 2014 \\

& & \textcolor{Bittersweet}{Natural Language Inference} & \underline{MedNLI} \\

\midrule
 \multirow{1}{*}{6} &    \multirow{1}{*}{ClinicalBERT (K)} &  \multirow{1}{*}{\textcolor{black}{Readmission Prediction}} & \multirow{1}{*}{30-day hospital
readmission prediction}  \\

\midrule
 \multirow{5}{*}{7} &    \multirow{5}{*}{BlueBERT} &  \textcolor{blue}{Named Entity Recognition}  &     \underline{BC5CDR-disease}, \underline{BC5CDR-chemical}, \underline{ShARe/CLEF} \\

 & & \textcolor{purple}{Relation Extraction}  & \underline{DDI}, \underline{ChemProt}, \underline{i2b2 2010}\\ 

& & \textcolor{Green}{Sentence Similarity}  & \underline{BIOSSES}, \underline{MedSTS}\\ 

& & \textcolor{brown}{Document Multi-label Classification}  & \underline{HOC}\\ 

& & \textcolor{Bittersweet}{Natural Language Inference} &  \underline{MedNLI} \hl{(All)}\\

\midrule
 8 &    G-BERT &  \textcolor{Fuchsia}{Medication Recommendation Task}  &     Medication Recommendation Task \\

\midrule
9 &    BEHRT &  \textcolor{BlueViolet}{Disease Prediction}  & \underline{Next Visit}, \underline{Next 6M}, \underline{Next 12M} \hl{(All)}   \\

\midrule
10 &    BioFLAIR &  \textcolor{blue}{Named Entity Recognition}  &  \underline{BC5CDR}, \underline{Species-800}, NCBI-disease, LINNAEUS, JNLPBA, BC5CDR-disease  \\

\midrule
 \multirow{1}{*}{11} &    \multirow{1}{*}{EhrBERT} & \multirow{1}{*}{\textcolor{black}{Entity Normalization}}  & \underline{MADE 1.0}, \underline{NCBI-disease}, \underline{CDR} \hl{(All)}  \\

\midrule
\multirow{5}{*}{12} &    \multirow{5}{*}{S2ORC-SciBERT} &  \textcolor{blue}{Named Entity Recognition} &  BC5CDR, SciERC, JNLPBA, NCBI-disease \\

&    & \textcolor{magenta}{PICO Extraction} & EBM-NLP \\
    
&    & \textcolor{teal}{Text Classification} & ACL-ARC, SciCite, Paper Field \\ 

&    & \textcolor{purple}{Relation Extraction} & ChemProt, SciERC \\ 

&    & \textcolor{cyan}{Dependency Parsing}  & GENIA-LAS, GENIA-UAS\\

\midrule
 \multirow{1}{*}{13} &    \multirow{1}{*}{Clinical XLNet} &  \textcolor{black}{Prolonged Mechanical Ventilation  Prediction}   & \multirow{1}{*}{\underline{PMV}, \underline{Mortality} \hl{(All)}}   \\

\midrule
14  &   SciGPT2 &  \textcolor{black}{Relationship Explanation Task} &     Relationship Explanation Task \\

\midrule
15  &   NukeBERT  &  \textcolor{orange}{Question Answering} &     NQUAD \hl{(All)} \\

\midrule
\multirow{2}{*}{16}  &   \multirow{2}{*}{GreenBioBERT}  &  \multirow{2}{*}{\textcolor{blue}{Named Entity Recognition}} & \underline{BC5CDR-chemical}, \underline{BC4CHEMD}, \underline{Species-800},  BC5CDR-disease, NCBI-disease, BC2GM, JNLPBA, LINNAEUS   \\

 \midrule
 \multirow{4}{*}{17} &    \multirow{4}{*}{SPECTER} &  \textcolor{teal}{Text Classification} &  \underline{MeSH}, \underline{MAG} \\ 

 & & \textcolor{NavyBlue}{Citation Prediction} &  \underline{Co-Cite}, Cite \\ 

 & & \textcolor{RubineRed}{User Activity Prediction} & \underline{Co-Views}, \underline{Co-Reads} \\ 

& &  \textcolor{Fuchsia}{Paper Recommendation Task} &  \underline{Paper Recommendation Task} \\

\midrule
18  &   BERT-XML  &  \textcolor{black}{ICD Classification} &  \underline{ICD Classification}  \hl{All} \\

\midrule
\multirow{3}{*}{19} & \multirow{3}{*}{Bio-ELECTRA} & \textcolor{blue}{Named Entity Recognition} & BC2GM, BC4CHEMD, NCBI-disease, LINNAEUS \\

  &  & \textcolor{purple}{Relation Extraction} & GAD, ChemProt  \\
 
&    & \textcolor{orange}{Question Answering} & \underline{BioASQ 8b-yes/no}, \underline{BioASQ 5b based}, BioASQ 8b-factoid \\

\midrule
 \multirow{2}{*}{20} &    \multirow{2}{*}{Med-BERT} &  \multirow{2}{*}{\textcolor{BlueViolet}{Disease Prediction}}  &  \underline{Heart failure in diabetes patients} (DHF-Cerner), \underline{Pancreatic cancer} (PaCa-Cerner, PaCa-Truven) \hl{(All)}   \\

\midrule
 \multirow{5}{*}{21} &    \multirow{5}{*}{ouBioBERT} &  \textcolor{blue}{Named Entity Recognition}  &     \underline{BC5CDR-disease}, BC5CDR-chemical, ShARe/CLEF \\

 & & \textcolor{purple}{Relation Extraction}  & \underline{DDI}, \underline{ChemProt}, i2b2 2010\\ 

& & \textcolor{Green}{Sentence Similarity}  & \underline{BIOSSES}, MedSTS\\ 

& & \textcolor{brown}{Document Multi-label Classification}  & HOC \\ 

& & \textcolor{Bittersweet}{Natural Language Inference} &  MedNLI \\

\midrule
 \multirow{6}{*}{22} &    \multirow{6}{*}{PubMedBERT} &  \textcolor{blue}{Named Entity Recognition} &  \underline{BC5CDR-disease}, \underline{BC5CDR-chemical}, \underline{JNLPBA}, \underline{BC2GM}, NCBI-disease  \\

  & & \textcolor{purple}{Relation Extraction} & \underline{ChemProt}, \underline{DDI}, \underline{GAD} \\ 

& & \textcolor{magenta}{PICO Extraction} &  \underline{EBM PICO}\\

& & \textcolor{Green}{Sentence Similarity} & \underline{BIOSSES} \\

& & \textcolor{brown}{Document Multi-label Classification}  & \underline{HOC} \\

& & \textcolor{orange}{Question Answering}  & \underline{BioASQ}, PubMedQA \\

\midrule
 \multirow{6}{*}{23} &    \multirow{6}{*}{MC-BERT} &  \textcolor{blue}{Named Entity Recognition} &  \underline{cEHRNER}, \underline{cMedQANER}  \\

& & \textcolor{orange}{Question Answering}  & \underline{cMedQNLI}, \underline{cMeQA}\\

&    & \textcolor{teal}{Text Classification} & \underline{cMedTC} \\ 

& & \textcolor{black}{Intent Classification} &  \underline{cMedIC}\\

& & \textcolor{black}{Paraphrase Identification} & \underline{cMedQQ} \\

& & \textcolor{RawSienna}{Information Retrieval}  & \underline{cMedIR} \hl{(All)} \\

\midrule
\multirow{7}{*}{24}  &   \multirow{7}{*}{BioALBERT}  &  \multirow{2}{*}{\textcolor{blue}{Named Entity Recognition}} &  \underline{NCBI-disease}, \underline{BC5CDR}, \underline{BC4CHEMD}, \underline{BC2GM}, \underline{JNLPBA}, \underline{LNNAEUS}, \underline{Species-800}, \underline{Share/Clefe} \\

  & & \textcolor{purple}{Relation Extraction} & \underline{ChemProt}, \underline{DDI}, \underline{i2b2}, GAD, Euadr \\ 

& & \textcolor{Green}{Sentence Similarity} & \underline{BIOSSES}, \underline{MedSTS} \\

& & \textcolor{brown}{Document Multi-label Classification}  & \underline{HOC} \\

& & \textcolor{orange}{Question Answering}  & \underline{BioASQ} \\

 & & \textcolor{Bittersweet}{Natural Language Inference} & MedNLI \\

\midrule
25  &   BRLTM  &  \textcolor{black}{Depression Prediction} &  \underline{Depression Prediction}  \hl{(All)} \\

\midrule
\multirow{3}{*}{26} & \multirow{3}{*}{BioMegatron} & \textcolor{blue}{Named Entity Recognition} &  \underline{BC5CDR-chemical}, \underline{BC5CDR-disease}, \underline{NCBI-disease} \\

  &  & \textcolor{purple}{Relation Extraction} &  \underline{ChemProt} \\
 
&    & \textcolor{orange}{Question Answering} & \underline{BioASQ 7b-factoid} \hl{(All)} \\

\midrule
 \multirow{4}{*}{27} &    \multirow{4}{*}{CharacterBERT} &  \textcolor{blue}{Named Entity Recognition}  &   \underline{i2b2 2010}   \\

 & & \textcolor{purple}{Relation Extraction}  & \underline{DDI}, \underline{ChemProt} \\ 

& & \textcolor{Green}{Sentence Similarity}  & ClinicalSTS \\ 

& & \textcolor{Bittersweet}{Natural Language Inference} &  \underline{MedNLI} \\

\midrule
28  &   ChemBERTa & MoleculeNet Task  &  BBBP, ClinTox, HIV, Tox21  \\

\midrule
29  &   ClinicalTransformer  &  \textcolor{blue}{Named Entity Recognition} & \underline{i2b2 2010}, \underline{i2b2 2012}, \underline{n2c2 2018}  \hl{(All)}  \\

\midrule
\multirow{2}{*}{30}  &   \multirow{2}{*}{SapBERT}  &  \multirow{2}{*}{\textcolor{Apricot}{Entity Linking}} & \underline{NCBI-disease}, \underline{BC5CDR-disease}, \underline{BC5CDR-chemical}, \underline{MedMentions}, \underline{AskAPatient}, \underline{COMETA}  \hl{(All)} \\

\midrule
\multirow{3}{*}{31} & \multirow{3}{*}{UmlsBERT} & \textcolor{blue}{Named Entity Recognition} &  \underline{i2b2 2010}, \underline{i2b2 2012} \\

 & & \textcolor{Bittersweet}{Natural Language Inference} & \underline{MedNLI} \\
 
 & & \textcolor{SeaGreen}{De-Identification} & \underline{i2b2 2006}, i2b2 2014 \\

\midrule
 \multirow{1}{*}{32} &    \multirow{1}{*}{bert-for-radiology} &  \textcolor{teal}{Text Classification}  &   \underline{Binary Classifier}, \underline{Classification of CT Reports}, Multi-label Classification  \\



\midrule
 \multirow{6}{*}{33} &    \multirow{6}{*}{Bio-LM} &  \multirow{2}{*}{\textcolor{blue}{Named Entity Recognition}} & \underline{BC5CDR-chemical}, \underline{BC5CDR-disease}, \underline{JNLPBA}, \underline{NCBI-disease}, \underline{BC4CHEMD}, \underline{BC2GM}, \underline{LINNAEUS}, \underline{Species-800}, \underline{i2b2-2010}, \underline{i2b2-2012}  \\

 & & \textcolor{purple}{Relation Extraction} & \underline{ChemProt}, \underline{DDI}, \underline{i2b2 2010}, GAD, EU-ADR \\ 

 & & \textcolor{SeaGreen}{De-Identification} & \underline{i2b2 2014} \\ 
 
 & & \textcolor{brown}{Document Multi-label Classification}  & HOC\\
 & & \textcolor{Bittersweet}{Natural Language Inference} & \underline{MedNLI} \\

\midrule
 \multirow{3}{*}{34} &    \multirow{3}{*}{CODER} &  \textcolor{black}{Term Normalization}  &   \underline{Cadec}, \underline{PsyTar}, MANTRA  \\

 & & \textcolor{black}{Medical Embeddings}  & \underline{MCSM} \\ 

&    & \textcolor{purple}{Relation Extraction} & \underline{DDBRC} (\textit{CODER-Eng version}) \\

\midrule
 \multirow{2}{*}{35} &    \multirow{2}{*}{exBERT} &  \textcolor{blue}{Named Entity Recognition} & \underline{MTL-Bioinformatics-2016}  \\

  & & \textcolor{purple}{Relation Extraction} & \underline{MTL-Bioinformatics-2016}  \hl{(All)} \\

\midrule
\multirow{3}{*}{36} & \multirow{3}{*}{BioMedBERT} & \textcolor{blue}{Named Entity Recognition} &  NCBI-disease, BC5CDR-chemical, BC5CDR-disease, BC4CHEMD, BC2GM, JNLPBA \\

  &  & \textcolor{purple}{Relation Extraction} &  GAD, EU-ADR \\
 
&    & \textcolor{orange}{Question Answering} & \underline{BioASQ 5b-factoid}, \underline{BioASQ 6b-factoid}, \underline{BioASQ 7b-factoid}, BioASQ 4b-factoid \\

\midrule
 \multirow{2}{*}{37} &    \multirow{2}{*}{LBERT} &  \multirow{2}{*}{\textcolor{purple}{Relation Extraction}} & \underline{HPDR50}, \underline{IEPA}, \underline{EU-ADR}, \underline{MedLine}, \underline{BioNLP Shared Task 2011 corpus}, ChemProt, AIMed, BioInfer, GAD, LLL  \\


\midrule
\multirow{8}{*}{38} &    \multirow{8}{*}{OAG-BERT} & \textcolor{Sepia}{Author Name Disambiguation}   &  \underline{whoiswho-v1} \\

 &     & \textcolor{RawSienna}{Scientific Literature Retrieval}  & \underline{OAG-QA} \\

 &     &  \textcolor{Fuchsia}{Paper Recommendation}  & \underline{SciDocs (Recommendation task)} \\

 &     & \textcolor{RubineRed}{User Activity Prediction}   & \underline{Co-Views}, \underline{Co-Reads} \\

 &     &  Entity Graph Completion  & \underline{CS Dataset} \\

 &     &  Fields-of-study Tagging  & \underline{MAG} \\

 &     &  Venue Prediction  &  \underline{arXiv} \\

 &     &  Affiliation Prediction  &  \underline{arXiv} \hl{(All)}  \\

\midrule
\multirow{2}{*}{39} & \multirow{2}{*}{CovidBERT} & \textcolor{purple}{Relation Extraction} &  \underline{BioCreative V} \\

  &  & \textcolor{teal}{Text Classification}  & \underline{DisGeNet Database} \hl{(All)} \\

\midrule
\multirow{3}{*}{40} & \multirow{3}{*}{ELECTRAMed} & \textcolor{blue}{Named Entity Recognition} &\underline{BC5CDR},  NCBI-disease, JNLPBA \\

  &  & \textcolor{purple}{Relation Extraction} &  ChemProt, DDI \\
 
&    & \textcolor{orange}{Question Answering} & \underline{BioASQ 7b-factoid} \\

\midrule
\multirow{2}{*}{41} & \multirow{2}{*}{KeBioLM} & \textcolor{blue}{Named Entity Recognition} &  \underline{NCBI-disease}, \underline{BC5CDR-chemical}, \underline{BC2GM}, \underline{JNLPBA}, BC5CDR-disease \\

  &  & \textcolor{purple}{Relation Extraction} &  \underline{GAD}, \underline{ChemProt}, DDI \\

\midrule
 \multirow{4}{*}{42} &    \multirow{4}{*}{SINA-BERT} & Fill-in-the-Blank  & \underline{10k-sentences}  \\

&    &  \textcolor{teal}{Text Classification}  &  \underline{4k-samples}  \\ 

&    &  \textcolor{OrangeRed}{Medical Sentiment Analysis}  &  \underline{5k-comments}  \\ 

&    &  \textcolor{RawSienna}{Medical Question Retrieval}  &  \underline{200k-pairs}  \hl{(All)}  \\

\midrule
\multirow{3}{*}{43} & \multirow{3}{*}{MathBERT (P)} &  \textcolor{RawSienna}{Mathematical Information Retrieval}  &   \underline{NTCIR-12 MathIR} \\

  &  &  Formula Topic Classification  &  \underline{TopicMath-100K}  \\
 
&    &  \textcolor{WildStrawberry}{Formula Headline Generation}   &  \underline{EXEQ-300K} \hl{(All)}  \\

\midrule
 \multirow{2}{*}{44} &    \multirow{2}{*}{NukeLM} & OSTI Multi-class Prediction  &  \underline{OSTI subject categories}  \\

&    & OSTI Binary Prediction & \underline{OSTI custom} \hl{(All)}   \\

\midrule
 \multirow{3}{*}{45} &    \multirow{3}{*}{ProteinBERT} & Protein Structure   &  Secondary Structure, Disorder, Remote Homology, Fold Classes  \\

&    & Post-translational Modifications  &  Neuropeptide Cleavage, Major PTMs, Signal Peptide  \\

&    &   Biophysical Properties &   \underline{Stability}, Fluorescence \\

\midrule
 \multirow{6}{*}{46} &    \multirow{6}{*}{SciFive} &  \multirow{2}{*}{\textcolor{blue}{Named Entity Recognition}} &\underline{JNLPBA},  \underline{Species-800}, \underline{BC5CDR-disease}, BC5CDR-chemical, NCBI-disease, BC4CHEMD, BC2GM  \\

  & & \textcolor{purple}{Relation Extraction} & \underline{ChemProt}, \underline{DDI} \\ 

 & & \textcolor{Bittersweet}{Natural Language Inference} &  \underline{MedNLI} \\

& & \textcolor{brown}{Document Multi-label Classification}  & HOC\\

& & \textcolor{orange}{Question Answering}  & \underline{BioASQ 4b-factoid}, \underline{BioASQ 5b-factoid}, \underline{BioASQ 6b-factoid} \\

\midrule
 \multirow{7}{*}{47} &    \multirow{7}{*}{BioELECTRA} &  \textcolor{blue}{Named Entity Recognition} & \underline{BC5CDR-chemical}, \underline{BC5CDR-disease}, \underline{JNLPBA}, \underline{NCBI-disease}, \underline{BC2GM}, \underline{ShARe/CLEFE}  \\

  & & \textcolor{purple}{Relation Extraction} & \underline{ChemProt}, \underline{DDI}, \underline{GAD}, \underline{i2b2-2010} \\ 

& & \textcolor{magenta}{PICO Extraction} &  \underline{EBM-NLP}\\

& & \textcolor{Green}{Sentence Similarity} & \underline{BIOSSES}, \underline{ClinicalSTS} \\

& & \textcolor{brown}{Document Multi-label Classification}  & \underline{HOC}\\

& & \textcolor{orange}{Question Answering}  & \underline{PubMedQA}, \underline{BioASQ} \\

 & & \textcolor{Bittersweet}{Natural Language Inference} &  \underline{MedNLI} \hl{(All)} \\

\midrule
 \multirow{2}{*}{48} &    \multirow{2}{*}{ChemBERT} & Reaction Role Labeling   &    \underline{Reaction Role Labeling} \\

&    & Product Extraction  &  \underline{Product Extraction}  \hl{(All)} \\

\midrule
 \multirow{4}{*}{49} &    \multirow{4}{*}{EntityBERT} &  \textcolor{purple}{Temporal Relation Extraction}  &   \underline{THYME+}  \\

&    &  Document Time Relation Classification & \underline{THYME+}  \\

&    &  Negation Detection & i2b2, MiPACQ, Seed, Strat  \\

&    &  \textcolor{orange}{Question Answering} & PubMedQA  \\

\midrule

\multirow{3}{*}{50} &    \multirow{3}{*}{MathBERT (S)} &  Knowledge Component Prediction  &  \underline{ASSISTments} \\

  &   & Auto-grading  &  \underline{ASSISTments} \\

  &   & Knowledge Tracing Correctness Prediction  &  \underline{ASSISTments} \hl{(All)} \\

\midrule
51  &   MedGPT & Next Disorder Prediction  &  \underline{MIMIC-III}, \underline{KTH}  \hl{(All)} \\

\midrule
 \multirow{6}{*}{52} &    \multirow{6}{*}{SMedBERT} &  \textcolor{blue}{Named Entity Recognition} &  \underline{cMedQANER}, \underline{DXY-NER} \\

  & & \textcolor{purple}{Relation Extraction} & \underline{DXY-RE}, \underline{CHIP-RE} \\ 

& & \textcolor{magenta}{Question Matching} &  \underline{cMedQQ}\\

& & \textcolor{Mahogany}{Intrinsic Evaluation}  & \underline{D1}, \underline{D2}, \underline{D3} \\

& & \textcolor{orange}{Question Answering}  & \underline{cMedQA}, \underline{WebMedQA} \\

 & & \textcolor{Bittersweet}{Natural Language Inference} &  \underline{cMedQNLI} \hl{(All)} \\

\midrule
 53 &  Bio-cli   &  \textcolor{blue}{Named Entity Recognition}  &  \underline{PharmaCoNER}, CANTEMIST, \underline{ICTUSnet} \\

\midrule
 \multirow{3}{*}{54} &    \multirow{3}{*}{MatSciBERT} & \textcolor{blue}{Named Entity Recognition}  &  \underline{SOFC}, \underline{SOFC-Slot}, \underline{Matscholar NER} \\

  & & \textcolor{teal}{Text Classification}  &  \underline{Glass vs. Non-glass Dataset} \\ 

  & & \textcolor{purple}{Relation Extraction}  &  \underline{Materials Synthesis Procedures Dataset} \hl{(All)} \\

\midrule
\multirow{2}{*}{55}  &   \multirow{2}{*}{ClimateBert} &  \textcolor{teal}{Text Classification}  &  \underline{Climate-related Classification} \\ 

&    &  \textcolor{OrangeRed}{Sentiment Analysis}  &  \underline{Climate-related Paragraphs} \hl{(All)} \\

\midrule
\multirow{2}{*}{56}  &   \multirow{2}{*}{MatBERT} &  \multirow{2}{*}{\textcolor{blue}{Named Entity Recognition}}  &  \underline{Solid-state Materials Abstracts}, \underline{Inorganic Doping Abstracts}, \underline{Gold Nanoparticle Synthesis}  \hl{(All)} \\

\midrule
57  &   UTH-BERT   &  MedWeb Task (8 labels)  &  \underline{MedWeb} \hl{(All)} \\

\midrule
58  &  ChestXRayBERT  &  \textcolor{Plum}{Summarization}  &  \underline{OPEN-I}, \underline{MIMIC-CXR} \hl{(All)} \\

\midrule
 \multirow{3}{*}{59} &    \multirow{3}{*}{MedRoBERTa.nl} &  Medical Odd-one-out Similarity Task & \underline{Medical odd-one-out Similarity Task}  \\

  & & ICF Classification Task & ICF Classification Task \\ 

  & & \textcolor{blue}{Named Entity Recognition} (Dutch) & CoNLL 2002 \\

\midrule
 \multirow{6}{*}{60} &    \multirow{6}{*}{PubMedELECTRA} &  \textcolor{blue}{Named Entity Recognition} &  BC5CDR-chemical, JNLPBA, NCBI-disease, BC5CDR-disease, BC2GM \\

  & & \textcolor{purple}{Relation Extraction} &  \underline{GAD}, ChemProt, DDI  \\ 

& & \textcolor{magenta}{PICO Extraction} &  \underline{EBM-NLP} \\

& & \textcolor{Green}{Sentence Similarity} &  BIOSSES \\

& & \textcolor{brown}{Document Multi-label Classification}  & HOC  \\

& & \textcolor{orange}{Question Answering}  &  PubMedQA, BioASQ \\

\midrule
 \multirow{4}{*}{61} &      \multirow{4}{*}{Clinical-Longformer} &  \textcolor{blue}{Named Entity Recognition} & \underline{i2b2 2006}, \underline{i2b2 2010}, \underline{i2b2 2012}, \underline{i2b2 2014}  \\

& & \textcolor{teal}{Text Classification}  & \underline{MIMIC-AKI}, \underline{OpenI}, OHSUMed\\

& & \textcolor{orange}{Question Answering}  & \underline{emrQA} \\

 & & \textcolor{Bittersweet}{Natural Language Inference} &  \underline{MedNLI} \\





\midrule
 \multirow{6}{*}{62} &    \multirow{6}{*}{BioLinkBERT} &  \textcolor{blue}{Named Entity Recognition} &  \underline{BC5CDR-chemical},  \underline{BC5CDR-disease}, \underline{JNLPBA}, \underline{NCBI-disease}, \underline{BC2GM} \\

  & & \textcolor{purple}{Relation Extraction} &  \underline{GAD}, \underline{ChemProt}, \underline{DDI}  \\ 

& & \textcolor{magenta}{PICO Extraction} &  \underline{EBM-NLP} \\

& & \textcolor{Green}{Sentence Similarity} &  \underline{BIOSSES} \\

& & \textcolor{brown}{Document Multi-label Classification}  & \underline{HOC}  \\

& & \textcolor{orange}{Question Answering}  &  \underline{PubMedQA}, \underline{BioASQ} \hl{(All)} \\

\midrule
 \multirow{4}{*}{63} &    \multirow{4}{*}{BioBART} &  \textcolor{blue}{Named Entity Recognition} & ShARe13, ShARe14, CADEC, GENIA \\

  & & \textcolor{Apricot}{Entity Linking} &  \underline{BC5CDR}, \underline{AskAPatient}, \underline{COMETA}, MedMentions, NCBI-disease  \\ 

& & \textcolor{Plum}{Summarization} &  \underline{MeQSum}, \underline{MEDIQA-ANS}, iCliniq, HealthCareMagic,  MEDIQA-QS, MEDIQA-MAS \\

& & \textcolor{RoyalBlue}{Dialogue} & \underline{CovidDialog} \\

\midrule
\multirow{2}{*}{64} &    \multirow{2}{*}{SecureBERT} &  \textcolor{blue}{Named Entity Recognition} & \underline{MalwareTextDB}  \\


  &    &  \textcolor{OrangeRed}{Sentiment Analysis} &  Rotten Tomatoes  \\

\midrule
 \multirow{2}{*}{65} &    \multirow{2}{*}{BatteryBERT} &  \textcolor{orange}{Question Answering} & Retional Data Extraction, Device Component Classification \\

& & \textcolor{teal}{Text Classification} & Battery Abstract Classification \\

\midrule
66  &   bsc-bio-ehr-es  &  \textcolor{blue}{Named Entity Recognition} &  \underline{PharmaCoNER}, \underline{CANTEMIST}, \underline{ICTUSnet} \hl{(All)} \\

\midrule
67  &   ChemGPT  &  N/A  &  N/A  \\



\midrule
 \multirow{1}{*}{68} &    \multirow{1}{*}{PathologyBERT} &  Cancer Severity Classification  & \underline{Breast Cancer Diagnose Severity Classification}  \hl{(All)} \\

\midrule
 \multirow{3}{*}{69} &    \multirow{3}{*}{ScholarBERT} &   \multirow{2}{*}{\textcolor{blue}{Named Entity Recognition}} & \underline{ScienceExam}, SciERC, NCBI-disease, 
ChemDNER, Matscholar NER, BC5CDR, JNLPBA,  Coleridge \\

  & & \textcolor{purple}{Relation Extraction} & \underline{SciERC}, ChemProt, Paper Field \\

\midrule
 \multirow{3}{*}{70} &    \multirow{3}{*}{RadBERT} &   \textcolor{teal}{Text Classification} & \underline{Abnormal Sentence Classification} \\

  & & Report Coding & \underline{AAA}, \underline{BI-RADS}, \underline{Lung-RADS}, \underline{Abnormal}, \underline{Alert} \\ 

  & & \textcolor{Plum}{Summarization} &  \underline{Summarization} \hl{(All)} \\

\midrule
 \multirow{3}{*}{71} &    \multirow{3}{*}{SciDeBERTa} &   \textcolor{blue}{Named Entity Recognition} & \underline{SciERC}, \underline{GENIA} \\

  & & \textcolor{purple}{Relation Extraction} & \underline{SciERC} \\ 

  & & Co-reference Resolution & \underline{GENIA}, SciERC  \\

\midrule
 \multirow{4}{*}{72} &    \multirow{4}{*}{SsciBERT} &  \textcolor{blue}{Named Entity Recognition}  & \underline{Scientometrics} \\

  & & Identification Results of Abstract Structures & \underline{Identification Results of Abstract Structures} \\ 

  & & Discipline Classification & \underline{JCR Social Science} \hl{(All)} \\

\midrule  
\multirow{3}{*}{73} &    \multirow{3}{*}{ViHealthBERT} &  \textcolor{blue}{Named Entity Recognition}   &  \underline{COVID-19}, \underline{ViMQ} \\

  & & \textcolor{Sepia}{Acronym Disambiguation Task} &  \underline{acrDrAid}  \\ 

  & & \textcolor{Plum}{Summarization} &  \underline{FAQ Summarization} \hl{(All)} \\

\midrule
74  &   Clinical Flair  & \textcolor{blue}{Named Entity Recognition}  &   \underline{CANTEMIST}, \underline{Clinical Trials}, \underline{NUBes}, PharmaCoNER  \\

\midrule
 \multirow{2}{*}{75} &    \multirow{2}{*}{KM-BERT} &  \textcolor{blue}{Named Entity Recognition}  & \underline{Korean NER} \\

  & & \textcolor{MidnightBlue}{Semantic Textual Similarity}  & \underline{Korean MedSTS} \hl{(All)} \\

\midrule
76  &   MaterialsBERT (S) & \textcolor{blue}{Named Entity Recognition}  & \underline{A polymer dataset annotated by the authors} \hl{(All)} \\

\midrule
77  &   ProcessBERT & \textcolor{teal}{Text Classification}    &    \underline{Phrases manually annotated dataset} \hl{(All)} \\

\midrule
 \multirow{4}{*}{78} &    \multirow{4}{*}{BioGPT} &  \textcolor{WildStrawberry}{Generation} & Self-created Text Generation Dataset \\

  & & \textcolor{purple}{Relation Extraction} & \underline{BC5CDR}, \underline{KD-DTI}, \underline{DDI} \\ 

& & \textcolor{brown}{Document Multi-label Classification}  & \underline{HOC} \\

& & \textcolor{orange}{Question Answering}  & \underline{PubMedQA}\\

\midrule
 \multirow{2}{*}{79} &    \multirow{2}{*}{ChemBERTa-2} &  Classification  & \underline{BBBP}, \underline{SR-p53}, BACE,  ClinTox \\

  & & Regression & \underline{Clearance}, \underline{Delaney}, \underline{Lipo}, BACE \\

\midrule
 \multirow{3}{*}{80} &    \multirow{3}{*}{CSL-T5} &  \textcolor{Plum}{Summarization} &  \underline{Predict Title from Abstract} (Created by authors) \\

  & & \textcolor{WildStrawberry}{Generation} & \underline{Chinese Keyword Generation Task} (Created by authors) \\ 

& & \textcolor{teal}{Text Classification}  & \underline{CTG}, \underline{DCP} (Created by authors) \hl{(All)}  \\

\midrule
 \multirow{2}{*}{81} &    \multirow{2}{*}{MaterialBERT (Y)} &  Visualization of Materials  & Visualization of Materials \\

  & & \textcolor{teal}{Text Classification} & CoLA \\

 \midrule
\multirow{1}{*}{82} &    \multirow{1}{*}{AcademicRoBERTa} &  \textcolor{teal}{Text Classification}  &  \underline{Author Identification}, \underline{Category Classification} \hl{(All)} \\


 \midrule
 83 &  Bioberturk   &  \textcolor{teal}{Radiology Report classification}  &  \underline{Impressions}, \underline{Findings} \hl{(All)}\\

\midrule
84  &   DRAGON & \textcolor{orange}{Question Answering}   &    \underline{MedQA}, \underline{PubMedQA}, \underline{BioASQ} \hl{(All)} \\

\midrule
 \multirow{5}{*}{85} &    \multirow{5}{*}{UCSF-BERT} &  \textcolor{blue}{Named Entity Recognition} & \underline{i2b2 2010}, i2b2 2012   \\

  & & \textcolor{purple}{Relation Extraction} & \underline{i2b2 2010} \\ 

& & \textcolor{Bittersweet}{Natural Language Inference} & MedNLI \\

& & ICD9-top50 Coding & \underline{ICD9-top50 Coding} \\

& & Therapeutic Class Prediction  & \underline{Therapeutic Class Prediction} \\

\midrule  
\multirow{3}{*}{86} &    \multirow{3}{*}{ViPubmedT5} &  \textcolor{Bittersweet}{Natural Language Inference}   &   \underline{ViMedNLI} \\

  & & \textcolor{Sepia}{Acronym Disambiguation Task} &  \underline{acrDrAid}  \\ 

  & & \textcolor{Plum}{Summarization}  &  FAQ Summarization \\

\midrule
 \multirow{9}{*}{87} &    \multirow{9}{*}{Galactica} & General Capabilities  & \underline{57 task selection from BIG-bench} \\

  &    &  Downstream Scientific NLP  &  \underline{MMLU benchmark}, \underline{other popular scientific QA benchmarks}  \\

  &    &  Citation Prediction  &  \underline{Citation Accuracy} (PWC Citations, Extended Citations, Contextual Citations)  \\ 

  &    &  Chemical Understanding  &  \underline{IUPAC Name Prediction}, MoleculeNet  \\ 
  
  &    &  Toxicity and Bias  &  \underline{Bias and Stereotypes} (CrowS-Pairs, StereoSet, Toxicity), \underline{TruthfulQA}  \\

 \cmidrule{3-4}
  &    &  \multirow{2}{*}{Knowledge Probes}  &  \underline{LaTeX Equations}, Domain Probes (\underline{AminoProbe}, BioLAMA, \underline{Chemical Reactions}, \underline{Galaxy Clusters}, \underline{Mineral Groups}), Reasoning (\underline{MMLU Mathematics}, MATH)  \\

 \cmidrule{3-4}
  &    &  \multirow{2}{*}{Biological Understanding}  &  \underline{Sequence Validation Perplexity}, \underline{Protein Keyword Prediction}, \underline{Protein Function Description}  \\ 

\midrule
 \multirow{4}{*}{88} &    \multirow{4}{*}{VarMAE} &  \textcolor{blue}{Named Entity Recognition} & \underline{JNLPBA}, \underline{IEE} \\

  & & \textcolor{teal}{Text Classification} & \underline{ACL-ARC}, \underline{SciCite}, \underline{OIR}, \underline{MTC}\\ 

& & \textcolor{magenta}{PICO Extraction} &  \underline{EBM-NLP} \\

& & Text Matching & \underline{PSM} \hl{(All)} \\

\midrule
89  &   AliBERT  & \textcolor{blue}{Named Entity Recognition}  &  \underline{BioNER}, \underline{QUAERO}  \hl{(All)} \\

\midrule
 \multirow{2}{*}{90} &    \multirow{2}{*}{BioMedLM} &  \textcolor{WildStrawberry}{Text Generation} & MeQSum \\
& & \textcolor{orange}{Question Answering}  & \underline{MedQA}, PubMedQA, and BioASQ \\

\midrule
 \multirow{6}{*}{91} &    \multirow{6}{*}{BioReader} &  \multirow{2}{*}{\textcolor{blue}{Named Entity Recognition}} & \underline{BC4CHEMD}, \underline{Species-800}, JNLPBA, NCBI-disease, BC5CDR-disease, BC5CDR-chemical, 
BC2GM  \\

  & & \textcolor{purple}{Relation Extraction} & ChemProt, \underline{DDI} \\ 

& & \textcolor{brown}{Document Multi-label Classification}  & \underline{HOC} \\

& & \textcolor{orange}{Question Answering}  & BioASQ 4b-factoid, BioASQ 5b-factoid, BioASQ 6b-factoid, MedQA-USMLE  \\

& & \textcolor{Bittersweet}{Natural Language Inference} & MedNLI \\

\midrule
 \multirow{4}{*}{92} &    \multirow{4}{*}{ClinicalT5} &  \textcolor{blue}{Named Entity Recognition} & \underline{NCBI-disease}, BC5CDR-disease  \\

  & & \textcolor{Mahogany}{Intrinsic Evaluation} & UMNSRS-Sim, UMNSRS-Rel \\ 

& & \textcolor{brown}{Document Multi-label Classification}  & \underline{HOC} \\

& & \textcolor{Bittersweet}{Natural Language Inference} & \underline{MedNLI} \\

\midrule   

 \multirow{5}{*}{93} &    \multirow{5}{*}{CySecBERT} &  \textcolor{blue}{Named Entity Recognition} &  \underline{NVD}  \\ 

  &      &  Word Similarity Task  &  \underline{Word Similarity}  \\ 

  &      &   Clustering Task  &  \underline{Log4J dataset}  \\

  &      &  Classification &  \underline{CySecAlert},  \underline{MS Exchange}  \\ 

  &      &  Combined Task &  SuperGLUE  \\

\midrule
 \multirow{5}{*}{94} &    \multirow{5}{*}{GatorTron} &  \textcolor{blue}{Named Entity Recognition} &  \underline{i2b2 2010},  \underline{i2b2 2012}, \underline{n2c2 2018} \\

  & & \textcolor{purple}{Relation Extraction} & \underline{n2c2 2018} \\ 

  & & \textcolor{MidnightBlue}{Semantic Textual Similarity}  & \underline{n2c2 2019} \\ 

& & \textcolor{Bittersweet}{Natural Language Inference} & \underline{MedNLI} \\

  & & \textcolor{orange}{Medical Question Answering} & \underline{emrQA}  \hl{(All)} \\

\midrule
 \multirow{2}{*}{95} &    \multirow{2}{*}{Med-PaLM} &  \multirow{2}{*}{\textcolor{orange}{Question Answering}} & \underline{MedQA-USMLE}, \underline{MedMCQA}, \underline{PubMedQA}, MMLU, LiveQA, MedicationQA, HealthSearchQA  \\

\midrule
96  &   Clinical-T5  & N/A  &  N/A  \\

\midrule
\multirow{1}{*}{97}  &    \multirow{1}{*}{CPT-BigBird}  & \multirow{1}{*}{\textcolor{orange}{Question Answering}}  &  \multirow{1}{*}{SQuAD, IS-QA}  \\

\midrule
 \multirow{3}{*}{98} &    \multirow{3}{*}{ProtST} &  Protein Localization Prediction &  \underline{Bin}, \underline{Sub} \\

& &  Fitness Landscape Prediction  &  \underline{$\beta$-lac}, \underline{AAV}, \underline{Thermo}, \underline{Flu}, \underline{Sta}    \\ 
  
& &   Protein Function Annotation    &   \underline{EC}, \underline{GO-BP}, \underline{GO-MF}, \underline{GO-CC}  \hl{(All)}    \\

\midrule
\multirow{2}{*}{99}  &   \multirow{2}{*}{SciEdBERT}   & Automatic Scoring of Students' Written Responses   &  \multirow{2}{*}{\underline{4T}, \underline{7T}  \hl{(All)}} \\


  

\midrule
 \multirow{4}{*}{100} &    \multirow{4}{*}{Bioformer} &  \multirow{2}{*}{\textcolor{blue}{Named Entity Recognition}} & BC5CDR-disease, NCBI-disease, BC5CDR-chemical, BC4CHEMD, BC2GM, JNLPBA, LINNAEUS, Species-800  \\

& & \textcolor{purple}{Relation Extraction} & ChemProt, DDI, EU-ADR, GAD \\ 
  
& & \textcolor{brown}{Document Multi-label Classification}  & HOC, BioCreative-LitCovid \\

\midrule
 \multirow{4}{*}{101} &    \multirow{4}{*}{\makecell[l]{Lightweight \\ Clinical \\ Transformers}} &  \textcolor{blue}{Named Entity Recognition} &  \underline{i2b2 2010}, i2b2 2012, \underline{i2b2 2014}   \\

& & \textcolor{purple}{Relation Extraction} & \underline{i2b2 2010} \\ 
  
& & \textcolor{Bittersweet}{Natural Language Inference}  & MedNLI \\

& & Sequence Classification  &  \underline{Isaric Clinical Notes (ICN)} \\

\midrule
102  &   ManuBERT   &  \textcolor{blue}{Named Entity Recognition} &  \underline{FabNER}  \\

\midrule
103  &   RAMM   &   \textcolor{orange}{Visual Question Answering}  &  \underline{Med-VQA 2019}, \underline{VQA-Med 2021}, \underline{VQARAD}, \underline{SLAKE}  \hl{(All)} \\

\midrule
 \multirow{5}{*}{104} &    \multirow{5}{*}{DrBERT} & \multirow{2}{*}{\textcolor{blue}{Named Entity Recognition}}   & \underline{MUSCA-DET}, \underline{QUAERO-EMEA}, \underline{QUAERO-MEDLINE}, \underline{Acute heart failure (aHF)}, \underline{Medical Report}   \\

&     &   \textcolor{teal}{Text Classification}   & \underline{aHF}, \underline{MUSCA-DET}, \underline{Technical Specialties Sorting}   \\ 

&     &   POS Tagging & \underline{ESSAIS}, CAS  \\

&     &  \textcolor{orange}{Question Answering}    &  FrenchMedMCQA \\

\midrule
 \multirow{4}{*}{105} &    \multirow{4}{*}{MOTOR} & \textcolor{WildStrawberry}{Medical Report Generation}  &  \underline{IU-Xray} \\

& & \textcolor{RawSienna}{Image Report Retrieval}  &  \underline{MIMIC-CXR}    \\ 
  
& &   Diagnosis Classification    &   \underline{MIMIC-CXR}, ChestX-ray14    \\

& &   \textcolor{orange}{Visual Question Answering}    &    \underline{SLAKE}, VQARAD   \\

\midrule
 \multirow{5}{*}{106} &    \multirow{5}{*}{BiomedGPT} &  Image Classification & \underline{MedMNIST v2}  \\

& & \textcolor{Bittersweet}{Natural Language Inference} &  MedNLI \\ 

& & \textcolor{Plum}{Summarization} &  MeQSum, iCliniq, HealthCareMagic \\ 

& & \textcolor{Emerald}{Image Captioning} & \underline{Peir Gross}, IU X-ray, ROCO \\ 

& & \textcolor{orange}{Visual Question Answering} &  \underline{SLAKE}, \underline{PathVQA}, VQARAD \\

\midrule
 \multirow{4}{*}{107} &    \multirow{4}{*}{Patton} &  Classification  &  PATTON (\underline{Mathematics, Clothes, Sports}), SciPATTON (\underline{Geology, Economics}) \\

& & Retrieval  &  PATTON (\underline{Clothes, Sports}), SciPATTON (\underline{Mathematics, Geology, Economics})    \\ 
  
& &  Reranking     &    PATTON (\underline{Clothes, Sports}), SciPATTON (\underline{Mathematics, Geology, Economics})   \\

& &    Link Prediction   &    SciPATTON (\underline{Mathematics, Geology, Economics})   \\

\midrule
108  &   TurkRadBERT   & \textcolor{teal}{Text Classification}  &  Turkish head CT reports  \\

\midrule
109  &   CamemBERT-bio   & \textcolor{blue}{Named Entity Recognition}  &  \underline{CAS1}, \underline{CAS2}, \underline{QUAERO-EMEA}, QUAERO-MEDLINE, E3C  \\

\midrule
 \multirow{4}{*}{110} &    \multirow{4}{*}{ClinicalGPT} &  \textcolor{RoyalBlue}{Medical Conversation} & \underline{MedDialog}  \\

& & Diagnosis & \underline{MD-EHR}  \\ 

& &  \textcolor{orange}{Medical Question Answering} & \underline{cMedQA2}, MEDQA-MCMLE  \\

\midrule

\multirow{3}{*}{111} &    \multirow{3}{*}{EriBERTa} & \multirow{2}{*}{\textcolor{blue}{Named Entity Recognition}}  &  BC5CDR-disease, BC5CDR-chemical, JNLPBA, NCBI-disease, BC4CHEM, DIANN, Treatment \\ 

& & \textcolor{blue}{Named Entity Recognition} (Spanish)  &   \underline{PharmaCoNER}, \underline{Cantemist-NER}   \\

\midrule
 \multirow{1}{*}{112} &    \multirow{1}{*}{K2} & \textcolor{orange}{Question Answering}  &  \underline{NPEE}, APTest \\



\midrule
 \multirow{3}{*}{113} &    \multirow{3}{*}{PharmBERT} & \multirow{2}{*}{Drug Labeling}  &  \underline{Detection of adverse drug reaction}, \underline{Extraction of drug–drug interaction}, \underline{ADME classification} \\

& &  \textcolor{OrangeRed}{Sentiment Labeling}  &  SST-2    \\

\midrule

 \multirow{2}{*}{114} &    \multirow{2}{*}{BioNART} & \textcolor{Plum}{Summarization}  &  iCliniq, HealthCareMagic \\ 

& &  \textcolor{RoyalBlue}{Clinical Dialogue}  &  CovidDialog    \\

\midrule
115  &   BIOptimus   & \textcolor{blue}{Named Entity Recognition}  &  \underline{BC5-chemical}, \underline{NCBI-disease}, \underline{BC2GM}, BC5-disease, JNLPBA  \\

\midrule
 \multirow{3}{*}{116} &    \multirow{3}{*}{KEBLM} & \textcolor{Apricot}{Entity Linking}  & \underline{COMETA}, \underline{NCBI-disease}, BC5CDR-chemical, BC5CDR-disease  \\

& &  \textcolor{Bittersweet}{Natural Language Inference} &   \underline{MedNLI}   \\ 
  
& &    \textcolor{orange}{Question Answering}   &   \underline{PubMedQA}    \\

\midrule
 \multirow{4}{*}{117} &    \multirow{4}{*}{GIT-Mol} & \textcolor{Emerald}{Captioning}  & \underline{Molecule Captioning}, \underline{Molecule Image Captioning}   \\

& & \textcolor{WildStrawberry}{Molecule Generation}  &  \underline{Molecule Generation}    \\

& &  Molecular Property Prediction     &    \underline{ToxCast}, \underline{Sider}, \underline{ClinTox}, \underline{Bace},  \underline{BBBP}, Tox21   \\

  \bottomrule

\end{xltabular}
}

\subsection{Exploring Task Performance Details}
\label{app_sec_exploring_task_performance}

\begin{figure}[h]
    \centering
    \begin{subfigure}{0.49\linewidth}
        \includegraphics[width=\linewidth]{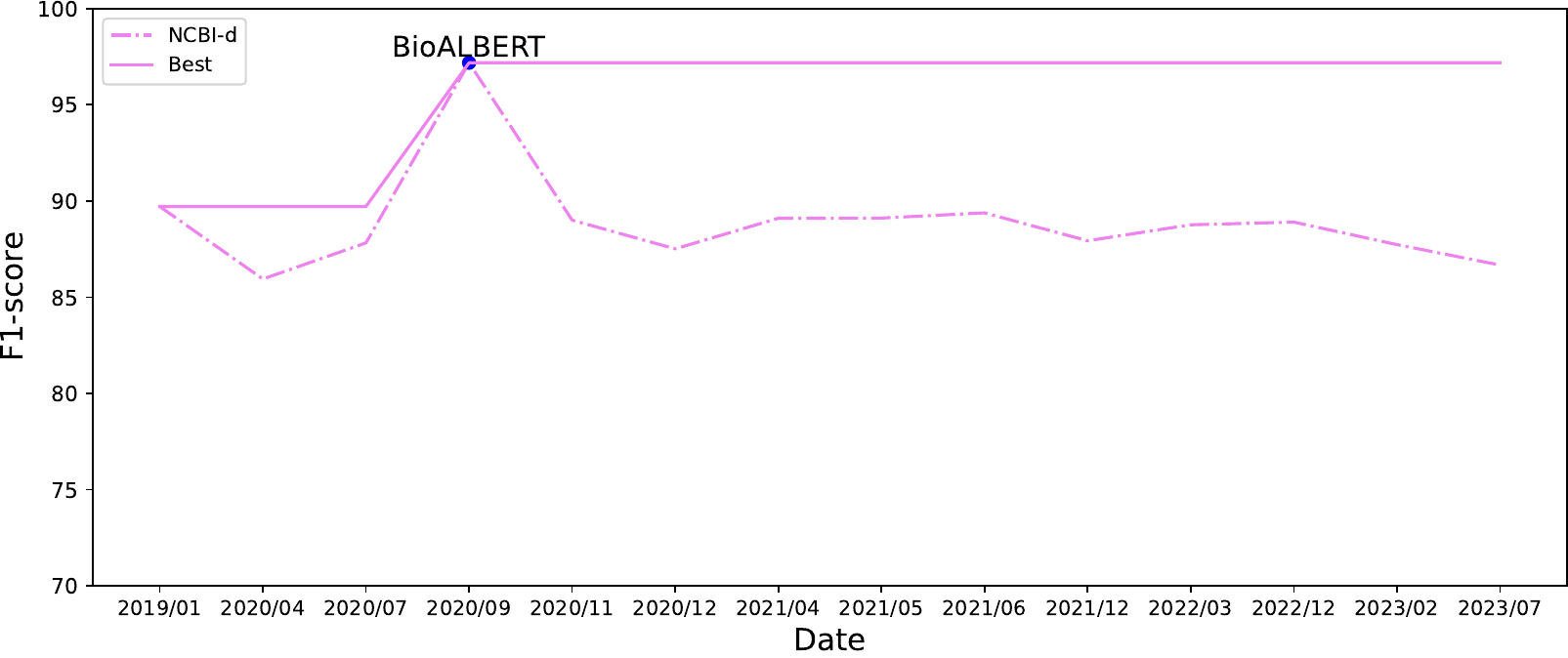}
        \caption{NCBI-disease}
    \end{subfigure}
    \begin{subfigure}{0.49\linewidth}
        \includegraphics[width=\linewidth]{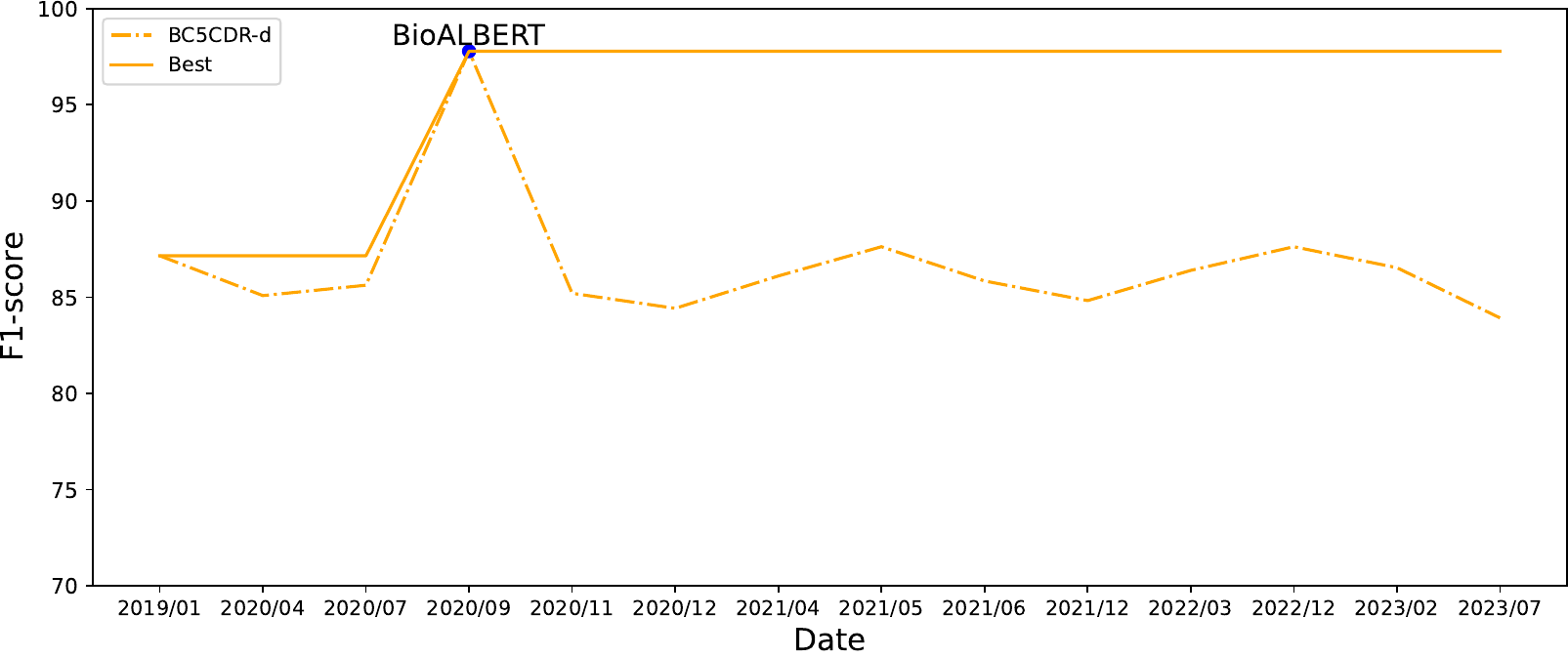}
        \caption{BC5CDR-disease}
    \end{subfigure}
    \begin{subfigure}{0.49\linewidth}
        \includegraphics[width=\linewidth]{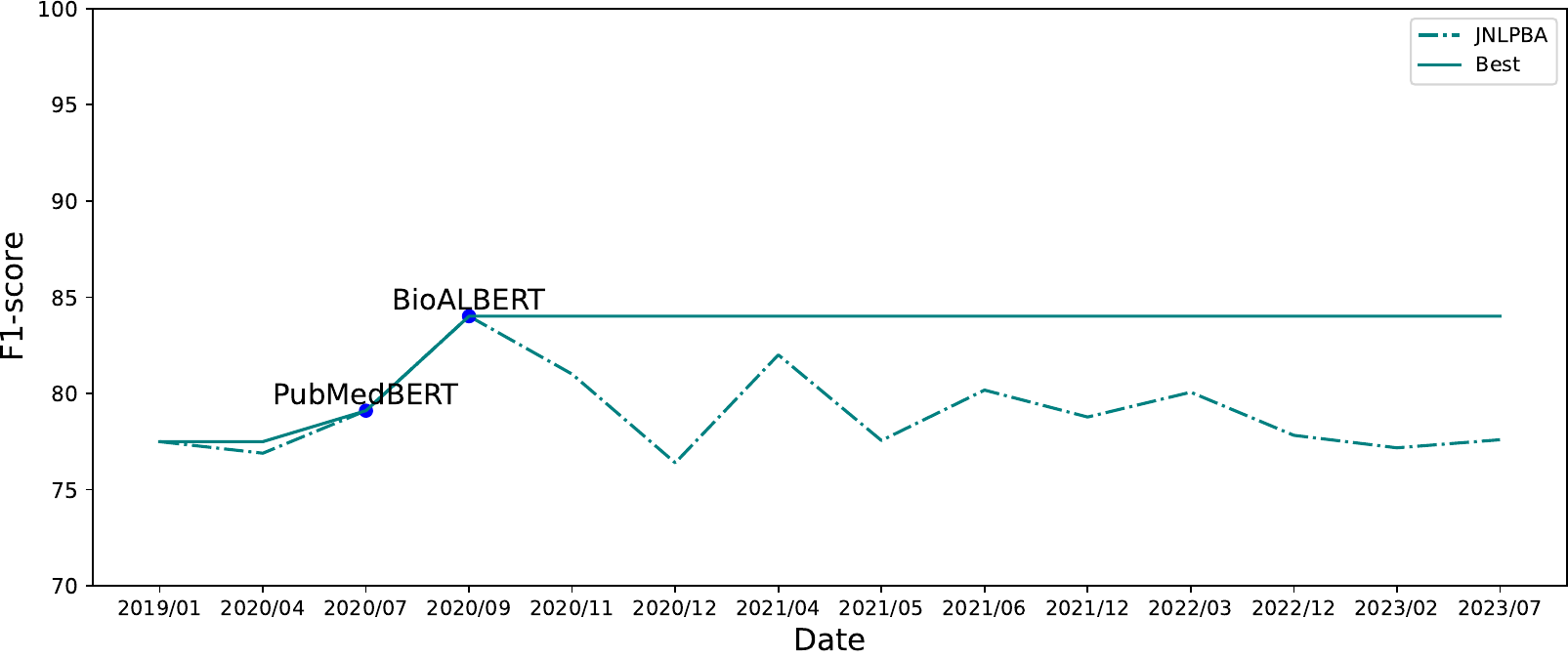}
        \caption{JNLPBA}
    \end{subfigure}
    \begin{subfigure}{0.49\linewidth}
        \includegraphics[width=\linewidth]{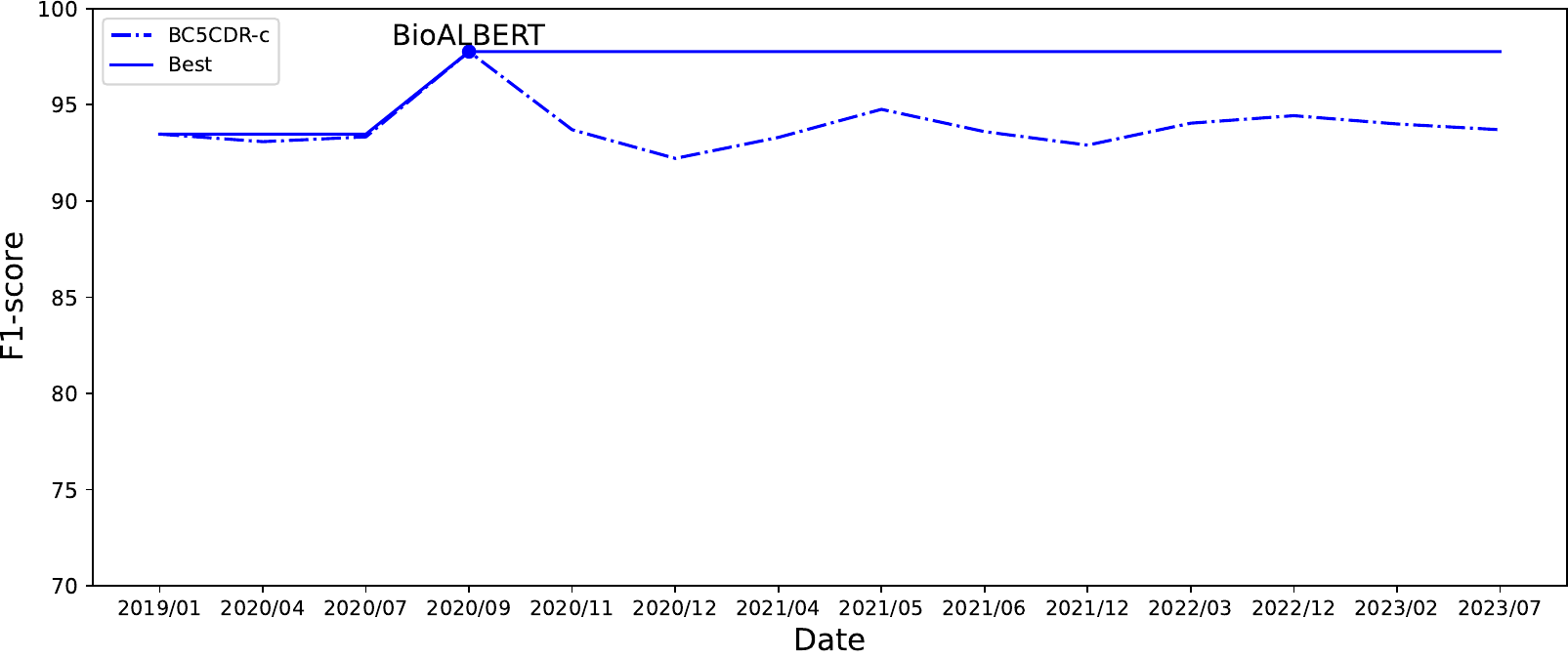}
        \caption{BC5CDR-chemical}
    \end{subfigure}
    \begin{subfigure}{0.49\linewidth}
        \includegraphics[width=\linewidth]{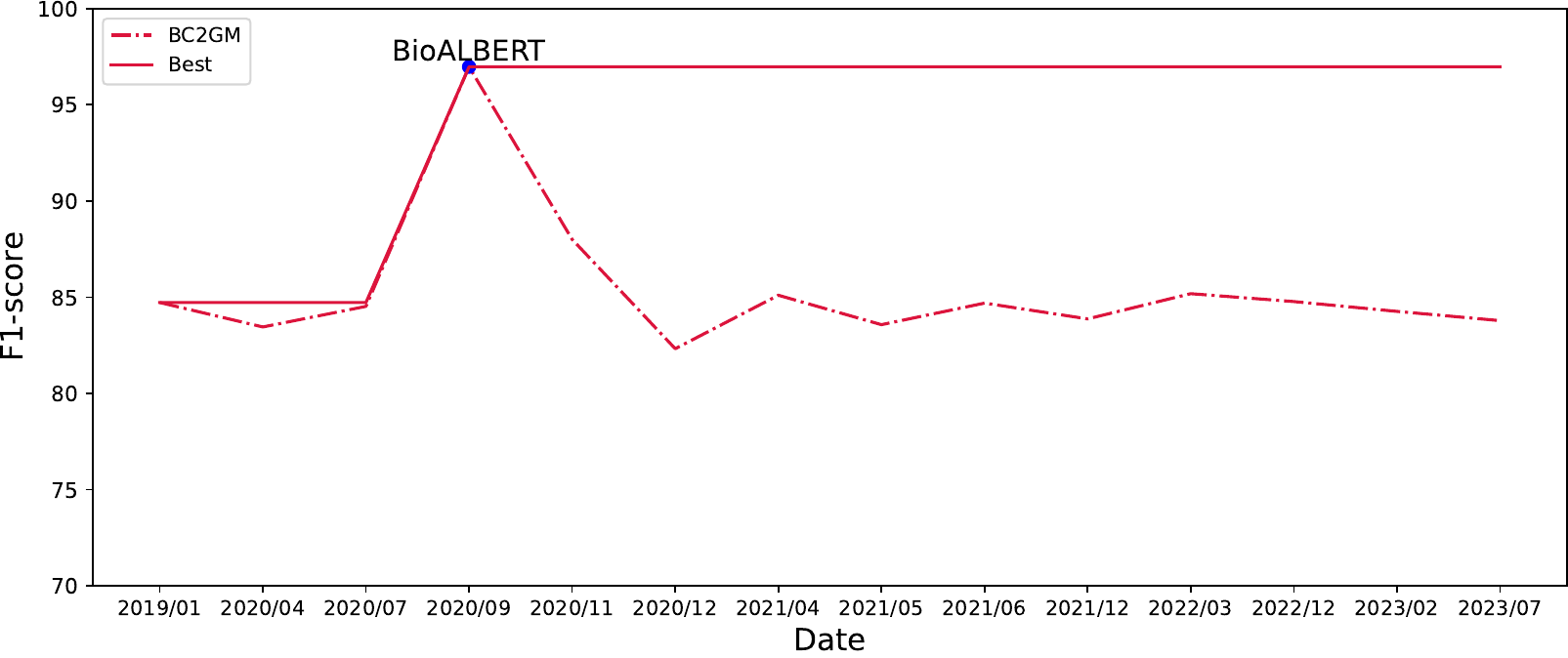}
        \caption{BC2GM}
    \end{subfigure}

    \caption{Details of performance changes for five NER datasets: (a) NCBI-disease, (b) BC5CDR-disease, (c) JNLPBA, (d) BC5CDR-chemical, and (e) BC2GM.}
    \label{result_ner_details}
\end{figure}

\newpage

\section{List of Abbreviations Used in This Paper}

\begin{table}

  \caption{List of abbreviations used in this paper.
  }
  \label{app_abbreviation}

  \begin{tabular}{l | l  | l}
    \toprule
  \textbf{Abbreviation} &  \textbf{Details}  &  \textbf{Defined in Section}   \\
    \midrule

PLMs &  Pre-trained Language Models &  1. Introduction  \\

LLMs & Large Language Models  &  1. Introduction  \\

NER &  Named Entity Recognition &   1. Introduction \\

RE & Relation Extraction  &   1. Introduction \\

QA &  Question-Answering &  1. Introduction  \\

SOTA &  state-of-the-art &   1. Introduction \\

SciLMs &  LMs for processing scientific texts &  1. Introduction  \\



\midrule
MLM &  Masked Language Modeling &  2.1. Existing LM Architectures  \\

NSP & Next Sentence Prediction  &  2.1. Existing LM Architectures  \\

SOP & Sentence Order Prediction  &  2.1. Existing LM Architectures  \\

RTP &  Replaced Token Prediction &  2.1. Existing LM Architectures  \\

NTP &   Next Token Prediction &  2.1. Existing LM Architectures  \\

GNN &  Graph Neural Network  &  2.1. Existing LM Architectures  \\

\midrule
KG & Knowledge Graph  &  2.2. Existing Tasks  \\

NLI & Natural Language Inference  &  2.2. Existing Tasks  \\

\midrule
SNS & Social Network Sites  &  2.3. Scientific Text  \\

\midrule 
Bi-LM &  Bidirectional Language Modeling  &  3. Existing SciLMs  \\

\midrule 
KBs & Knowledge Bases  &  5.1. Foundation SciLMs  \\

 


  \bottomrule
\end{tabular}

 \end{table}

\end{document}